\pdfoutput=1

\documentclass[11pt]{article}

\usepackage{ACL2023}

\usepackage{times}
\usepackage{latexsym}
\usepackage{booktabs}
\usepackage{graphicx}
\usepackage{subcaption}
\usepackage{multicol}
\usepackage{float}
\usepackage{eucal}
\usepackage{amsmath, amsthm, amssymb, amsfonts}
\usepackage[T1]{fontenc}

\usepackage[utf8]{inputenc}

\usepackage{microtype}

\usepackage{inconsolata}

\title{Model Interpretability and Rationale Extraction by Input Mask Optimization}

\author{Marc Brinner \and Sina Zarrieß \\
  Bielefeld University\\
  Faculty for Linguistics and Literary Studies\\
  \texttt{\{marc.brinner,sina.zarriess\}@uni-bielefeld.de}}

\begin{document}
\maketitle
\begin{abstract}
Concurrent with the rapid progress in neural network-based models in NLP, the need for creating explanations for the predictions of these black-box models has risen steadily. Yet, especially for complex inputs like texts or images, existing interpretability methods still struggle with deriving easily interpretable explanations that also accurately represent the basis for the model's decision. To this end, we propose a new, model-agnostic method to generate extractive explanations for predictions made by neural networks, that is based on masking parts of the input which the model does not consider to be indicative of the respective class. The masking is done using gradient-based optimization combined with a new regularization scheme that enforces sufficiency, comprehensiveness, and compactness of the generated explanation.
Our method achieves state-of-the-art results in a challenging paragraph-level rationale extraction task, showing that this task can be performed without training a specialized model.
We further apply our method to image inputs and obtain high-quality explanations for image classifications, which indicates that the objectives for optimizing explanation masks in text generalize to inputs of other modalities.
\end{abstract}

\section{Introduction}

Black-box machine-learning models like transformers \citep{22} or convolutional neural networks \citep{23} are state-of-the-art in natural language processing and computer vision. Their complexity enables them to perform well on a variety of tasks, but this comes at the cost of a lack of interpretability: The question of why a model made a specific prediction cannot be answered reliably. Especially if such black-box models are used in critical real-world applications (e.g., in the medical domain), this creates a demand for methods that explain network predictions while fulfilling a variety of requirements, like being easy to implement, model and task agnostic, faithful to the inner workings of the network, and producing results that are easily interpretable for humans.

To this end, a variety of interpretability methods have been proposed \cite{49, 50}, but as the aforementioned requirements are often at odds, at least one of them often remains unfulfilled.
Reasons for this include the reliance on complex message passing schemes that require laborious implementations (e.g., \citealp{5}, \citealp{6}), the applicability only to specific model architectures (e.g., \citealp{TAM}, \citealp{40}), or the fact that explanations often highlight individual, disconnected input features (e.g., standard gradient-based saliency), which contradicts human intuition of a sensible explanation (compare Section \ref{subsec:interpretability} for details).

As an example, in a text classification setting, interpretability methods often highlight individual words that explain the prediction, but do not include their context \citep{51}, even though the context of a word is crucial in determining its meaning: 
The word "good" influences the prediction in a completely different way if it is preceded by the word "not", meaning that this context has an impact on the classification and should therefore be part of the rationale. Notably, this holds true even in the absence of such modifiers, since the context must be available to confirm this absence.

In this work, we propose a new method for model explainability that is able to identify parts of the input that are, on the one hand, most indicative of a class and, on the other hand, perceived as a sensible rationale by humans.
Our method is applicable to all input types that define a spatial structure between individual features (e.g., texts, images) and builds on the assumption that interpretable explanations correspond to smooth and connected regions of features with respect to this spatial structure. It uses numerical optimization to mask out parts of the input that the model does not consider indicative of the class of interest, thus leaving only the parts of the input that are indicative of this class. The masking is done using gradient-based optimization combined with a new regularization scheme that enforces sufficiency, comprehensiveness, and compactness of the generated explanation \citep{15}, three criteria that have been established in the domain of rationale extraction but are less common in network interpretability methods. In this way, our method bridges the gap between model interpretability and rationale extraction, thereby showing that the latter of which can be performed without training a specialized model, only on the basis of a trained classifier.

\section{Background}
\label{sec:background}

Methods that explain the predictions made by black-box models to users  
can be broadly categorized into (i) interpretability methods that aim at creating explanations for existing classifiers after they have been trained (Section \ref{subsec:interpretability}) and (ii) rationale extraction approaches that are designed to create a rationale as a model output in addition to the usual label prediction (Section \ref{subsec:rationale}). Our interpretability method relies on gradient-based input optimization, discussed in detail in Section \ref{subsec:inputopt}.

\subsection{Neural Network Interpretability}
\label{subsec:interpretability}

Interpretability methods usually assign importance scores to features or parts of a given input, 
indicating how relevant the respective feature is for making the prediction.
Many early methods focus on convolutional neural networks and use backpropagation-like procedures to compute saliency scores for each input feature. \citet{2} use the network's gradient at the input image as saliency scores, while \citet{7}'s integrated gradients method sums over gradients at different inputs that are created by gradually transforming a neutral input into the input of interest. The DeconvNet architecture \citep{1} and the guided backpropagation algorithm \citep{3} again rely on a single evaluation but change the standard gradient computation to produce visually improved importance maps.
Attribution methods like layer-wise relevance propagation \citet{4} extend this idea by defining a backward pass that redistributes the total function value layer-wise backwards using a propagation rule that makes the total relevancy within each layer add up to the function value that is to be explained. 
Deep Taylor Decomposition \citep{5} and DeepLIFT \citep{6} then introduced different rules for redistributing the relevance between layers. For transformer models, methods like \citep{40} track the attention flow through the network.
This has been extended to incorporate information from attribution methods like the Deep Taylor Decomposition to more accurately identify neurons that have a strong influence on the final prediction \citep{29, 28}.

Other well-known explainability methods rely on input perturbations. LIME \citep{8} identifies important input features by perturbing the input, observing the change in the model predictions, and fitting an interpretable model to the observed data, while other methods occlude parts of the input to detect features that are important for the classification \citep{1,9,11,10}.

A further approach to model interpretability is to generate an input that maximally activates specific neurons, thereby yielding insights about the responsibilities of these neurons, as was done for CNNs by \citet{2}. \citet{13} then used a similar idea to remove class-indicative information from input images to detect the parts of the image responsible for the classification.

The method we propose in this paper differs from standard gradient-based techniques by not relying on evaluations at a single point or at fixed perturbations, but at points that are determined by a dynamic optimization process. This control via optimization is also a key difference from methods that rely on random permutations or masking of input features. Compared to message-passing schemes and model-specific methods (like methods for transformer interpretability), relying only on the gradient makes our method applicable to models with a variety of architectures and layer types without requiring additional implementational effort. 

\subsection{Rationale Extraction}
\label{subsec:rationale}

The task of \textit{rationale extraction}, also commonly referred to as \textit{selective rationalization}, is concerned with designing models that can produce human-interpretable rationales in addition to the usual model output \citep{14}, with the domain usually being textual inputs and the rationales being a subset of the input text that is determined to be responsible for the prediction. \citet{14} approached this task by developing a two-step procedure in which a proposal network extracts a rationale from the input text and a subsequent classification network only has access to the rationale to make the final prediction. 
By training this model end-to-end, the proposal network learns to extract the most useful text fragments from the input, which thus corresponds to an explanation for the classification. Later, \citet{15} proposed three criteria that rationales should satisfy to be perceived as sensible:
\begin{description}
    \item[Sufficiency:] The rationale should be sufficient to correctly classify the sample only by its rationale.
    \item[Comprehensiveness:] All relevant information should be contained in the rationale, meaning that the correct label can not be inferred by just considering the words not included in the rationale.
    \item[Compactness:] The rationale should be sparse but should nevertheless consist of consecutive text fragments instead of single words.
\end{description}
\citet{15}'s methods enforce these criteria through regularizers and by using a complement predictor that predicts the correct label based on all words that are not part of the rationale. Training the proposal network to fool the complement predictor then enforces the comprehensiveness constraint. Other approaches extend this and extract class-dependent rationales \citep{16} or select complete paragraphs as rationales \citep{17}.

The two main differentiators of rationale extraction models to the interpretability methods discussed in Section \ref{subsec:interpretability} are that, one the one hand, models are explicitly trained to produce rationales instead of creating them post hoc, and, on the other hand, the focus is on creating human interpretable rationales while the focus for interpretability methods often is on mathematical faithfulness measures.

Our method combines the focus on faithfulness with the desire for human interpretability to create rationales that faithfully explain model predictions post hoc \textit{and} correspond to human rationales, as these properties substantially enhance the usefulness of explanations for many applications.

\subsection{Input Optimization}
\label{subsec:inputopt}

As mentioned in Section \ref{subsec:rationale}, optimization of input images for CNNs has been used to explain the responsibilities of specific neurons, but notably, the resulting images do not resemble naturally occurring images. This is caused by the huge complexity and highly nonlinear behavior of neural networks, leading to the property of having unpredictable behavior on out-of-domain inputs that quickly arise during the optimization. 
In different experiments, this has led to behaviors like making highly confident class predictions for images that resemble random noise \citep{41} or predicting a completely different class after adding almost imperceivable noise to a given image \citep{19}. Unconstrained optimization of the input to a neural network to optimize the activation of specific neurons will therefore inevitably result in inputs that are out-of-domain, do not resemble natural images, or seem downright counter-intuitive. 
Different strategies for mitigating this problem in the context of input optimization exist (e.g., the use of GANs, \citealp{20}), with the most common being extensive regularization to prevent high-frequency information in images from influencing the prediction \citep{12, 21} or using lower-resolution inputs and blurring to limit the degrees of freedom within the optimization \citep{13}. 

In this study, we use input optimization to perform model interpretability by optimizing a mask to suppress all parts of a given input that a given model does not consider indicative of the given class. Compared to \citep{13}, we propose a new optimization objective as well as a new regularization scheme that allows for the creation of more detailed masks. Additionally, we expand the scope of input optimization methods from the domain of images to text processing.

\section{MaRC}
\label{sec:MaRC}
In this section, we introduce \textbf{MaRC}, our framework for \textbf{Ma}sk-based \textbf{R}ationale \textbf{C}reation. Section \ref{sec:method} develops the general framework.  Sections \ref{sec:method_texts} and \ref{sec:method_images} address the specificities of applying MaRC to texts and images, respectively.

\subsection{Method}
\label{sec:method}

We design an interpretability method that detects parts of an input $x$ that a model  $M$ considers most indicative of a specific class $c$. 
We assume an input $x$ with $n$ input features, each of which could be high-dimensional, e.g., token embeddings or pixels with color channels.
The main idea of the approach is to detect input features that are highly indicative of class $c$ by replacing as much of the input as possible with an uninformative input $b$, i.e., an input that the model does not consider indicative of any class, while having the model assign a high score for class $c$ to the altered input.
We define a mask $\lambda \in \mathbb{R}^n$, $\lambda_i \in [0, 1]$ to obtain a masked input  $\tilde{x}$ in the following way:
\begin{align}\label{eq:1}
    \tilde{x} = \lambda \cdot x + (1 - \lambda) \cdot b
\end{align}
When $\lambda_i$ is close to $1$, feature $i$ is mostly retained in  $\tilde{x}$‚ while $\lambda_i$ values close to $0$ replace feature $i$ almost completely with the uninformative $b$.

MaRC tackles this masking as an optimization problem: it optimizes $\lambda$ to obtain rationales that fulfill the properties of sufficiency, comprehensiveness, and compactness (compare Section \ref{subsec:rationale}). 
It models these properties via dedicated regularizers, which we will develop step by step in the following.

\paragraph{Sufficiency}
 We want to find a mask $\lambda$ such that the probability that model $M$ assigns to $\tilde{x}$ for class $c$ is close to $1$.
We optimize this criterion as follows:
\begin{align}
\label{eq:opt_suff}
    \underset{\lambda \in [0, 1]^n}{\textrm{arg\,min}}\:\:\: -\mathcal{L}(\tilde{x}, c) + \underbrace{\alpha_{\lambda} \left[\frac{1}{n}\sum_{i=1}^{n} \lambda_i\right]^2}_{\Omega_{\lambda}}
\end{align}
Here, $\mathcal{L}(\tilde{x}, c)$ is a scoring function for $c$ under $M$ and $\Omega_{\lambda}$ is a sparsity regularizer that enforces 
the detection of the smallest set of input features that still induces a high score for $c$.
An obvious choice for $\mathcal{L}(\tilde{x}, c)$ is the log-likelihood of $c$, maximizing the probability of $c$ under $M$ and leading $\lambda$ to highlight \textit{class-discriminative information}, i.e., 
input features that indicate \textit{only} class $c$.
A different choice would be the logarithm of the sigmoid of the logit for $c$, which does not suppress other classes and therefore leads $\lambda$ to highlight \textit{class-indicative information}, i.e., all input features relevant for $c$, even if they are indicative of other classes as well. 
In both cases, $M$ considers $\tilde{x}$ to be highly indicative of class $c$, thereby fulfilling the sufficiency criterion.
\paragraph{Comprehensiveness} Optimizing Equation \ref{eq:opt_suff} leads to sufficiency but not comprehensiveness, as the smallest set of highly indicative input features is detected.
To detect \textit{all} information relevant for $c$, 
we introduce the \textit{complement of rationale} \citep{15}:
\begin{align}
    \tilde{x}^\mathsf{c} = (1-\lambda) \cdot x + \lambda \cdot b
\end{align}
which leaves features unmasked that were masked for $\tilde{x}$. Minimizing the score of $\tilde{x}^\mathsf{c}$ for $c$ enforces all parts that indicate class $c$ to be masked in $\tilde{x}^\mathsf{c}$ (meaning that they will be unmasked in $\tilde{x}$), resulting in the following optimization:
\begin{align}
    \underset{\lambda \in [0, 1]^n}{\textrm{arg\,min}}\:\:\: -\mathcal{L}(\tilde{x}, c) + \mathcal{L}(\tilde{x}^\mathsf{c}, c) + \Omega_{\lambda}
\end{align}
This formulation combines the "deletion game" and "preservation game" that were introduced by \citet{13} but treated as separate objectives. Optimizing the mask with respect to both objectives greatly supports the detection of precise boundaries of the relevant features.

\begin{figure*}
    \centering
    \includegraphics[width=0.97\linewidth]{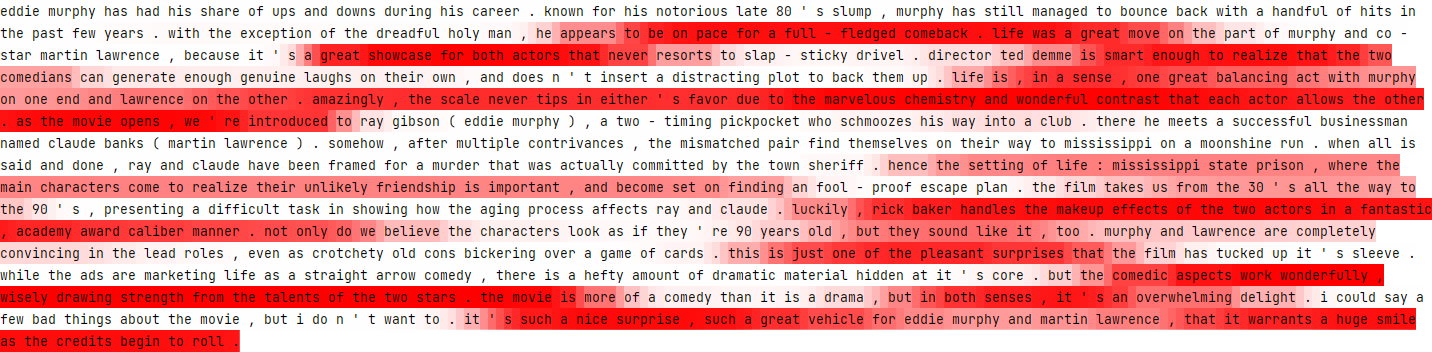}
    \caption{An exemplary rationale created by MaRC for the prediction of the \textit{positive} sentiment label.}
    \label{fig:my_label}
\end{figure*}

\paragraph{Compactness} The original compactness criterion states that a rationale shall consist of longer but fewer meaningful spans of \textit{text}. 
Here, we generalize this to all input types that possess a spatial structure that defines neighborhoods around input variables.
The underlying assumption is, that for these types of inputs, a feature is only meaningful in the context of its neighborhood, as, for example, is the case for single words in text or individual pixels in images, so that a sensible rationale must include larger groups of closely located features.

Thus, we now assume a general spatial structure on the input $x$ that defines distances $d(i, j)$ between the features $i$ and $j$, with features that are closer together having a higher chance of belonging to the same meaningful entity.
We enforce the selection of larger groups of features by reparameterizing our mask, i.e., 
 we introduce two new parameters, $w \in \mathbb{R}^n$ and $\sigma \in \mathbb{R}^n_{>0}$ from which the mask values $\lambda$ can subsequently be calculated. The optimization is then performed with respect to $w$ and $\sigma$.

The mask values $\lambda$ are mainly determined by $w$, in a way that $w_i$ largely determines the final value of $\lambda_i$. Crucially, $w_i$ now also influences the values of $\lambda$ around $i$, so that, for example, $\lambda_{i-1}$ and $\lambda_{i+1}$ are also strongly influenced by $w_i$. $\sigma_i$ then determines the strength and extent of $w_i$'s influence on its neighbors, as it parameterizes an unnormalized Gaussian placed at position $i$, so that the influence $w_{i \rightarrow j}$ of a weight $w_i$ onto $\lambda_j$ is then given by:
\begin{align}
    w_{i \rightarrow j} = w_i \cdot \exp\big(-\frac{d(i, j)^2}{\sigma_i}\big)
\end{align}
The final value for $\lambda_j$ is then calculated as follows:
\begin{align}
    \lambda_j = \textrm{sigmoid}(\sum_i w_{i \rightarrow j})
\end{align}
This parameterization of $\lambda$ enforces neighboring inputs to have similar values if the corresponding $\sigma$ values are large, which also plays a key role in regularizing the optimization to avoid the issues discussed in Section \ref{subsec:inputopt}.
Large values for $\sigma$ are softly enforced by introducing an additional regularizer:
\begin{align}
    \Omega_{\sigma} = -\alpha_{\sigma} \cdot \frac{1}{n} \sum_{i=1}^n \log(\sigma_i)
\end{align}
The logarithm was chosen to enforce positive values of $\sigma_i$ while gradually discounting the effect that increases in $\sigma_i$ have on the loss function. Notably, this regularizer does not enforce large values of $\sigma$ by means of hard constraints, meaning that low values and therefore sharper boundaries between mask values for neighboring features can be optimal if the other parts of the optimization objective support this behavior. This is in contrast to \citep{13}, who used a lower resolution mask in combination with upsampling and Gaussian blur to detect smooth masks, which does not allow for sharp masks even if they were optimal.

In summary, the final optimization objective looks as follows:
\begin{align}
\label{eq:general_objective}
    \underset{w, \sigma \in \mathbb{R}^n}{\textrm{arg\,min}}\:\:\: -\mathcal{L}(\tilde{x}, c) + \mathcal{L}(\tilde{x}^\mathsf{c}, c) + \Omega_{\lambda} + \Omega_\sigma
\end{align}
This objective can be optimized using stochastic gradient descent, but in practice, we found using an optimizer that incorporates momentum (e.g., Adam, \citealp{42}) to be key for avoiding local optima and obtaining optimal results.

\subsection{Textual Inputs}
\label{sec:method_texts}
As MaRC only requires the gradient of a model prediction at the input, it can be applied to all common text processing models. In the following, we discuss specific aspects of using MaRC with state-of-the-art transformer architectures like BERT \cite{30}.


As uninformative input $b$, we choose a sequence of \textit{PAD}-tokens of the same length as $x$. 
During training, 
the model learns to treat these tokens as uninformative since they are added to inputs irrespective of their content or the desired output.

As we want importance scores for each individual word, we define $n$ to be the number of words in the input sequence. Notably, this is different from the actual input dimension, as it is common to use WordPiece embeddings \cite{43} which could split words into multiple input tokens. In this case, we use parameter tying to only have a single parameter for all pieces of a word representation. The distance function is then simply defined as $d(i, j) = |i-j|$, with $i$ and $j$ being the positions of the words in the text.

Finally, we found that introducing noise into the optimization process is beneficial for regularization (see Section \ref{subsec:inputopt} for discussion of regularization in input optimization). Thus, for text inputs, we add Gaussian noise to $\tilde{x}$ and $\tilde{x}^\mathsf{c}$ and randomly set mask values to $0$ or to $1$ in each optimization step.

\subsection{Image Inputs}
\label{sec:method_images}

Image inputs also fulfill the requirements on the presence of a spatial structure that is needed for our method.
They also provide natural choices for uninformative inputs, as uniformly colored images can generally be assumed to be uninformative in most prediction settings. Therefore, obvious choices for $b$ would, for example, be a white image, a black image, or an image of the mean color within the given dataset. A different option is to remove usable information from the input image by blurring it and using this blurred image as uninformative input \cite{13}. As parts of the input image could have the same color as the uninformative input (which renders the corresponding mask values meaningless) and even uniformly colored patches could be seen as informative by neural networks, we chose to alter the optimization objective to be the average over different choices for $b$, with $B$ being the set of all uninformative inputs:
\begin{align}
\label{eq:objective_image}
    \nonumber\underset{w, \sigma \in \mathbb{R}^{w \times h}}{\textrm{arg\,min}}\:\:\: \frac{1}{|B|}\sum_{b \in B}\mathcal{L}&(\tilde{x}(b, \tau), c) - \mathcal{L}(\tilde{x}^\mathsf{c}(b, \tau), c)\\ + \Omega_{\lambda} &+ \Omega_\sigma + \Omega_{\textrm{NB}}
\end{align}
As images generally have more variables and therefore more degrees of freedom in the optimization, further regularization is needed to obtain sensible optimization results. To this end, this formulation includes an additional regularizer $\Omega_{\textrm{NB}}$, which denotes the average squared difference between mask values that are neighboring with respect to the 8-connected grid structure of the image, weighted by a corresponding parameter $\alpha_{\textrm{NB}}$.

To complete the specification of the optimization problem, we define the distance function $d$ between two pixels to be the euclidean distance between their two-dimensional position vectors in the image grid. In contrast to the textual inputs, the introduction of noise to the optimization process did not prove to be beneficial.

\begin{table}[t]
  \centering
  \begin{tabular}{lccccc}
    \toprule
    \hspace{-3px}Method     \hspace{-15px}& \hspace{-15px}Token F1\hspace{-10px} & mAP\hspace{-5px} & IoU F1\hspace{-7px} & Suff.\hspace{-3px} $\downarrow$\hspace{-7px} & Comp.\hspace{-4px} $\uparrow$\hspace{-7px} \\
    \midrule
    \hspace{-3px}MaRC \hspace{-15px}& \hspace{-10px}\textbf{.473} & \textbf{.469} & \textbf{.163} & .028 & .518 \\
    \hspace{-3px}Occlusion\hspace{-15px} &\hspace{-10px} .432 & .448 & .125 & .022 & .415 \\
    \hspace{-3px}$\textrm{Saliency}_{\textit{\small{n}}}$\hspace{-15px} & \hspace{-10px}.435 &  .392 & .04 & .132 & .287 \\
    \hspace{-3px}$\textrm{Saliency}_{\textit{\small{s}}}$\hspace{-15px} & \hspace{-10px}.425 & .340 & .076 & .260 & .246 \\
    \hspace{-3px}$\textrm{InXGrad}_{\textit{\small{n}}}$ \hspace{-15px}& \hspace{-10px}.436 &  .396 & .040 & .136 & .292 \\
    \hspace{-3px}$\textrm{InXGrad}_{\textit{\small{s}}}$ \hspace{-15px}& \hspace{-10px}.425 & .340 & .084 & .239 & .248 \\
    \hspace{-3px}$\textrm{Int. Grads}_{\textit{\small{n}}}$\hspace{-15px} & \hspace{-10px}.428 & .369 & .036 & .122 & .274 \\
    \hspace{-3px}$\textrm{Int. Grads}_{\textit{\small{s}}}$\hspace{-15px} &\hspace{-10px} .431 & .381 & .071 & .048 & .528 \\
    \hspace{-3px}LIME\hspace{-15px} & \hspace{-10px}.436 & .380 & .076 & .047 & .496 \\
    \hspace{-3px}Shapley \hspace{-15px}&\hspace{-10px} .428 & .439 & .079 & \textbf{-.015} & \textbf{.728} \\
    \cmidrule(r){1-6}
    Noise & .454 & .450 & .139 & .034 & .487 \\
    $\Omega_{\lambda}$ & .349 & .350 & .046 & .120 & .266 \\
    $\Omega_{\sigma}$ & .447 & .425 & .091 & .036 & .535 \\
    $\mathcal{L}(\tilde{x}^\mathsf{c}, c)$ & .396 & .436 & .123 & .052 & .304 \\
    \bottomrule
  \end{tabular}
  \caption{Results on rationale extraction on the movie reviews dataset \cite{26}, including faithfulness evaluation. See Section \ref{sec:exp_detials} for an overview of the methods tested and for experimental details.}
  \label{tab:rationales}
\end{table}
\begin{table}[t]
  \centering
  \begin{tabular}{lcccc}
    \toprule
    &\multicolumn{2}{c}{ResNet-101} &\multicolumn{2}{c}{ViT-B/16}\\
    \midrule Method\hspace{-15px} &\hspace{-15px}  Suff. $\downarrow$  &\hspace{-5px}  Comp.$\uparrow$ \hspace{-5px} &  Suff. $\downarrow$ & \hspace{-5px} Comp.$\uparrow$ \hspace{-0px} \\
    \midrule
    MaRC \hspace{-15px} & \hspace{-15px} \textbf{.196} & .612 & \textbf{.139} & .596\\
    M-Perturb\hspace{-15px} & \hspace{-15px} .260 & .605 & .174 & .572 \\
    Grad-CAM\hspace{-15px} & \hspace{-15px} .197 & .600 & .161 & .640 \\
    Exc-BP\hspace{-15px} & \hspace{-15px} .302 & .600 & - & - \\
    Saliency\hspace{-15px} &\hspace{-15px}  .442 & .599 & .355 & .528\\
    InputXGrad\hspace{-15px} & \hspace{-15px} .430 & .586 & .366 & .506\\
    Guided BP\hspace{-15px} & \hspace{-15px} .343 & .630 & - & -\\
    Intgr. Grads \hspace{-15px}&\hspace{-15px}  .344 & \textbf{.641} & .261 & .641 \\
    Occlusion \hspace{-15px}&\hspace{-15px}  .324 & .606 & .194 & .486 \\
    Attention\hspace{-15px} &\hspace{-15px}  - & - & .241 & .562 \\
    Attribution \hspace{-15px}&\hspace{-15px}  - & - & .176 & .608 \\
    Rollout\hspace{-15px} &\hspace{-15px}  - & - & .205 & .580 \\
    TAM \hspace{-15px}& \hspace{-15px} - & - & .146 & \textbf{.658} \\
    \bottomrule
  \end{tabular}
  \caption{Results for the faithfulness evaluation of different explainability methods for ResNet-101 and ViT-B/16. Compare Section \ref{sec:exp_detials} for an overview of the methods tested and for experimental details.}
  \label{tab:img_faith}
\end{table}

\section{Experiments on Rationale Extraction}
\label{sec:rationale_exp}

We evaluate MaRC on rationale extraction, a task that is concerned with predicting the correct label for a given textual input while also providing a subset of the input as a rationale for the prediction. 
\subsection{Data}
\begin{figure*}[t]
\makebox[0.99\linewidth][c]{
\begin{subfigure}{.163\textwidth}
  \centering
  \includegraphics[width=0.97\linewidth]{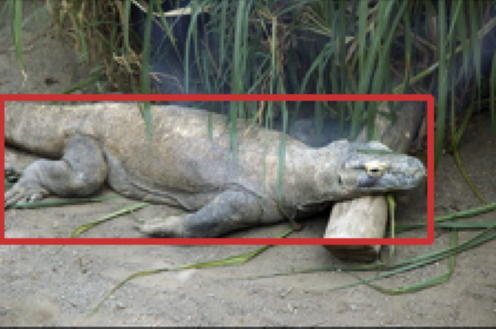}
  \caption{Input}
\end{subfigure}
\begin{subfigure}{.163\textwidth}
  \centering
  \includegraphics[width=0.97\linewidth]{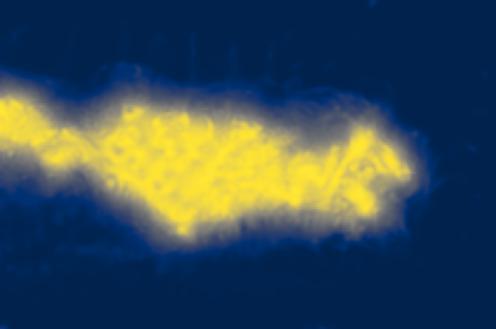}
  \caption{MaRC}
\end{subfigure}
\begin{subfigure}{.163\textwidth}
  \centering
  \includegraphics[width=0.97\linewidth]{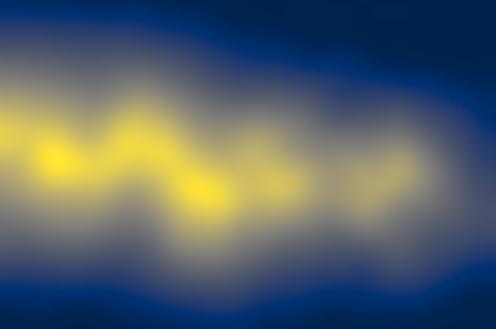}
  \caption{M-Perturb}
\end{subfigure}
\begin{subfigure}{.163\textwidth}
  \centering
  \includegraphics[width=0.97\linewidth]{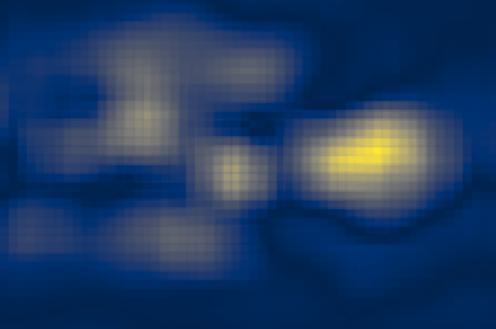}
  \caption{Occlusion}
\end{subfigure}
\begin{subfigure}{.163\textwidth}
  \centering
  \includegraphics[width=0.97\linewidth]{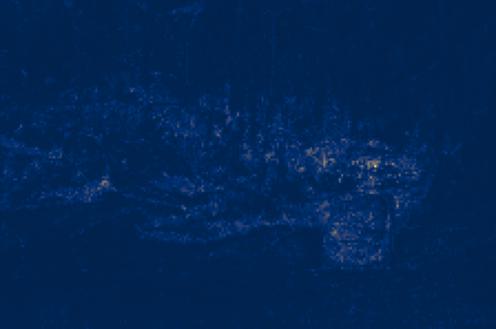}
  \caption{Integr. Grads}
\end{subfigure}
\begin{subfigure}{.163\textwidth}
  \centering
  \includegraphics[width=0.97\linewidth]{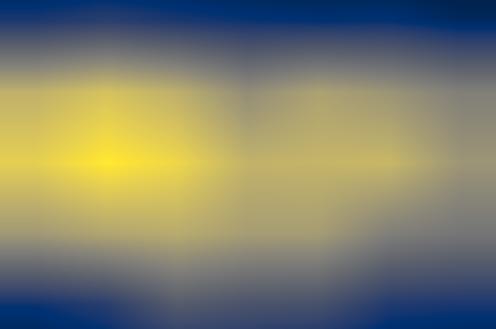}
  \caption{Grad-CAM}
\end{subfigure}
}
\vspace{-3px}
\caption{Comparison of masks created for ResNet-101 by different explainability methods.}
\label{fig:resnet}
\vspace{-5px}
\end{figure*}

We use the movie review data set \citep{27} with 2000 movie reviews annotated with sentiment labels (\textit{positive} or \textit{negative}) as well as span-level rationales. 
We test on the additional rationales created by \citet{26}, which are more comprehensive and thus, on average, comprise a much larger fraction of words (7.2\% vs. 31.4\%). As our approach is designed for extracting span-level rationales and most other datasets for rationale extraction are not annotated on span-level \citep{26}, this is the only dataset suitable for an evaluation of MaRC. We use a standard $\textrm{BERT}_{\textit{base}}$ model \citep{30} and train it as a standard binary classification model on the training data, therefore only using the class labels and not the annotated rationales.

\subsection{Evaluation}

There are two common ways of evaluating the rationales produced by different models \citep{26, 34}:

\begin{enumerate}
    \item \textbf{Agreement with human annotator rationales:} A strong overlap between rationales given by human annotators and rationales produced by an explainability model is a good indicator that sensible rationales have been selected. Additionally, similarity to human rationales could be considered a desirable property (depending on the use case), even if it is not perfectly in line with the actual reasoning process of the neural network.
    \item \textbf{Faithfulness:} Ideally, the rationales produced by a model fulfill the conditions of sufficiency and comprehensiveness, meaning that they actually reveal the information that the model considered indicative for the predicted label.
\end{enumerate}
Different metrics exist to evaluate the performance of approaches that produce "soft" scores (i.e., continuous values) or binary values as the output of the rationale generation. As we see use cases for both outputs, we evaluate our approach with respect to both. To create a binary mask from the continuous mask values that MaRC produces, we train a kernel regression model to predict the optimal percentage of words that need to be included in the rationale (described in Appendix \ref{sec:exp_detials}), which we do in the same way for all methods tested in this study.

To evaluate the agreement with human rationales, we calculate the token F1 score for the binary masks by using precision and recall of the "positive" class of words belonging to the rationale, while the soft-scoring models are evaluated using the mean average precision (mAP). To evaluate the agreement of larger detected spans with the spans present in the human rationales, we evaluate the IoU F1 score that counts a ground-truth span as correctly detected if there is a predicted span with an IoU of over $0.5$, which again allows for the calculation of an F1 score for the "positive" class of detecting the spans. These three metrics were used in the ERASER benchmark \cite{26}, which also proposed metrics to evaluate sufficiency and comprehensiveness. For these metrics, we slightly deviate from their evaluation metrics by evaluating these scores for a given sample $x$ and rationale $r$ in the following way:
\begin{align}
\label{eq:comp}
    \textrm{comp}(x, r) = \frac{1}{19} \sum_{i=1}^{19} M(x) - M(x \backslash r_i)
\end{align}
\begin{align}
\label{eq:suff}
    \textrm{sufficiency}(x, r) = \frac{1}{19} \sum_{i=1}^{19} M(x) - M(r_i)
\end{align}
Here, $M(x)$ denotes the class probability prediction (for the ground-truth class) of our model, $r_i$ denotes the top $(i\cdot 5)\%$ of words according to the soft rationale scores (all other words are removed), and $x \backslash r_i$ denotes sample $x$ with all words that belong to $r_i$ removed, where we "remove" words by replacing the corresponding tokens with \textit{PAD}-tokens. Therefore, the comprehensiveness score evaluates, how much removing the rationale decreases the model performance (higher scores are better) while the sufficiency score evaluates how well the correct label can be predicted from the rationale alone (lower scores are better).
\begin{figure*}[t]
\makebox[0.99\linewidth][c]{
\begin{subfigure}{.163\textwidth}
  \centering
  \includegraphics[width=0.97\linewidth]{images/1/0.png}
  \caption{Input}
\end{subfigure}
\begin{subfigure}{.163\textwidth}
  \centering
  \includegraphics[width=0.97\linewidth]{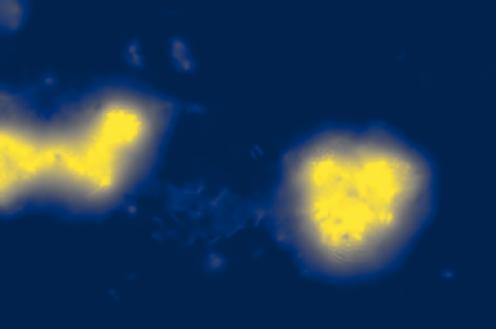}
  \caption{MaRC Softmax}
\end{subfigure}
\begin{subfigure}{.163\textwidth}
  \centering
  \includegraphics[width=0.97\linewidth]{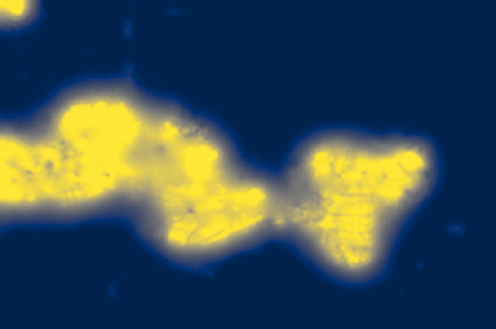}
  \caption{MaRC Sigmoid}
\end{subfigure}
\begin{subfigure}{.163\textwidth}
  \centering
  \includegraphics[width=0.97\linewidth]{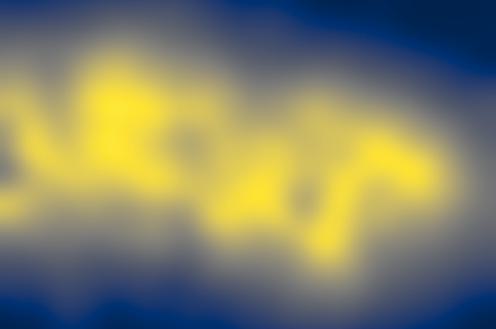}
  \caption{M-Perturb}
\end{subfigure}
\begin{subfigure}{.163\textwidth}
  \centering
  \includegraphics[width=0.97\linewidth]{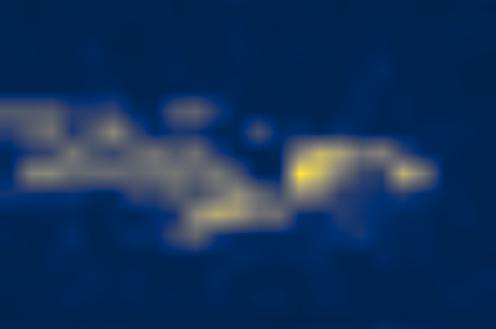}
  \caption{TAM}
\end{subfigure}
\begin{subfigure}{.163\textwidth}
  \centering
  \includegraphics[width=0.97\linewidth]{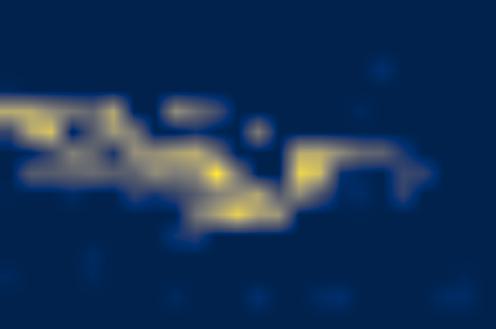}
  \caption{Grad-CAM}
\end{subfigure}
}
\vspace{-2px}
\caption{Comparison of masks created for ViT-B/16 by different explainability methods.}
\label{fig:vit}
\vspace{-5px}
\end{figure*}

\subsection{Results}

The evaluation results are displayed in the upper part of Table \ref{tab:rationales} (see Appendix \ref{sec:exp_detials} for more details on the setup). We compare MaRC against other interpretability methods that are commonly used in the context of NLP but omit specialized rationale extraction models as they (i) usually produce binary masks, making it impossible to perform the soft-scoring evaluation, and (ii) do not produce explanations for existing models, which makes the faithfulness evaluation inapplicable.

We see that MaRC achieves state-of-the-art results on all measures that evaluate agreement with human rationales, i.e. Token F1, mAP, and IoU F1, showing that MaRC is the best method for obtaining rationales that match human intuition. Especially with respect to the IoU F1 score, MaRC outperforms all other methods by a large margin, even though the hyperparameters for other methods were set to explicitly support high scores in this measure (e.g., masking larger spans for occlusion and LIME). This highlights that MaRC is suitable for detecting span-level rationales in a  paragraph-long text that agree with spans that humans annotate, without being trained to do so and without additional model components as in state-of-the-art rationale extraction models. 

For sufficiency and comprehensiveness, MaRC also achieves impressive results, being outperformed in both metrics only by Shapley value sampling. The excellent performance of this method with regard to these evaluation metrics is not surprising, though, as it is based on choosing a random permutation of input features, adding them successively to the input, and using the change in the model's output as the resulting score. This method 
is very closely connected to the sufficiency and comprehensiveness calculations, thereby rendering the great results of this method unsurprising. It should be noted, that multiple methods, including MaRC, achieve close to optimal results for sufficiency, as scores close to $0$ indicate that the removal of very few high-scoring tokens is enough to completely throw off the classifier.
Notably, MaRC can produce good results while aiming to create human-like rationales, showing that this kind of rationale to some extent corresponds to the inner workings of the neural network.

We also conduct an ablation study that tests the importance of the different parts of the optimization objective by leaving these parts out in turn and reporting the results with the altered objective. The results are displayed in the lower part of Table \ref{tab:rationales}, with the "Method" column specifying with part of the objective is omitted. The full optimization objective is almost uniformly the best-performing variant, proving that all parts a essential to achieve optimal performance.

\section{Experiments on Image Classification} 

We evaluate MaRC on the task of creating rationales for classifications of ImageNet \citep{44} images. A visual comparison of masks created by MaRC and other interpretability approaches for ResNet-101 \citep{45} and the vision transformer ViT-B/16 \citep{46} is displayed in Figures \ref{fig:resnet} and \ref{fig:vit}, respectively (see Appendix \ref{sec:vis} for with more visualizations). 
We see that MaRC is able to produce sharp masks that often cover the complete object of interest in the image. 
For ViT-B/16, we include a visualization highlighting the distinction between  class-discriminative vs. class-indicative information (compare Section \ref{sec:method}): Figure \ref{fig:vit}b) used the softmax of the model output as scoring function, which leads MaRC to highlight only the head and tail of the animal, the two parts that the model uses to differentiate the correct class from the other classes. For Figure \ref{fig:vit}c), on the other hand, a mixture of the sigmoid of the class logit and the softmax of the model output was used with a ratio of 9:1, making the model highlight all parts in the image that indicate the ground-truth class, as long as they are not significant indicators of other classes.

We also evaluate the faithfulness of the explanation created by MaRC and a variety of other interpretability methods using the same metrics as for textual inputs. For this experiment, we use pretrained ResNet-101 and ViT-B/16 models on a random sample of $500$ ImageNet validation images, with further implementational details being described in Appendix \ref{sec:exp_detials}. As shown in Table \ref{tab:img_faith}, MaRC is the best-performing model with respect to sufficiency for both ResNet-101 and ViT-B/16, showing that the areas that MaRC highlights are indeed the areas that allow the model to predict the correct class based solely on these regions. With respect to comprehensiveness, MaRC achieves competitive results, only falling behind model-specific architectures that heavily use the knowledge about the inner workings of the model and the information flow inside it (e.g., transition attention maps (TAM), \citealp{TAM}), as well as two other methods in the form of Guided Backpropagation \citep{3} and Integrated Gradients \citet{7}. The latter two methods often predict individual pixels that are spread over many areas of the image as the most indicative input features, indicating that the removal of key pixels at different positions of the image is a good strategy to quickly decrease the classifier performance, an approach that MaRC is actively discouraged to pursuit.

\section{Conclusion}

We propose a new method for creating explanations for neural network predictions that are faithful to the model's reasoning process as well as being sensible with respect to human judgment. We achieve state-of-the-art results on the task of rationale extraction, achieve competitive or state-of-the-art results with respect to faithfulness, and provide visually sensible explanations for classifications of images. As MaRC is model-agnostic, we believe it to be a useful tool in many areas of machine learning that include textual or image inputs. We further believe that other domains can make use of MaRC, including multimodal tasks that, for example, combine textual and image inputs, as well as other domains that fulfill the requirements on the spatial structure of the input, like auditory data.

\section{Limitations}

Compared to other interpretability methods, MaRC is able to create explanations that more closely resemble human rationales. Nevertheless, the similarity to human rationales is always limited by the inner workings of the respective neural network: If a network's reasoning does not mirror human reasoning, the resulting rationales will be incomprehensible to humans.

Additionally, rationales created by MaRC are the result of a complete input optimization process. Therefore, the rationale creation usually requires hundreds of forward passes and gradient evaluations for the respective neural network, which makes the process of creating the rationale time-consuming and therefore infeasible for many real-time applications. On modern hardware, creating a rationale for $\textrm{BERT}_{\textit{base}}$ can take two to three minutes depending on the length of the input text, while ResNet-101 and ViT-B/16 are faster at about one minute.



\bibliography{anthology,custom}
\bibliographystyle{acl_natbib}

\appendix

\section{Experimental Details}
\label{sec:exp_detials}
The implementations of MaRC for the experiments conducted in this study is available at \url{https://github.com/inas-argumentation/Explainability}.
\subsection{Rationale Extraction}
We perform rationale detection using $\textrm{BERT}_{\textit{base}}$ (uncased) \cite{30}, which we train as a binary classifier for at most $20$ epochs on the first eight folds of the movie review dataset \cite{27}, with the ninth and tenth fold being used for validation and testing, respectively. For the optimization, we use the Adam optimizer and achieve a $96.5\%$ test set accuracy.

For rationale creation, the results from Table \ref{tab:rationales} as well as the example images for MaRC were created by using the optimization objective given by Equation \ref{eq:general_objective} with all specifications as described in Section \ref{sec:method_texts}, hyperparameters set to $\alpha_{\lambda} = 1$, $\alpha_{\sigma} = 1.2$ and $w$ and $\sigma$ being uniformly initialized to $1.2$ and $2$, respectively. We add zero-mean Gaussian noise to $\tilde{x}$ and $\tilde{x}^\mathsf{c}$ ($\sigma=0.03$) and randomly set $5\%$ of mask values to $0$ or $1$, respectively, in each optimization step. We use the log-likelihood of the respective class as scoring function. All these choices were made by using the validation split, with the measure of quality being visual coherence of the created explanation, as the data set does not offer a validation split with the same label distribution, thus making validation with respect to scores infeasible. Texts that surpass the limit of $510$ input tokens for $\textrm{BERT}_{\textit{base}}$ are split into multiple segments, with consecutive segments overlapping for $100$ tokens, and a separate mask is predicted for each segment. The resulting masks are concatenated, with the overlapping parts being linearly blended. We proceed in the same way for all other interpretability methods.

The following models and parameters were used in the method comparison:
\begin{itemize}
	\item \textit{Occlusion} \cite{1}: We chose to mask slightly larger spans of $5$ tokens as this produced smoother masks which resulted in higher IoU F1 scores. Occluded parts were replaced by \textit{PAD}-tokens.
	\item \textit{Saliency} \cite{2}: No special hyperparameter settings required.
	\item \textit{InXGrad} (Input times gradient, \citealp{6}): No special hyperparameter settings required.
	\item \textit{Int. Grads} (Integrated Gradients, \citealp{7}): We use a sequence of \textit{PAD}-tokens as background and do $50$ gradient evaluation steps per sample.
    \item \textit{LIME} \cite{8}: We do 50 function evaluations per sample. In each evaluation, we randomly select $5-13\%$ of tokens and replace them as well as the next three tokens with \textit{PAD}-tokens. We train a linear classifier and use the resulting weights as rationale.
    \item \textit{Shapley} (Shapley value sampling,  \citealp{33}): We evaluate the token contributions for $25$ feature permutations per sample. Removed tokens are replaced by \textit{PAD}-tokens.
\end{itemize}
We use the implementations provided by \cite{47} for all methods. All methods have access to the ground truth label and therefore do not have to rely on a correct classifier prediction.

For methods that produce scores for each entry of the embedding vector, we report results for two different methods of combining these scores to single values per token, with one being taking the vector norm (results are reported for the L1 norm, but we did not see a significant difference for the L2 norm), and the other one being summing over the resulting scores (indicated by subscript \textit{n} and \textit{s} in Table \ref{tab:rationales}, respectively). In the latter case, the resulting value was negated if the target label is $0$.

To evaluate the token F1 score and the IoU F1 score, we need to create a binary mask from the continuous scores produced by the different interpretability methods. We do this by selecting the top-scoring words as rationale, with the percentage of words that are selected being decided by a Nadaraya-Watson kernel regression model using an RBF kernel. The input to the kernel regression for a given sample is the percentage of words that have a score greater than a fixed threshold (a hyperparameter, here set to 0.1), while the output is the percentage of words to be selected as rationale. As we use the rationales from \citep{26} (who only annotated 200 samples) for our experiment, we do not have access to a separate training set to train the kernel regression, so we resort to a leave-one-out scheme to use the same set for training and testing.

For the faithfulness evaluation, we note that we deviate from the common practice of evaluating the area under the curve (AUC) (e.g., used by \citealp{10}) and instead take the average over the tested range of values. We do this, to accommodate for the possibility of negative scores in the sufficiency calculation, which undermine the theoretical foundation of the AUC. We also adapt the comprehensiveness calculation accordingly for consistency.

\subsection{ImageNet Explanations}

We use MaRC with the optimization objective given by Equation \ref{eq:objective_image}. We use pretrained ResNet-101 \cite{45} and vision transformer ViT-B/16 (\citealp{46}, input image size=$384$) models and use the following hyperparameter setting for MaRC:
\begin{itemize}
    \item ResNet-101: We set $\alpha_{\lambda}=0.6$, $\alpha_{\sigma}=1.2$, $\alpha_{\textrm{NB}}=10$ and initialize $w$ and $\sigma$ uniformly to $0.5$ and $1.2$ respectively. As ResNet models seem to treat uniformly colored images as uninformative, we chose $B$ to be a set containing a black image, a white image, and an image with the mean color from the dataset. As the scoring function, we chose the log of a combination of the softmax output of $c$ (weighted by $0.9$) and the sigmoid of the logit of $c$ (weighted by $0.1$).
    \item Vit-B/16: We set $\alpha_{\lambda}=0.25$, $\alpha_{\sigma}=1.2$, $\alpha_{\textrm{NB}}=10$ and initialize $w$ and $\sigma$ uniformly to $0.5$ and $1.2$ respectively. As the vision transformer often seems to interpret the uniformly colored backgrounds as indicative of specific classes, we instead opted to use a blurred version of the input image as $b$. As the scoring function, we use the log-likelihood of $c$.
\end{itemize}
All visualizations and experiments were, if not stated otherwise, conducted with these hyperparameter settings. 
We compared MaRC to the following methods:
\begin{itemize}
    \item M-Perturb \citep{13}: We used the original implementation and parameter settings by \citet{13} with minor adaptations to work with pytorch.
    \item \textit{Grad-CAM} \citep{38}: We used the implementation by \citet{48}.
    \item \textit{Exc-BP} (Excitation Backpropagation, \citealp{35}): We used the implementation available at \href{https://github.com/greydanus/excitationbp}{https://github.com/greydanus/excitationbp}
    \item \textit{Saliency} \cite{2}: We used the implementation by \citet{47}.
    \item \textit{InputXGrad} (Input times gradient, \citealp{6}): We used the implementation by \citet{47}.
    \item \textit{Guided BP} \cite{3}: We used the implementation by \citet{47}.
    \item \textit{Intgr. Grads} (Integrated Gradients, \citealp{7}):  We used the implementation by \citet{47}. As background, a blurred version of the input image was used, as this produced optimal results. For each sample, $100$ gradient evaluation steps were performed.
    \item \textit{Occlusion} \cite{1}:  We used the implementation by \citet{47}. We occluded patches of $\frac{1}{9}$ of the input image size and used a stride of $\frac{1}{56}$ of the input image size. Occluded patches were replaced by a blurred version of the input image.
    \item \textit{Attention} \cite{TAM}: We used the implementation by \citet{TAM}.
    \item \textit{Attribution} \cite{29}: We used the implementation by \citet{TAM}.
    \item \textit{Rollout} \cite{40}: We used the implementation by \citet{TAM}.
    \item \textit{TAM} (Transition attention maps, \citealp{TAM}): We used the implementation by \citet{TAM}.
\end{itemize}
All methods had access to the ground-truth label to create the explanation. The random sample of ImageNet validation images that was used in this experiment can be viewed at the GitHub page given at the beginning of Appendix \ref{sec:exp_detials}.

To evaluate model faithfulness, we used Equation \ref{eq:comp} and \ref{eq:suff}, with the difference that the evaluation for each sample and percentage was performed with respect to four uninformative inputs to prevent skewed evaluation results for methods that only work well with respect to one background. The four backgrounds we used were a white image, a black image, an image of the mean color of the dataset, as well as a blurred version of the input image.
\onecolumn
\section{Movie Review Rationale Examples}
\label{sec:vis_text}
\begin{multicols}{2}
\setlength{\parindent}{0pt}
This section includes four additional exemplary rationales created by MaRC on movie reviews from \citep{27}.
\vspace{200px}
\end{multicols}

\begin{figure}[H]
    \centering
    \includegraphics[width=0.99\linewidth]{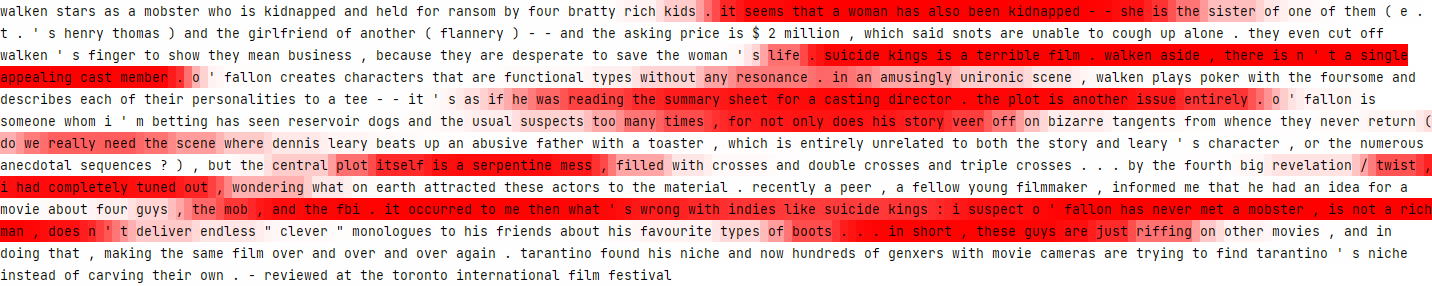}
    \caption{Rationale for a negative movie review.}
\end{figure}

\begin{figure}[H]
    \centering
    \includegraphics[width=0.99\linewidth]{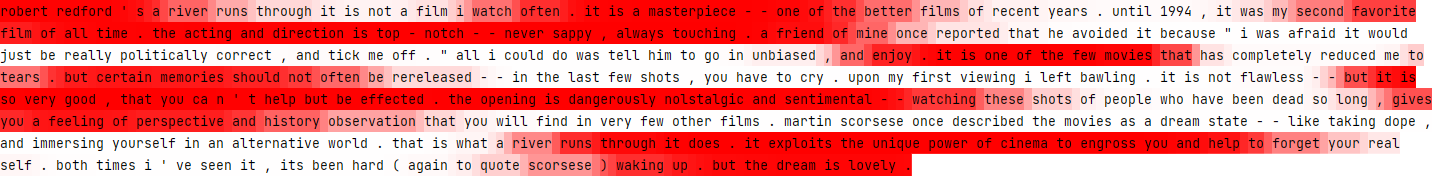}
    \caption{Rationale for a positive movie review.}
\end{figure}

\begin{figure}[H]
    \centering
    \includegraphics[width=0.99\linewidth]{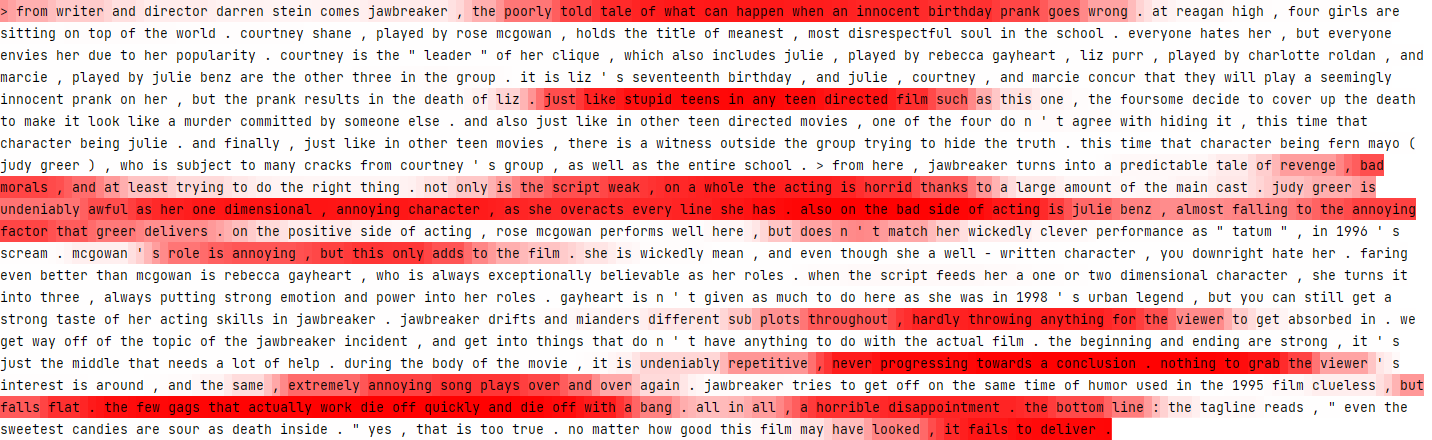}
    \caption{Rationale for a negative movie review.}
\end{figure}

\begin{figure}[H]
    \centering
    \includegraphics[width=0.99\linewidth]{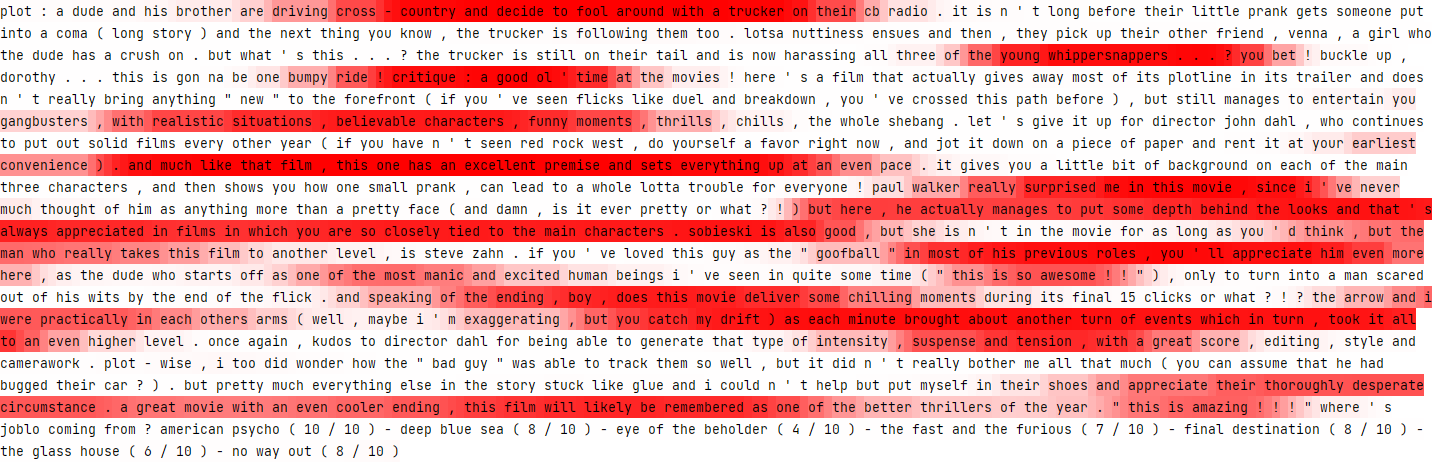}
    \caption{Rationale for a positive movie review.}
\end{figure}
\newpage
\section{ImageNet Mask Comparison}
\label{sec:vis}
\begin{multicols}{2}
\setlength{\parindent}{0pt}
This section includes a more extensive comparison of masks created by different interpretability methods on the selection of images used by \citet{13}. For MaRC, an additional visualization is added to more easily see the unmasked regions in the image. For an overview of methods as well as hyperparameter settings, compare Appendix \ref{sec:exp_detials}.
\end{multicols}
\subsection{ResNet-101}
\begin{figure}[H]
\makebox[0.99\linewidth][c]{
\begin{subfigure}{.163\textwidth}
  \centering
  \caption*{Input}
  \includegraphics[width=0.97\linewidth]{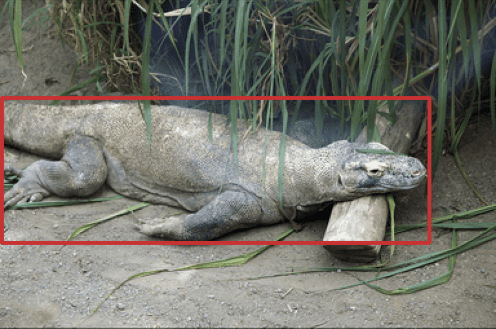}
\end{subfigure}
\begin{subfigure}{.163\textwidth}
  \centering
  \caption*{MaRC}
  \includegraphics[width=0.97\linewidth]{images/1/ResNet_n01695060_5541.JPEG_Optim.jpg}
\end{subfigure}
\begin{subfigure}{.163\textwidth}
  \centering
  \caption*{M-Perturb}
  \includegraphics[width=0.97\linewidth]{images/1/ResNet_n01695060_5541.JPEG_Perturb.jpg}
\end{subfigure}
\begin{subfigure}{.163\textwidth}
  \centering
  \caption*{Grad-CAM}
  \includegraphics[width=0.97\linewidth]{images/1/ResNet_n01695060_5541.JPEG_grad_cam.jpg}
\end{subfigure}
\begin{subfigure}{.163\textwidth}
  \centering
  \caption*{Excitation BP}
  \includegraphics[width=0.97\linewidth]{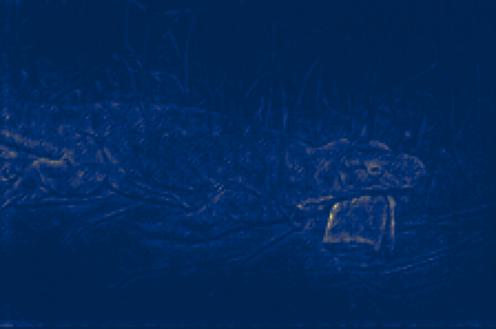}
\end{subfigure}
\begin{subfigure}{.163\textwidth}
  \centering
  \caption*{Occlusion}
  \includegraphics[width=0.97\linewidth]{images/1/ResNet_n01695060_5541.JPEG_Occ.jpg}
\end{subfigure}
}\vspace{2px}\\
\makebox[0.99\linewidth][c]{
\hspace{.163\textwidth}
\begin{subfigure}{.163\textwidth}
  \centering
  \includegraphics[width=0.97\linewidth]{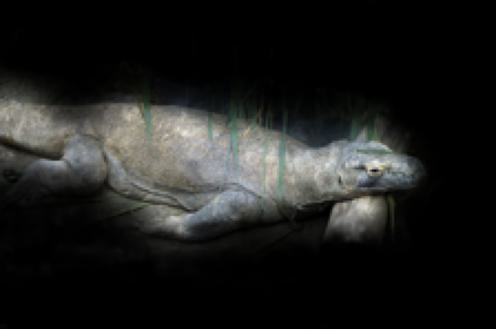}
  \caption*{MaRC Vis.}
\end{subfigure}
\begin{subfigure}{.163\textwidth}
  \centering
  \includegraphics[width=0.97\linewidth]{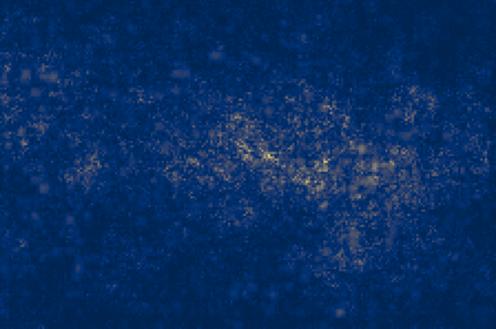}
  \caption*{Saliency}
\end{subfigure}
\begin{subfigure}{.163\textwidth}
  \centering
  \includegraphics[width=0.97\linewidth]{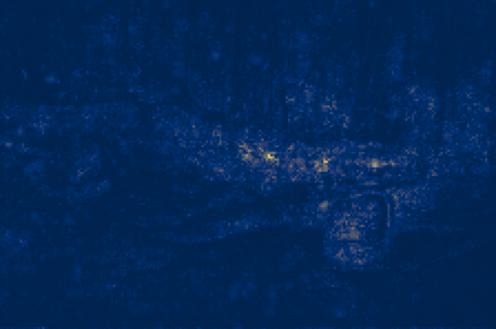}
  \caption*{InputXGrad}
\end{subfigure}
\begin{subfigure}{.163\textwidth}
  \centering
  \includegraphics[width=0.97\linewidth]{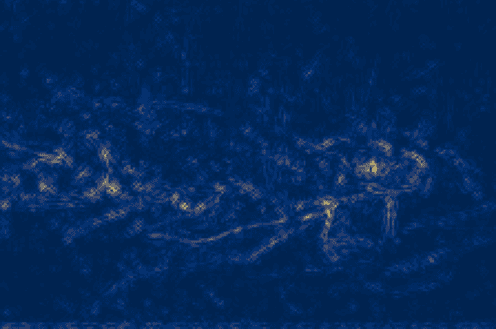}
  \caption*{Guided BP}
\end{subfigure}
\begin{subfigure}{.163\textwidth}
  \centering
  \includegraphics[width=0.97\linewidth]{images/1/ResNet_n01695060_5541.JPEG_IG.jpg}
  \caption*{Integr. Grads}
\end{subfigure}
}
\caption{Komodo dragon}
\end{figure}

\begin{figure}[H]
\makebox[0.99\linewidth][c]{
\begin{subfigure}{.163\textwidth}
  \centering
  \caption*{Input}
  \includegraphics[width=0.97\linewidth]{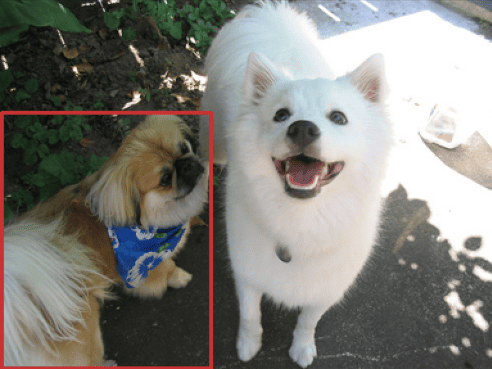}
\end{subfigure}
\begin{subfigure}{.163\textwidth}
  \centering
  \caption*{MaRC}
  \includegraphics[width=0.97\linewidth]{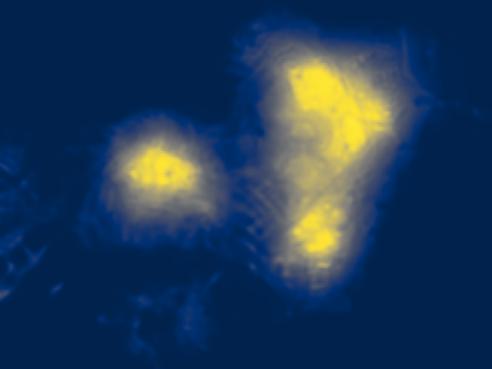}
\end{subfigure}
\begin{subfigure}{.163\textwidth}
  \centering
  \caption*{M-Perturb}
  \includegraphics[width=0.97\linewidth]{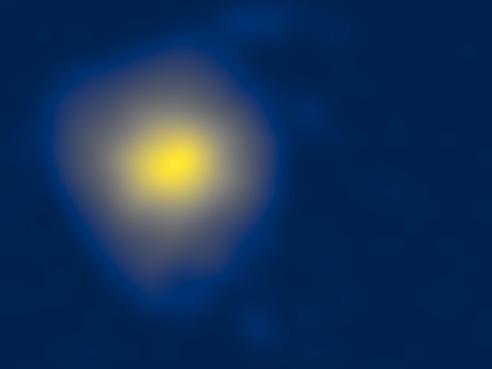}
\end{subfigure}
\begin{subfigure}{.163\textwidth}
  \centering
  \caption*{Grad-CAM}
  \includegraphics[width=0.97\linewidth]{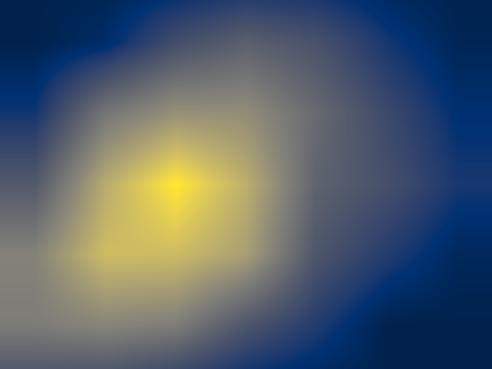}
\end{subfigure}
\begin{subfigure}{.163\textwidth}
  \centering
  \caption*{Excitation BP}
  \includegraphics[width=0.97\linewidth]{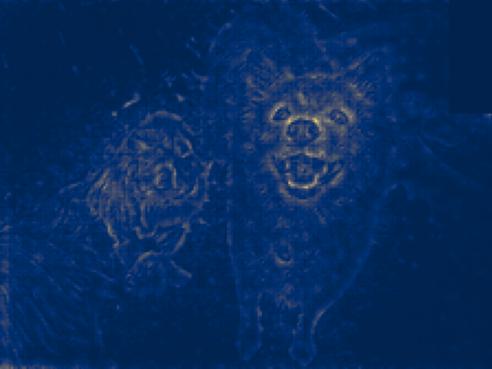}
\end{subfigure}
\begin{subfigure}{.163\textwidth}
  \centering
  \caption*{Occlusion}
  \includegraphics[width=0.97\linewidth]{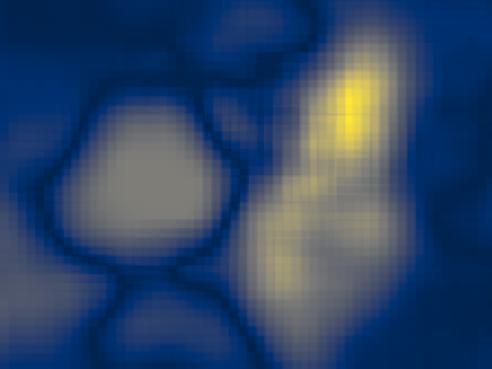}
\end{subfigure}
}\vspace{2px}\\
\makebox[0.99\linewidth][c]{
\hspace{.163\textwidth}
\begin{subfigure}{.163\textwidth}
  \centering
  \includegraphics[width=0.97\linewidth]{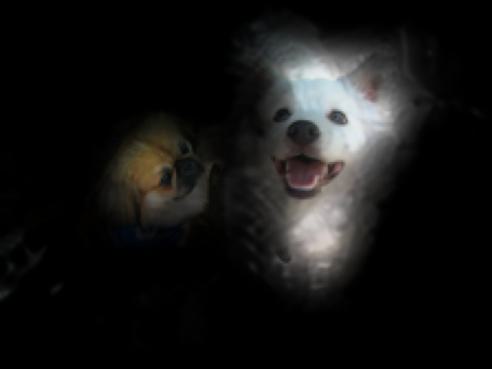}
  \caption*{MaRC Vis.}
\end{subfigure}
\begin{subfigure}{.163\textwidth}
  \centering
  \includegraphics[width=0.97\linewidth]{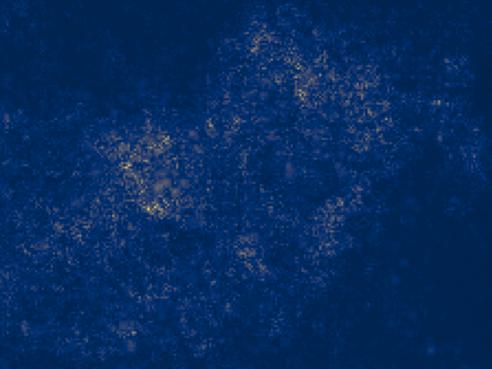}
  \caption*{Saliency}
\end{subfigure}
\begin{subfigure}{.163\textwidth}
  \centering
  \includegraphics[width=0.97\linewidth]{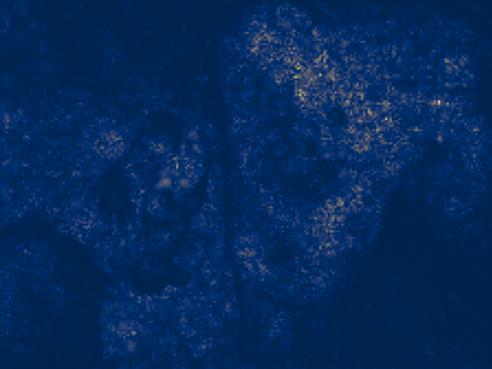}
  \caption*{InputXGrad}
\end{subfigure}
\begin{subfigure}{.163\textwidth}
  \centering
  \includegraphics[width=0.97\linewidth]{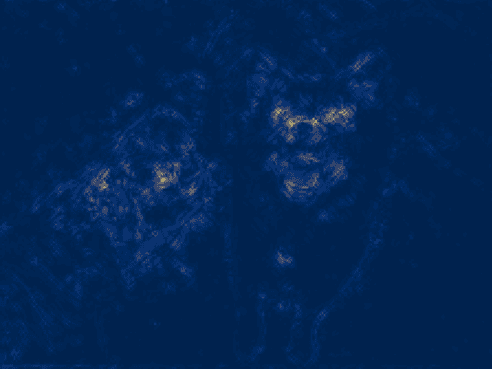}
  \caption*{Guided BP}
\end{subfigure}
\begin{subfigure}{.163\textwidth}
  \centering
  \includegraphics[width=0.97\linewidth]{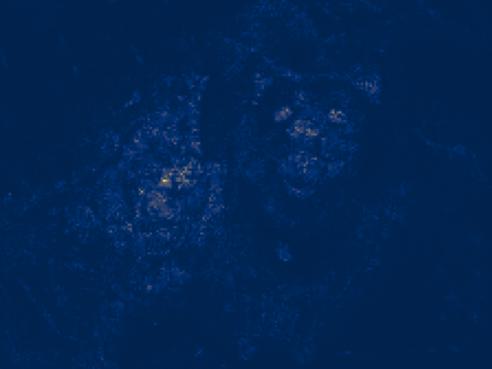}
  \caption*{Integr. Grads}
\end{subfigure}
}
\caption{Pekinese}
\end{figure}

\begin{figure}[H]
\makebox[0.99\linewidth][c]{
\begin{subfigure}{.163\textwidth}
  \centering
  \caption*{Input}
  \includegraphics[width=0.97\linewidth]{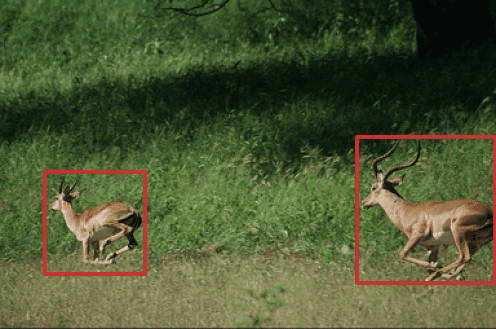}
\end{subfigure}
\begin{subfigure}{.163\textwidth}
  \centering
  \caption*{MaRC}
  \includegraphics[width=0.97\linewidth]{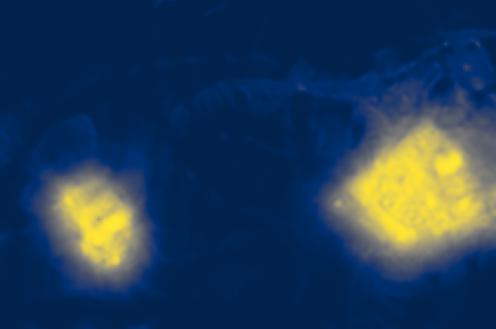}
\end{subfigure}
\begin{subfigure}{.163\textwidth}
  \centering
  \caption*{M-Perturb}
  \includegraphics[width=0.97\linewidth]{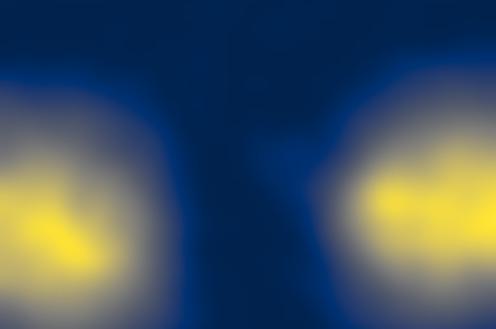}
\end{subfigure}
\begin{subfigure}{.163\textwidth}
  \centering
  \caption*{Grad-CAM}
  \includegraphics[width=0.97\linewidth]{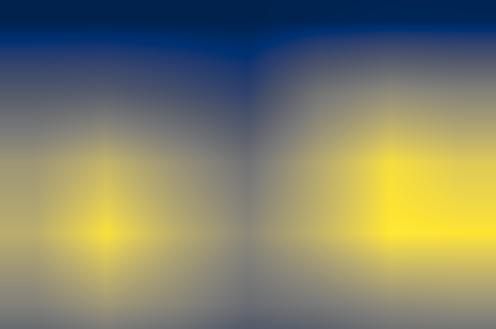}
\end{subfigure}
\begin{subfigure}{.163\textwidth}
  \centering
  \caption*{Excitation BP}
  \includegraphics[width=0.97\linewidth]{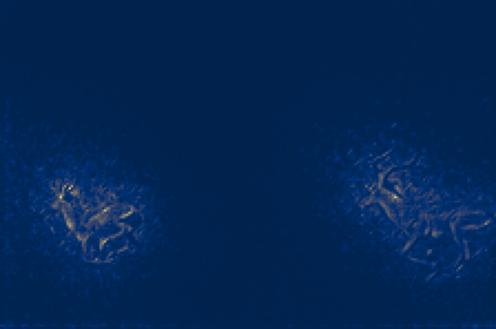}
\end{subfigure}
\begin{subfigure}{.163\textwidth}
  \centering
  \caption*{Occlusion}
  \includegraphics[width=0.97\linewidth]{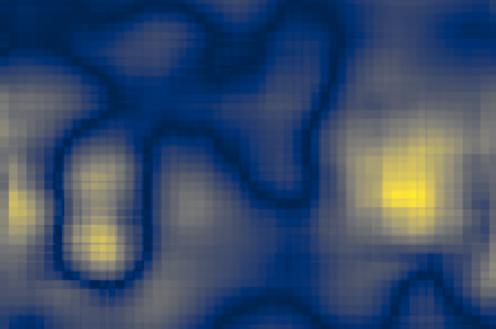}
\end{subfigure}
}\vspace{2px}\\
\makebox[0.99\linewidth][c]{
\hspace{.163\textwidth}
\begin{subfigure}{.163\textwidth}
  \centering
  \includegraphics[width=0.97\linewidth]{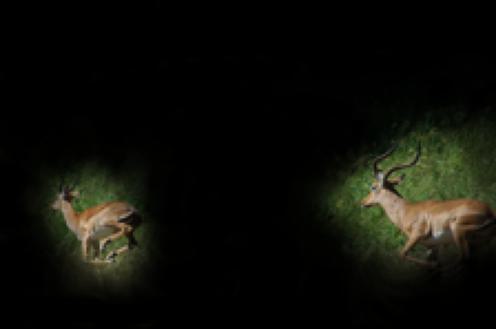}
  \caption*{MaRC Vis.}
\end{subfigure}
\begin{subfigure}{.163\textwidth}
  \centering
  \includegraphics[width=0.97\linewidth]{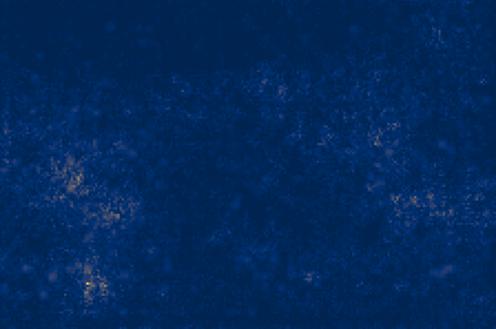}
  \caption*{Saliency}
\end{subfigure}
\begin{subfigure}{.163\textwidth}
  \centering
  \includegraphics[width=0.97\linewidth]{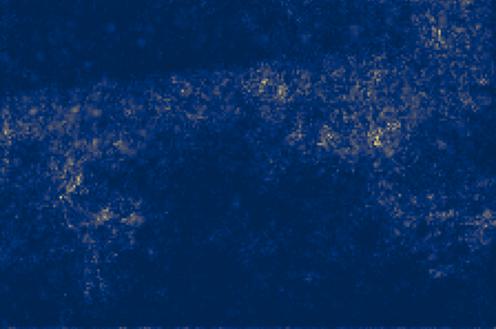}
  \caption*{InputXGrad}
\end{subfigure}
\begin{subfigure}{.163\textwidth}
  \centering
  \includegraphics[width=0.97\linewidth]{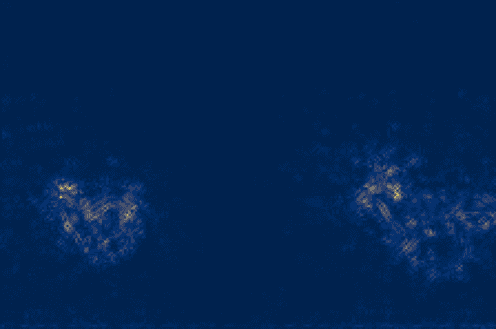}
  \caption*{Guided BP}
\end{subfigure}
\begin{subfigure}{.163\textwidth}
  \centering
  \includegraphics[width=0.97\linewidth]{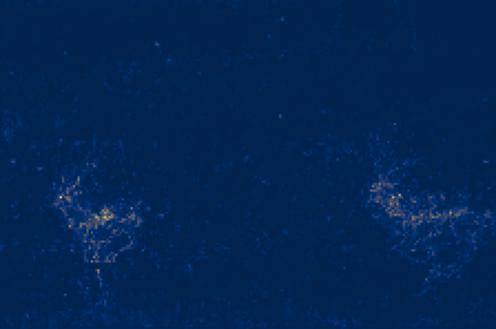}
  \caption*{Integr. Grads}
\end{subfigure}
}
\caption{Impala}
\end{figure}

\begin{figure}[H]
\makebox[0.99\linewidth][c]{
\begin{subfigure}{.163\textwidth}
  \centering
  \caption*{Input}
  \includegraphics[width=0.97\linewidth]{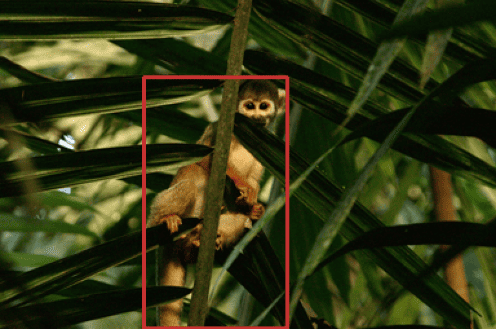}
\end{subfigure}
\begin{subfigure}{.163\textwidth}
  \centering
  \caption*{MaRC}
  \includegraphics[width=0.97\linewidth]{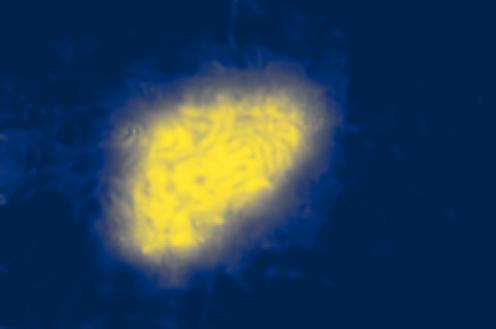}
\end{subfigure}
\begin{subfigure}{.163\textwidth}
  \centering
  \caption*{M-Perturb}
  \includegraphics[width=0.97\linewidth]{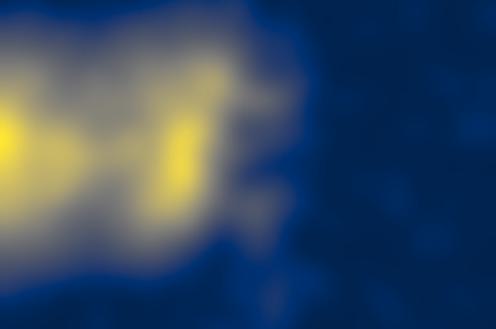}
\end{subfigure}
\begin{subfigure}{.163\textwidth}
  \centering
  \caption*{Grad-CAM}
  \includegraphics[width=0.97\linewidth]{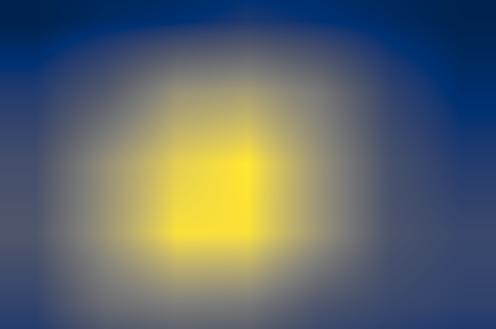}
\end{subfigure}
\begin{subfigure}{.163\textwidth}
  \centering
  \caption*{Excitation BP}
  \includegraphics[width=0.97\linewidth]{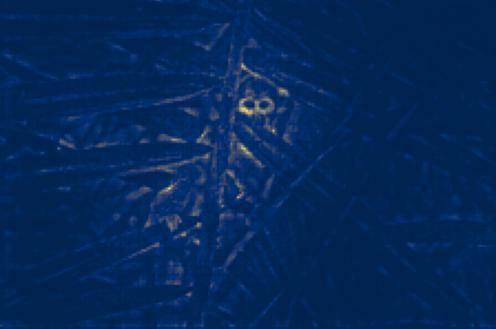}
\end{subfigure}
\begin{subfigure}{.163\textwidth}
  \centering
  \caption*{Occlusion}
  \includegraphics[width=0.97\linewidth]{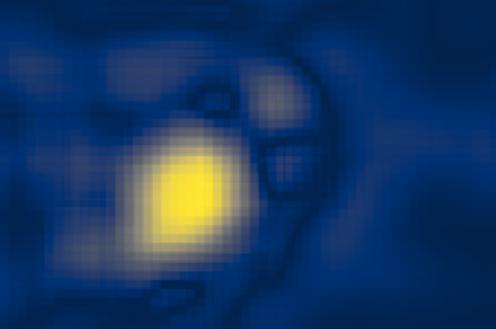}
\end{subfigure}
}\vspace{2px}\\
\makebox[0.99\linewidth][c]{
\hspace{.163\textwidth}
\begin{subfigure}{.163\textwidth}
  \centering
  \includegraphics[width=0.97\linewidth]{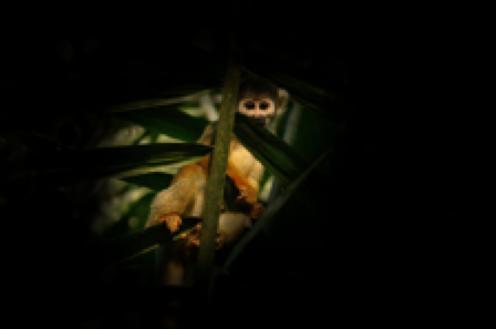}
  \caption*{MaRC Vis.}
\end{subfigure}
\begin{subfigure}{.163\textwidth}
  \centering
  \includegraphics[width=0.97\linewidth]{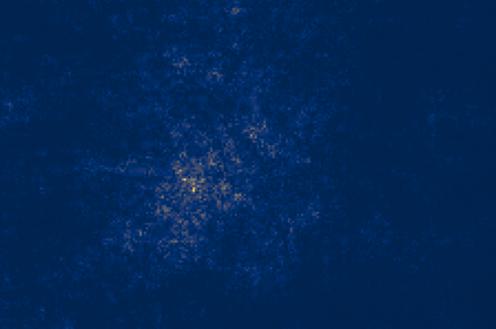}
  \caption*{Saliency}
\end{subfigure}
\begin{subfigure}{.163\textwidth}
  \centering
  \includegraphics[width=0.97\linewidth]{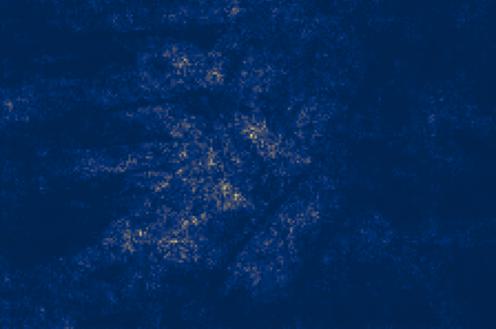}
  \caption*{InputXGrad}
\end{subfigure}
\begin{subfigure}{.163\textwidth}
  \centering
  \includegraphics[width=0.97\linewidth]{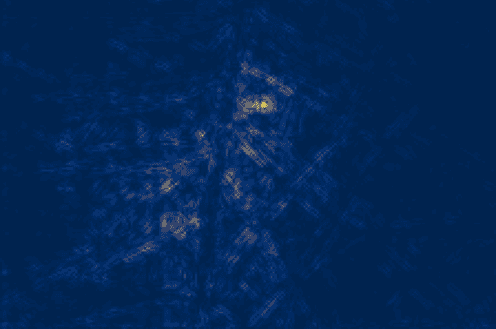}
  \caption*{Guided BP}
\end{subfigure}
\begin{subfigure}{.163\textwidth}
  \centering
  \includegraphics[width=0.97\linewidth]{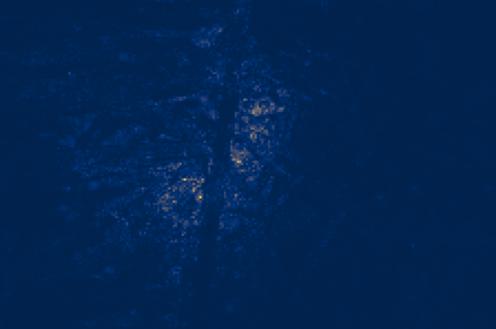}
  \caption*{Integr. Grads}
\end{subfigure}
}
\caption{Squirrel monkey}
\end{figure}

\begin{figure}[H]
\makebox[0.99\linewidth][c]{
\begin{subfigure}{.163\textwidth}
  \centering
  \caption*{Input}
  \includegraphics[width=0.97\linewidth]{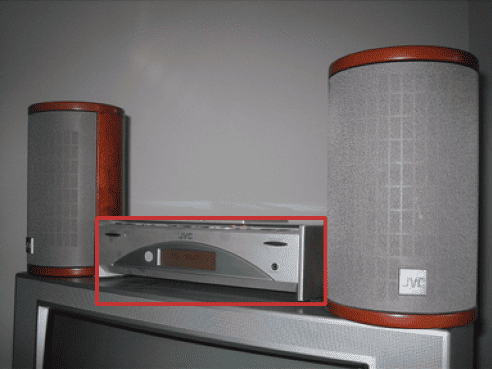}
\end{subfigure}
\begin{subfigure}{.163\textwidth}
  \centering
  \caption*{MaRC}
  \includegraphics[width=0.97\linewidth]{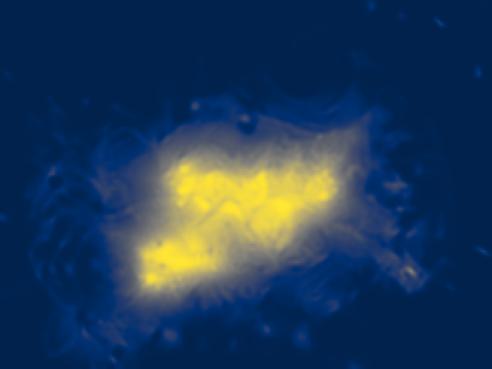}
\end{subfigure}
\begin{subfigure}{.163\textwidth}
  \centering
  \caption*{M-Perturb}
  \includegraphics[width=0.97\linewidth]{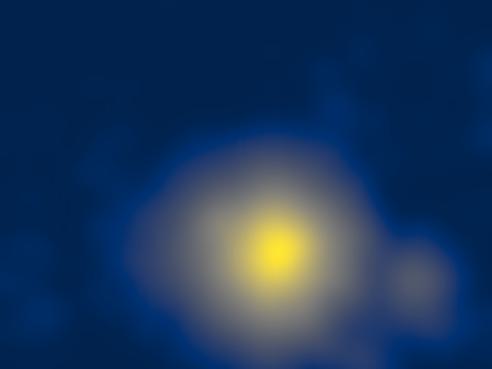}
\end{subfigure}
\begin{subfigure}{.163\textwidth}
  \centering
  \caption*{Grad-CAM}
  \includegraphics[width=0.97\linewidth]{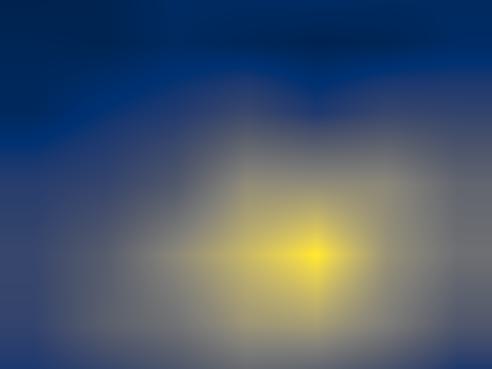}
\end{subfigure}
\begin{subfigure}{.163\textwidth}
  \centering
  \caption*{Excitation BP}
  \includegraphics[width=0.97\linewidth]{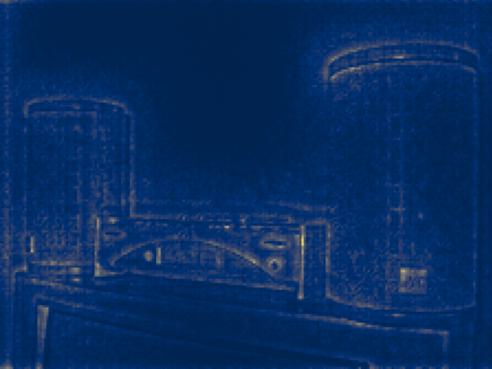}
\end{subfigure}
\begin{subfigure}{.163\textwidth}
  \centering
  \caption*{Occlusion}
  \includegraphics[width=0.97\linewidth]{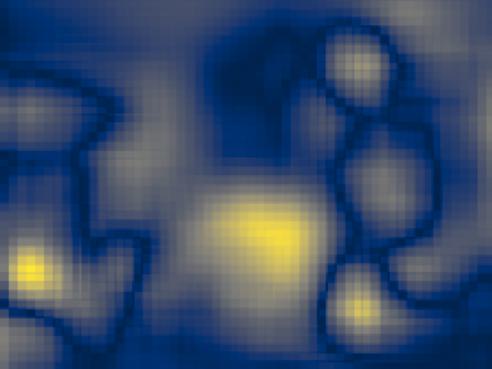}
\end{subfigure}
}\vspace{2px}\\
\makebox[0.99\linewidth][c]{
\hspace{.163\textwidth}
\begin{subfigure}{.163\textwidth}
  \centering
  \includegraphics[width=0.97\linewidth]{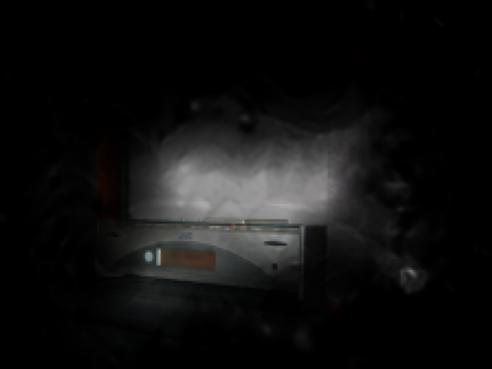}
  \caption*{MaRC Vis.}
\end{subfigure}
\begin{subfigure}{.163\textwidth}
  \centering
  \includegraphics[width=0.97\linewidth]{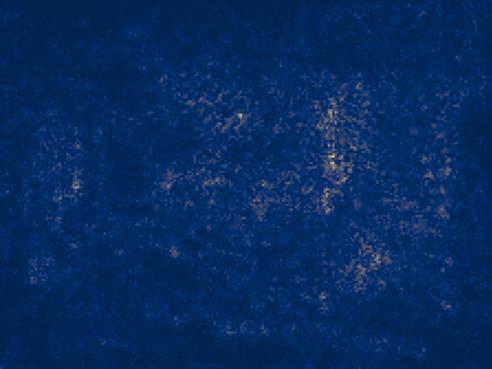}
  \caption*{Saliency}
\end{subfigure}
\begin{subfigure}{.163\textwidth}
  \centering
  \includegraphics[width=0.97\linewidth]{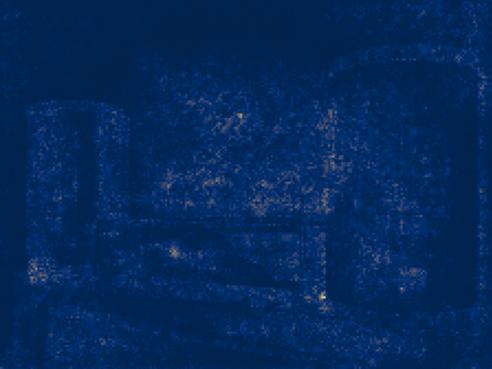}
  \caption*{InputXGrad}
\end{subfigure}
\begin{subfigure}{.163\textwidth}
  \centering
  \includegraphics[width=0.97\linewidth]{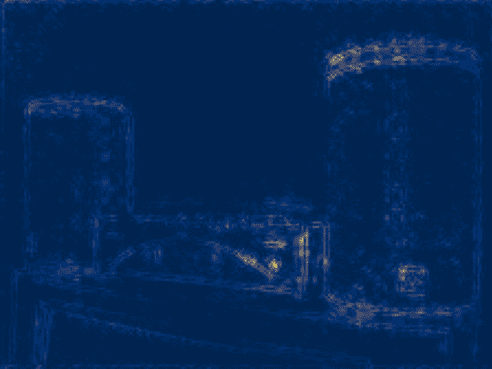}
  \caption*{Guided BP}
\end{subfigure}
\begin{subfigure}{.163\textwidth}
  \centering
  \includegraphics[width=0.97\linewidth]{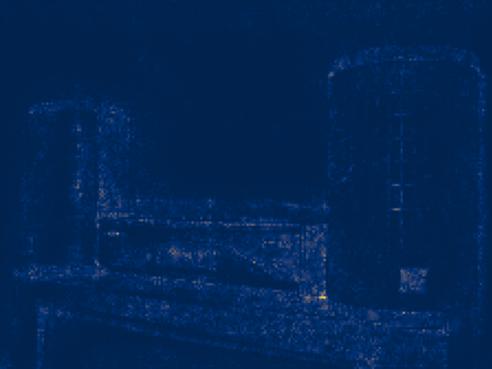}
  \caption*{Integr. Grads}
\end{subfigure}
}
\caption{CD player}
\end{figure}

\begin{figure}[H]
\makebox[0.99\linewidth][c]{
\begin{subfigure}{.163\textwidth}
  \centering
  \caption*{Input}
  \includegraphics[width=0.97\linewidth]{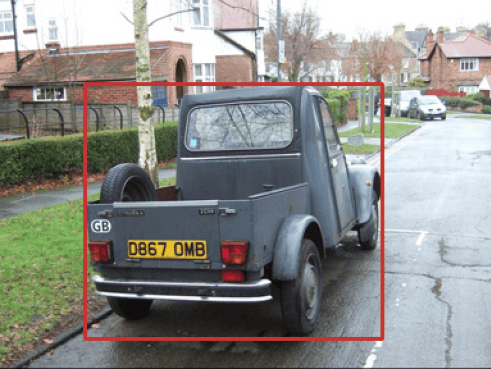}
\end{subfigure}
\begin{subfigure}{.163\textwidth}
  \centering
  \caption*{MaRC}
  \includegraphics[width=0.97\linewidth]{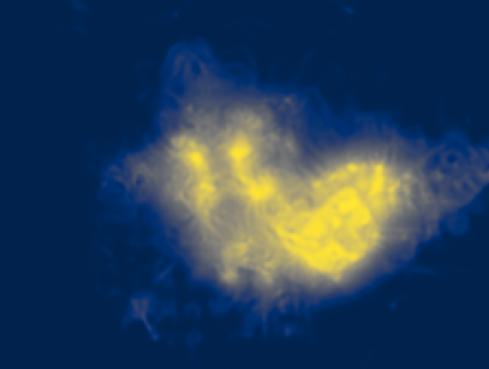}
\end{subfigure}
\begin{subfigure}{.163\textwidth}
  \centering
  \caption*{M-Perturb}
  \includegraphics[width=0.97\linewidth]{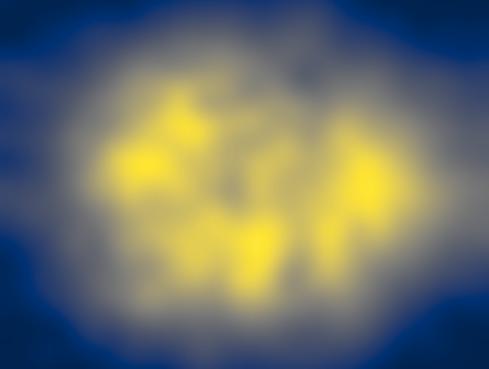}
\end{subfigure}
\begin{subfigure}{.163\textwidth}
  \centering
  \caption*{Grad-CAM}
  \includegraphics[width=0.97\linewidth]{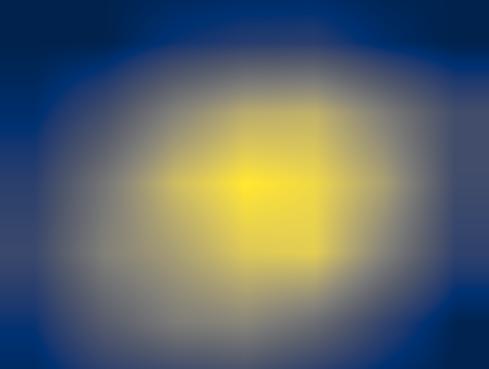}
\end{subfigure}
\begin{subfigure}{.163\textwidth}
  \centering
  \caption*{Excitation BP}
  \includegraphics[width=0.97\linewidth]{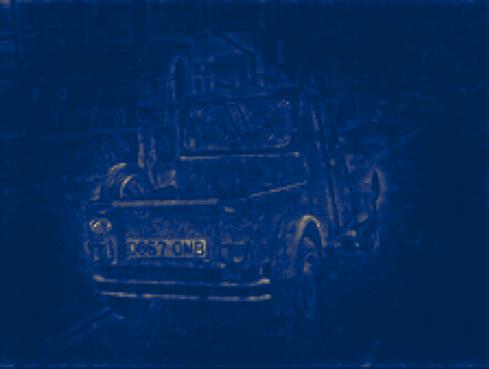}
\end{subfigure}
\begin{subfigure}{.163\textwidth}
  \centering
  \caption*{Occlusion}
  \includegraphics[width=0.97\linewidth]{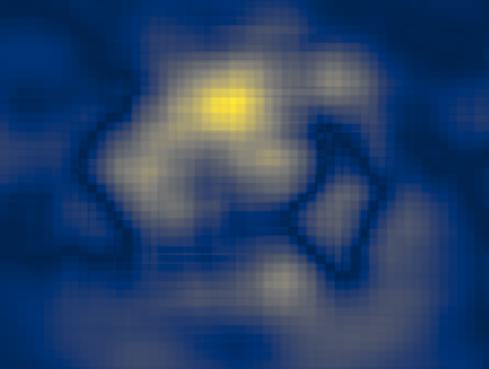}
\end{subfigure}
}\vspace{2px}\\
\makebox[0.99\linewidth][c]{
\hspace{.163\textwidth}
\begin{subfigure}{.163\textwidth}
  \centering
  \includegraphics[width=0.97\linewidth]{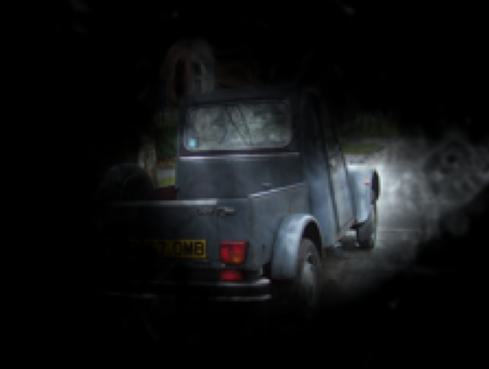}
  \caption*{MaRC Vis.}
\end{subfigure}
\begin{subfigure}{.163\textwidth}
  \centering
  \includegraphics[width=0.97\linewidth]{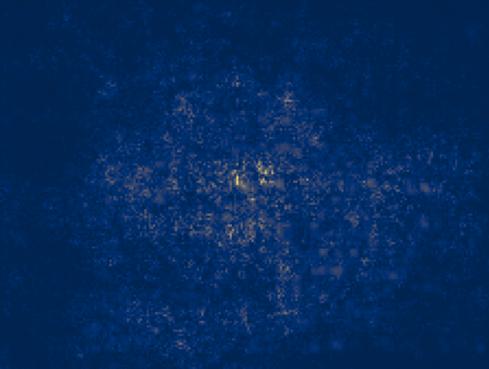}
  \caption*{Saliency}
\end{subfigure}
\begin{subfigure}{.163\textwidth}
  \centering
  \includegraphics[width=0.97\linewidth]{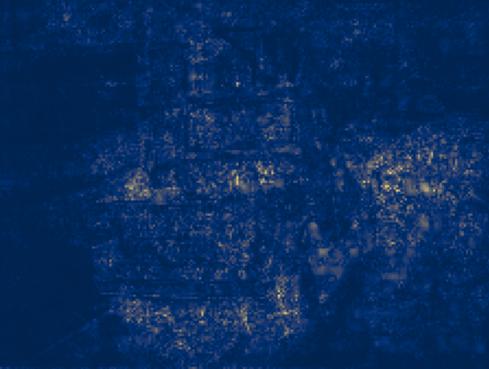}
  \caption*{InputXGrad}
\end{subfigure}
\begin{subfigure}{.163\textwidth}
  \centering
  \includegraphics[width=0.97\linewidth]{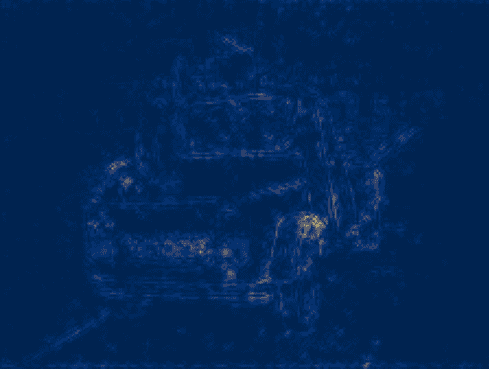}
  \caption*{Guided BP}
\end{subfigure}
\begin{subfigure}{.163\textwidth}
  \centering
  \includegraphics[width=0.97\linewidth]{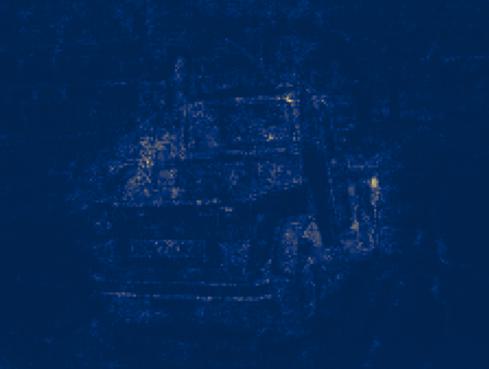}
  \caption*{Integr. Grads}
\end{subfigure}
}
\caption{Pickup}
\end{figure}

\begin{figure}[H]
\makebox[0.99\linewidth][c]{
\begin{subfigure}{.163\textwidth}
  \centering
  \caption*{Input}
  \includegraphics[width=0.97\linewidth]{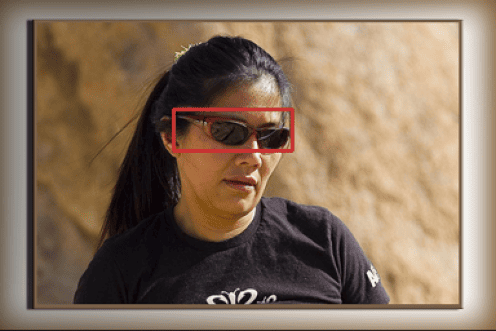}
\end{subfigure}
\begin{subfigure}{.163\textwidth}
  \centering
  \caption*{MaRC}
  \includegraphics[width=0.97\linewidth]{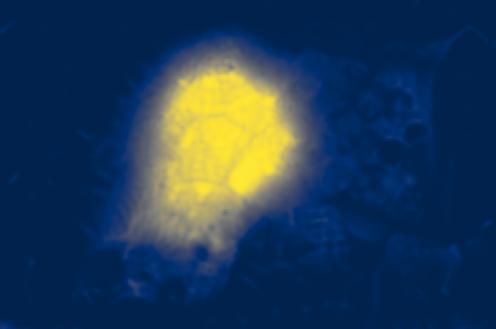}
\end{subfigure}
\begin{subfigure}{.163\textwidth}
  \centering
  \caption*{M-Perturb}
  \includegraphics[width=0.97\linewidth]{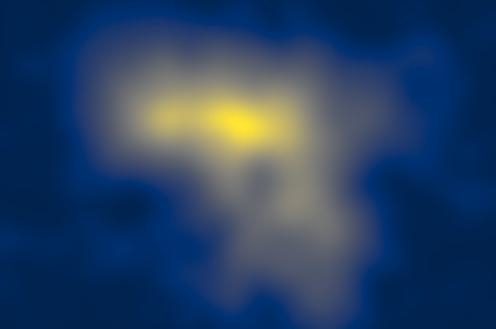}
\end{subfigure}
\begin{subfigure}{.163\textwidth}
  \centering
  \caption*{Grad-CAM}
  \includegraphics[width=0.97\linewidth]{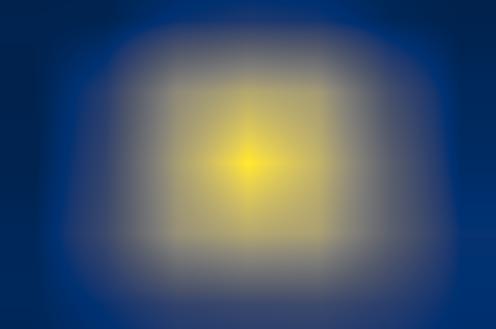}
\end{subfigure}
\begin{subfigure}{.163\textwidth}
  \centering
  \caption*{Excitation BP}
  \includegraphics[width=0.97\linewidth]{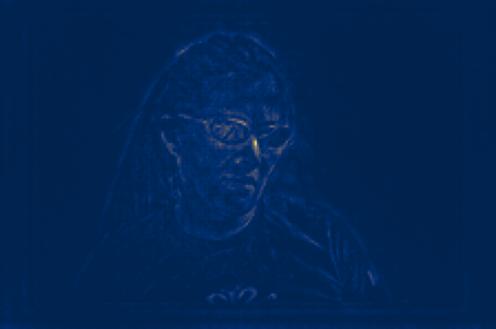}
\end{subfigure}
\begin{subfigure}{.163\textwidth}
  \centering
  \caption*{Occlusion}
  \includegraphics[width=0.97\linewidth]{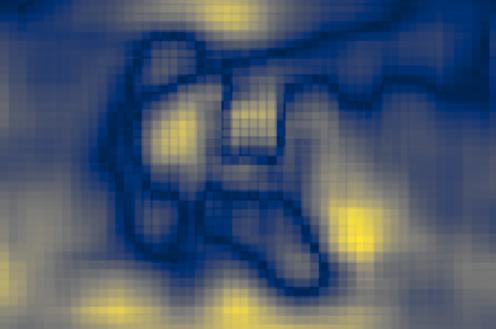}
\end{subfigure}
}\vspace{2px}\\
\makebox[0.99\linewidth][c]{
\hspace{.163\textwidth}
\begin{subfigure}{.163\textwidth}
  \centering
  \includegraphics[width=0.97\linewidth]{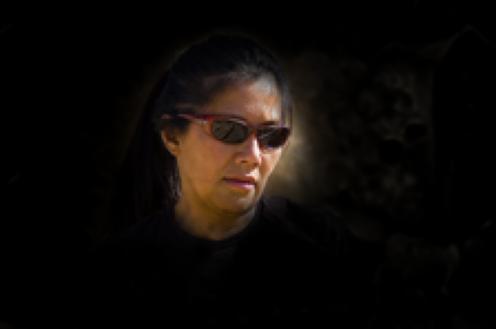}
  \caption*{MaRC Vis.}
\end{subfigure}
\begin{subfigure}{.163\textwidth}
  \centering
  \includegraphics[width=0.97\linewidth]{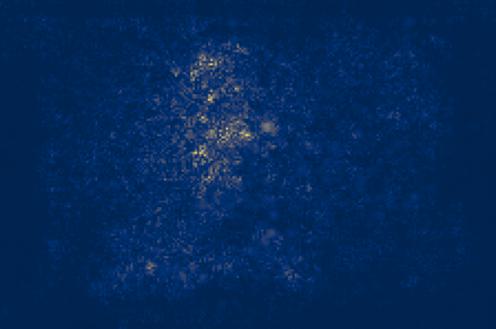}
  \caption*{Saliency}
\end{subfigure}
\begin{subfigure}{.163\textwidth}
  \centering
  \includegraphics[width=0.97\linewidth]{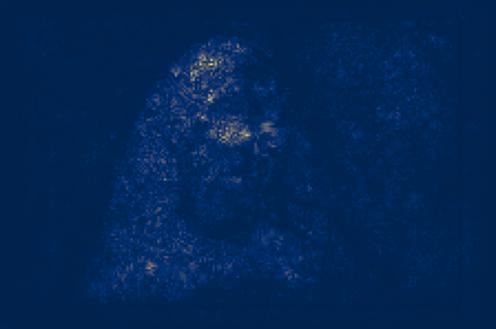}
  \caption*{InputXGrad}
\end{subfigure}
\begin{subfigure}{.163\textwidth}
  \centering
  \includegraphics[width=0.97\linewidth]{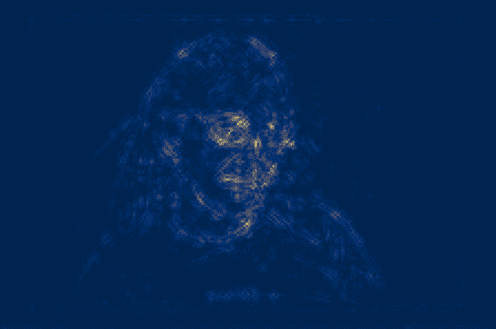}
  \caption*{Guided BP}
\end{subfigure}
\begin{subfigure}{.163\textwidth}
  \centering
  \includegraphics[width=0.97\linewidth]{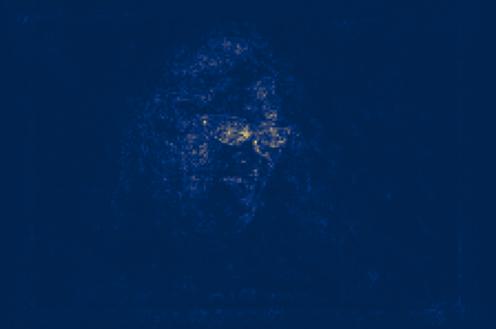}
  \caption*{Integr. Grads}
\end{subfigure}
}
\caption{Sunglasses}

\end{figure}

\begin{figure}[H]
\makebox[0.99\linewidth][c]{
\begin{subfigure}{.163\textwidth}
  \centering
  \caption*{Input}
  \includegraphics[width=0.97\linewidth]{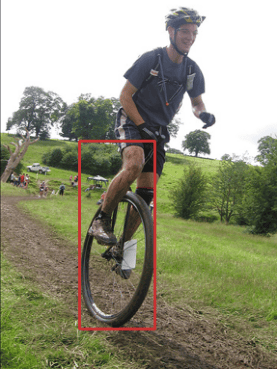}
\end{subfigure}
\begin{subfigure}{.163\textwidth}
  \centering
  \caption*{MaRC}
  \includegraphics[width=0.97\linewidth]{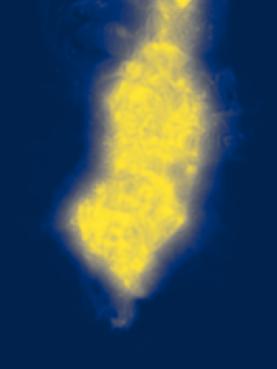}
\end{subfigure}
\begin{subfigure}{.163\textwidth}
  \centering
  \caption*{M-Perturb}
  \includegraphics[width=0.97\linewidth]{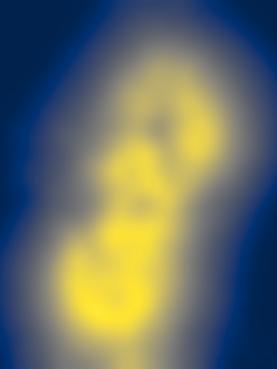}
\end{subfigure}
\begin{subfigure}{.163\textwidth}
  \centering
  \caption*{Grad-CAM}
  \includegraphics[width=0.97\linewidth]{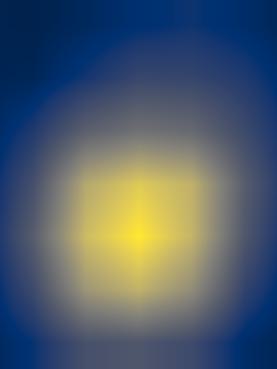}
\end{subfigure}
\begin{subfigure}{.163\textwidth}
  \centering
  \caption*{Excitation BP}
  \includegraphics[width=0.97\linewidth]{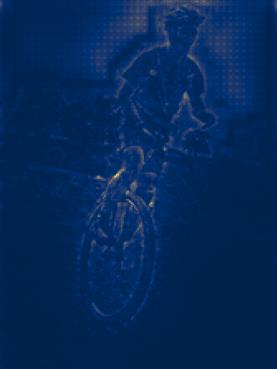}
\end{subfigure}
\begin{subfigure}{.163\textwidth}
  \centering
  \caption*{Occlusion}
  \includegraphics[width=0.97\linewidth]{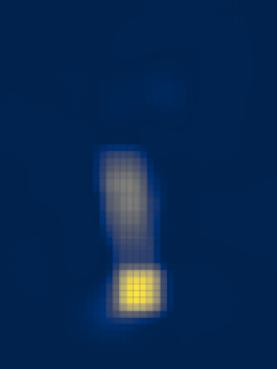}
\end{subfigure}
}\vspace{2px}\\
\makebox[0.99\linewidth][c]{
\hspace{.163\textwidth}
\begin{subfigure}{.163\textwidth}
  \centering
  \includegraphics[width=0.97\linewidth]{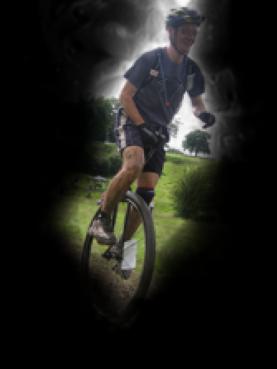}
  \caption*{MaRC Vis.}
\end{subfigure}
\begin{subfigure}{.163\textwidth}
  \centering
  \includegraphics[width=0.97\linewidth]{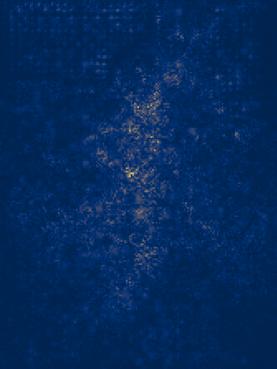}
  \caption*{Saliency}
\end{subfigure}
\begin{subfigure}{.163\textwidth}
  \centering
  \includegraphics[width=0.97\linewidth]{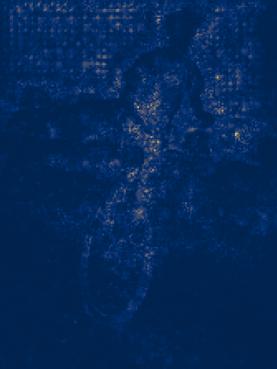}
  \caption*{InputXGrad}
\end{subfigure}
\begin{subfigure}{.163\textwidth}
  \centering
  \includegraphics[width=0.97\linewidth]{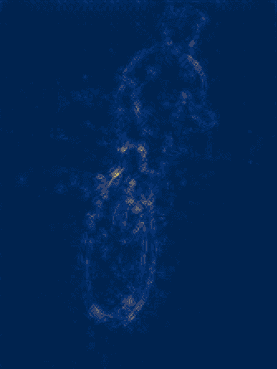}
  \caption*{Guided BP}
\end{subfigure}
\begin{subfigure}{.163\textwidth}
  \centering
  \includegraphics[width=0.97\linewidth]{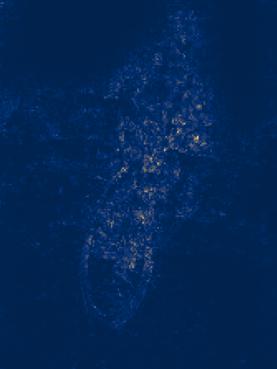}
  \caption*{Integr. Grads}
\end{subfigure}
}
\caption{Unicycle}
\end{figure}

\begin{figure}[H]
\makebox[0.99\linewidth][c]{
\begin{subfigure}{.163\textwidth}
  \centering
  \caption*{Input}
  \includegraphics[width=0.97\linewidth]{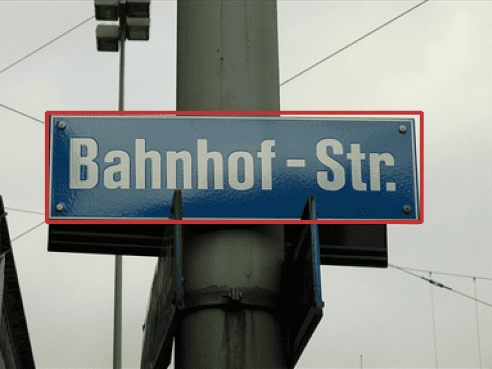}
\end{subfigure}
\begin{subfigure}{.163\textwidth}
  \centering
  \caption*{MaRC}
  \includegraphics[width=0.97\linewidth]{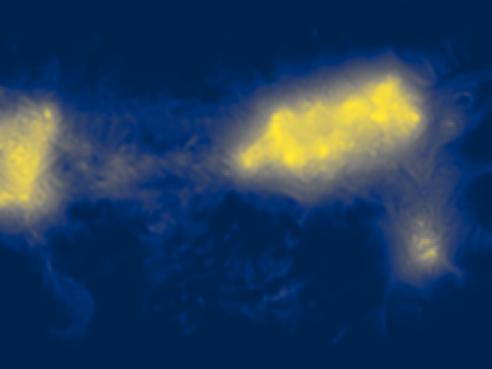}
\end{subfigure}
\begin{subfigure}{.163\textwidth}
  \centering
  \caption*{M-Perturb}
  \includegraphics[width=0.97\linewidth]{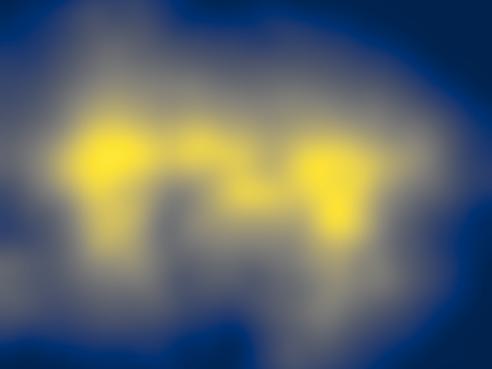}
\end{subfigure}
\begin{subfigure}{.163\textwidth}
  \centering
  \caption*{Grad-CAM}
  \includegraphics[width=0.97\linewidth]{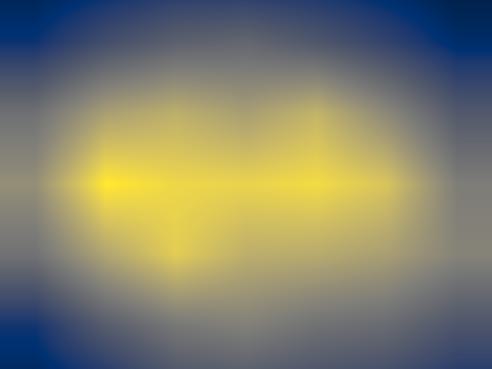}
\end{subfigure}
\begin{subfigure}{.163\textwidth}
  \centering
  \caption*{Excitation BP}
  \includegraphics[width=0.97\linewidth]{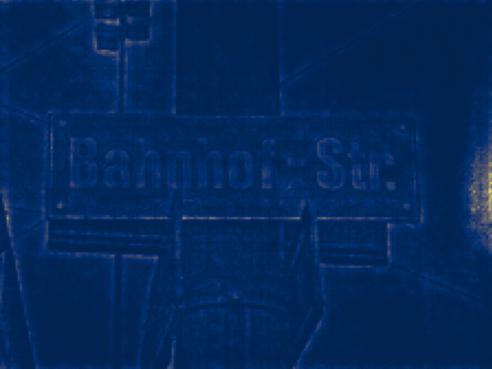}
\end{subfigure}
\begin{subfigure}{.163\textwidth}
  \centering
  \caption*{Occlusion}
  \includegraphics[width=0.97\linewidth]{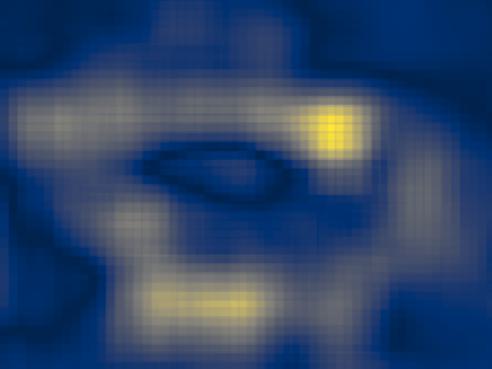}
\end{subfigure}
}\vspace{2px}\\
\makebox[0.99\linewidth][c]{
\hspace{.163\textwidth}
\begin{subfigure}{.163\textwidth}
  \centering
  \includegraphics[width=0.97\linewidth]{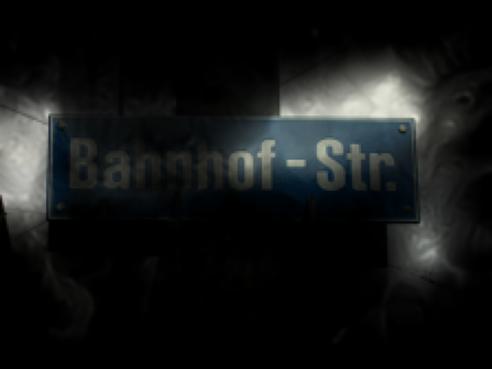}
  \caption*{MaRC Vis.}
\end{subfigure}
\begin{subfigure}{.163\textwidth}
  \centering
  \includegraphics[width=0.97\linewidth]{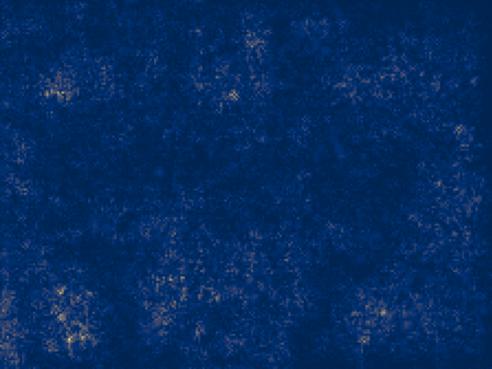}
  \caption*{Saliency}
\end{subfigure}
\begin{subfigure}{.163\textwidth}
  \centering
  \includegraphics[width=0.97\linewidth]{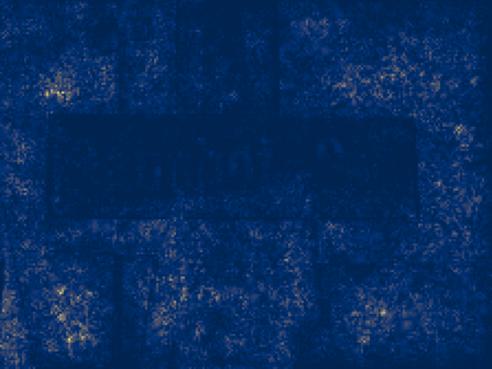}
  \caption*{InputXGrad}
\end{subfigure}
\begin{subfigure}{.163\textwidth}
  \centering
  \includegraphics[width=0.97\linewidth]{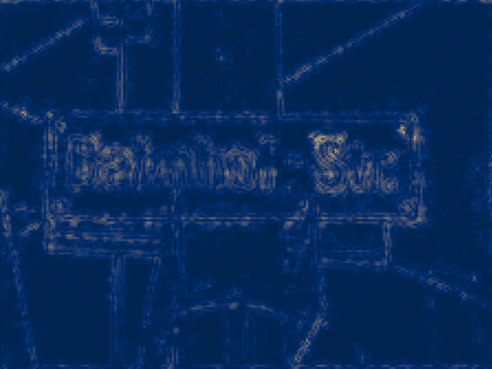}
  \caption*{Guided BP}
\end{subfigure}
\begin{subfigure}{.163\textwidth}
  \centering
  \includegraphics[width=0.97\linewidth]{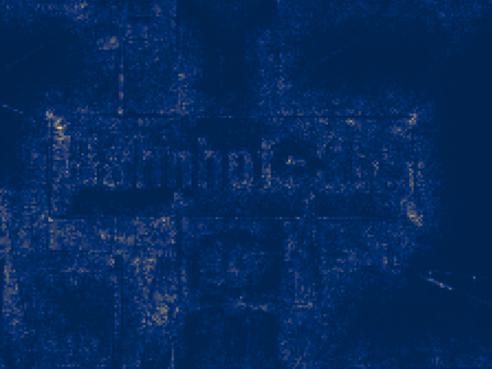}
  \caption*{Integr. Grads}
\end{subfigure}
}
\caption{Street sign}
\end{figure}

\begin{figure}[H]
\makebox[0.99\linewidth][c]{
\begin{subfigure}{.163\textwidth}
  \centering
  \caption*{Input}
  \includegraphics[width=0.97\linewidth]{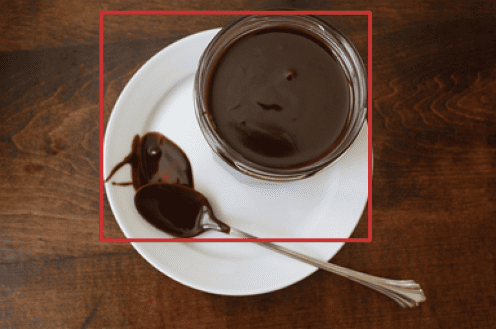}
\end{subfigure}
\begin{subfigure}{.163\textwidth}
  \centering
  \caption*{MaRC}
  \includegraphics[width=0.97\linewidth]{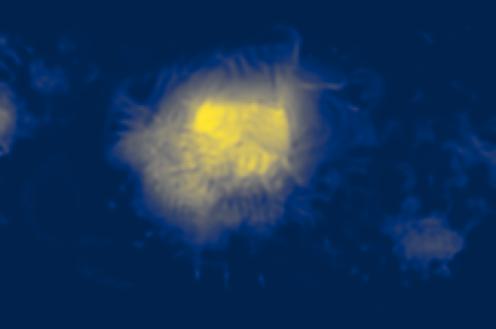}
\end{subfigure}
\begin{subfigure}{.163\textwidth}
  \centering
  \caption*{M-Perturb}
  \includegraphics[width=0.97\linewidth]{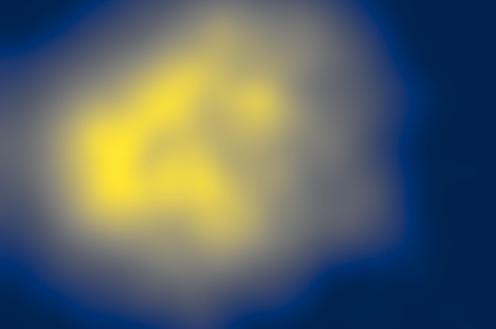}
\end{subfigure}
\begin{subfigure}{.163\textwidth}
  \centering
  \caption*{Grad-CAM}
  \includegraphics[width=0.97\linewidth]{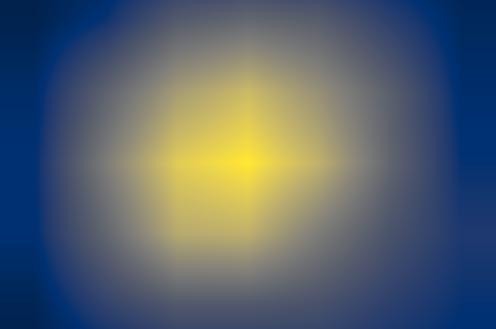}
\end{subfigure}
\begin{subfigure}{.163\textwidth}
  \centering
  \caption*{Excitation BP}
  \includegraphics[width=0.97\linewidth]{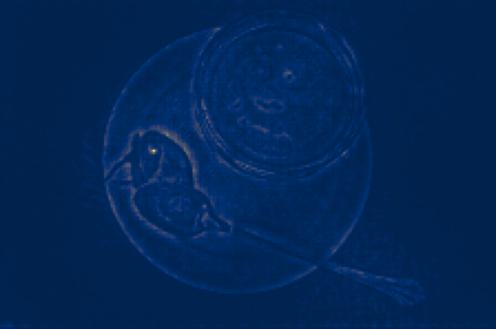}
\end{subfigure}
\begin{subfigure}{.163\textwidth}
  \centering
  \caption*{Occlusion}
  \includegraphics[width=0.97\linewidth]{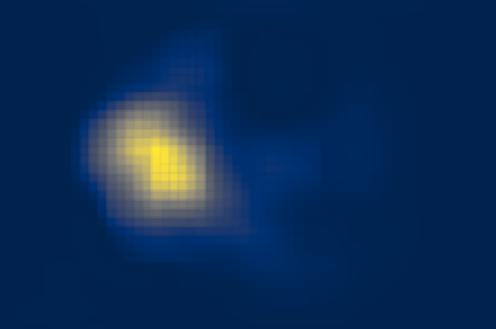}
\end{subfigure}
}\vspace{2px}\\
\makebox[0.99\linewidth][c]{
\hspace{.163\textwidth}
\begin{subfigure}{.163\textwidth}
  \centering
  \includegraphics[width=0.97\linewidth]{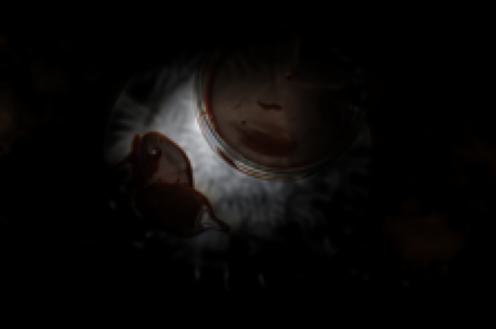}
  \caption*{MaRC Vis.}
\end{subfigure}
\begin{subfigure}{.163\textwidth}
  \centering
  \includegraphics[width=0.97\linewidth]{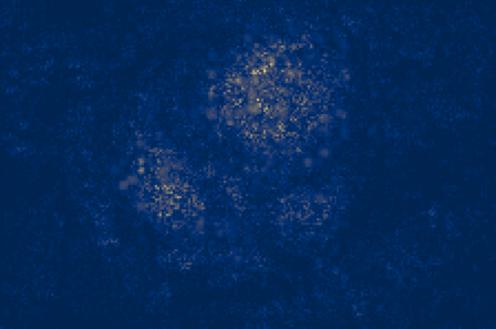}
  \caption*{Saliency}
\end{subfigure}
\begin{subfigure}{.163\textwidth}
  \centering
  \includegraphics[width=0.97\linewidth]{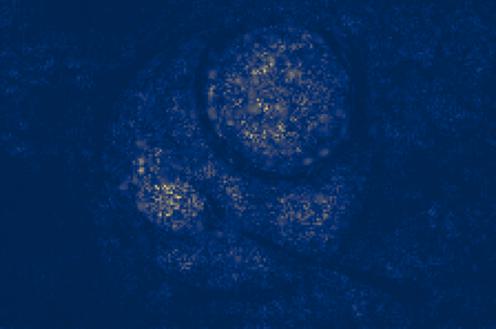}
  \caption*{InputXGrad}
\end{subfigure}
\begin{subfigure}{.163\textwidth}
  \centering
  \includegraphics[width=0.97\linewidth]{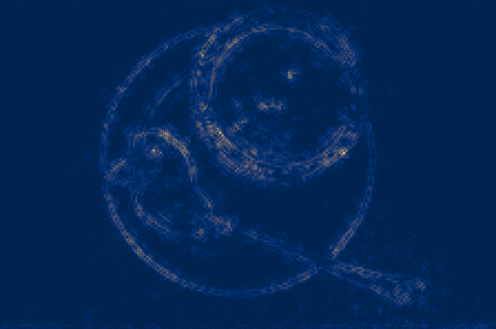}
  \caption*{Guided BP}
\end{subfigure}
\begin{subfigure}{.163\textwidth}
  \centering
  \includegraphics[width=0.97\linewidth]{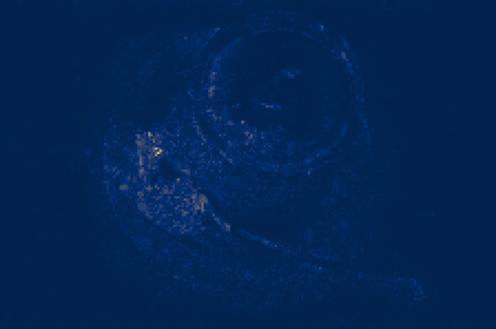}
  \caption*{Integr. Grads}
\end{subfigure}
}
\caption{Chocolate sauce}
\end{figure}

\begin{figure}[H]
\makebox[0.99\linewidth][c]{
\begin{subfigure}{.163\textwidth}
  \centering
  \caption*{Input}
  \includegraphics[width=0.97\linewidth]{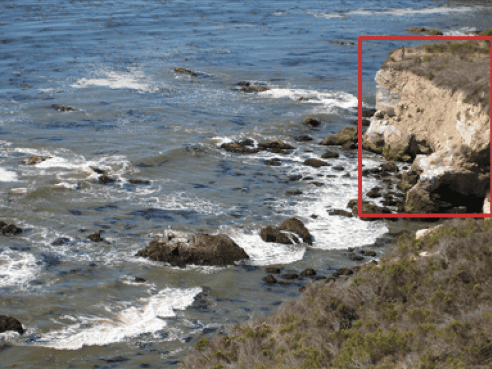}
\end{subfigure}
\begin{subfigure}{.163\textwidth}
  \centering
  \caption*{MaRC}
  \includegraphics[width=0.97\linewidth]{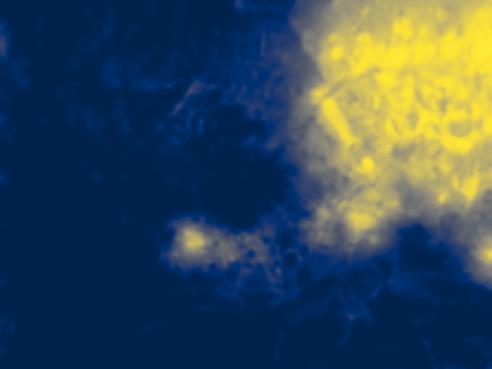}
\end{subfigure}
\begin{subfigure}{.163\textwidth}
  \centering
  \caption*{M-Perturb}
  \includegraphics[width=0.97\linewidth]{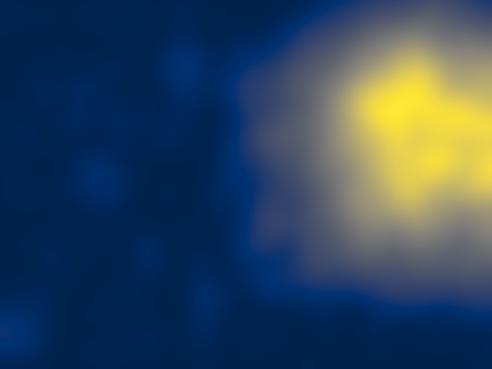}
\end{subfigure}
\begin{subfigure}{.163\textwidth}
  \centering
  \caption*{Grad-CAM}
  \includegraphics[width=0.97\linewidth]{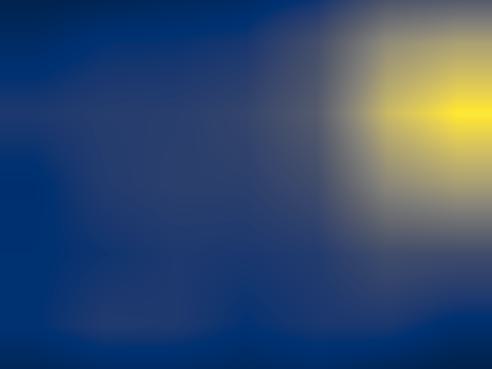}
\end{subfigure}
\begin{subfigure}{.163\textwidth}
  \centering
  \caption*{Excitation BP}
  \includegraphics[width=0.97\linewidth]{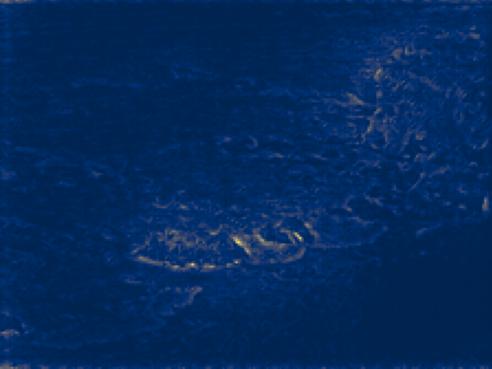}
\end{subfigure}
\begin{subfigure}{.163\textwidth}
  \centering
  \caption*{Occlusion}
  \includegraphics[width=0.97\linewidth]{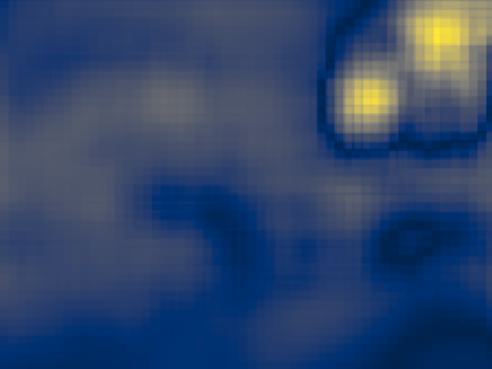}
\end{subfigure}
}\vspace{2px}\\
\makebox[0.99\linewidth][c]{
\hspace{.163\textwidth}
\begin{subfigure}{.163\textwidth}
  \centering
  \includegraphics[width=0.97\linewidth]{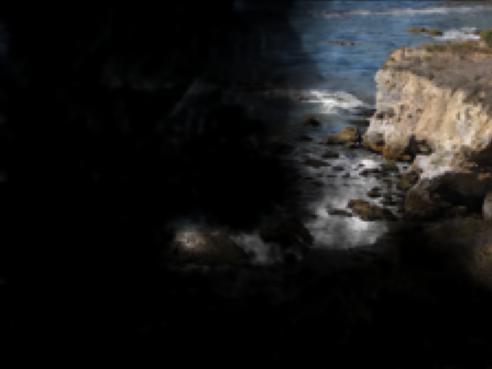}
  \caption*{MaRC Vis.}
\end{subfigure}
\begin{subfigure}{.163\textwidth}
  \centering
  \includegraphics[width=0.97\linewidth]{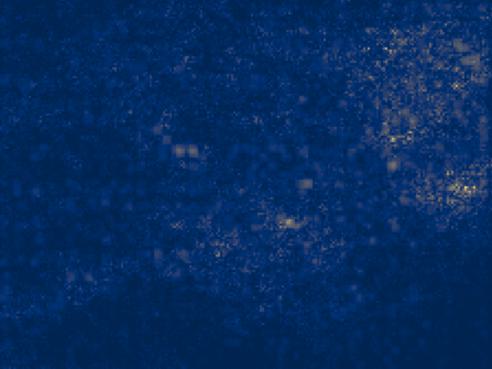}
  \caption*{Saliency}
\end{subfigure}
\begin{subfigure}{.163\textwidth}
  \centering
  \includegraphics[width=0.97\linewidth]{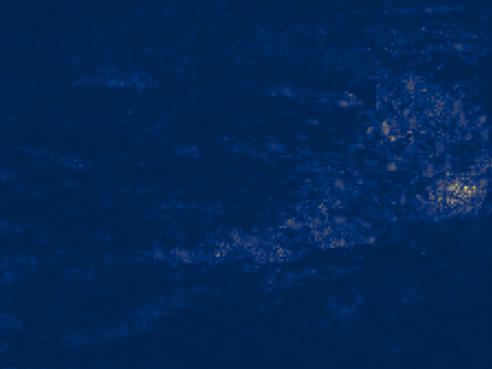}
  \caption*{InputXGrad}
\end{subfigure}
\begin{subfigure}{.163\textwidth}
  \centering
  \includegraphics[width=0.97\linewidth]{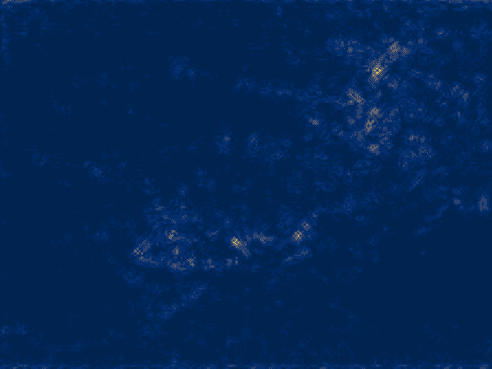}
  \caption*{Guided BP}
\end{subfigure}
\begin{subfigure}{.163\textwidth}
  \centering
  \includegraphics[width=0.97\linewidth]{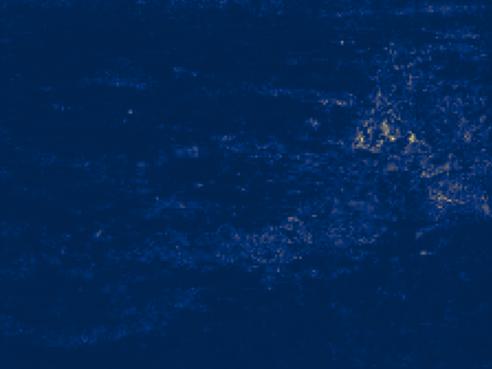}
  \caption*{Integr. Grads}
\end{subfigure}
}
\caption{Cliff}
\end{figure}

\subsection{ViT-B/16}
\begin{figure}[H]
\makebox[0.99\linewidth][c]{
\begin{subfigure}{.139\textwidth}
  \centering
  \caption*{Input}
  \includegraphics[width=0.97\linewidth]{images/gt/0.png}
\end{subfigure}
\begin{subfigure}{.139\textwidth}
  \centering
  \caption*{MaRC}
  \includegraphics[width=0.97\linewidth]{images/1/vit_n01695060_5541.JPEG_Optim.jpg}
\end{subfigure}
\begin{subfigure}{.139\textwidth}
  \centering
  \caption*{M-Perturb}
  \includegraphics[width=0.97\linewidth]{images/1/vit_n01695060_5541.JPEG_Perturb.jpg}
\end{subfigure}
\begin{subfigure}{.139\textwidth}
  \centering
  \caption*{Raw attention}
  \includegraphics[width=0.97\linewidth]{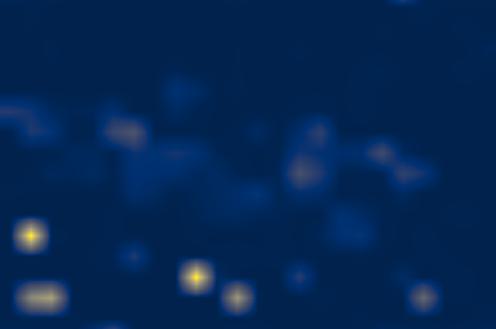}
\end{subfigure}
\begin{subfigure}{.139\textwidth}
  \centering
  \caption*{Rollout}
  \includegraphics[width=0.97\linewidth]{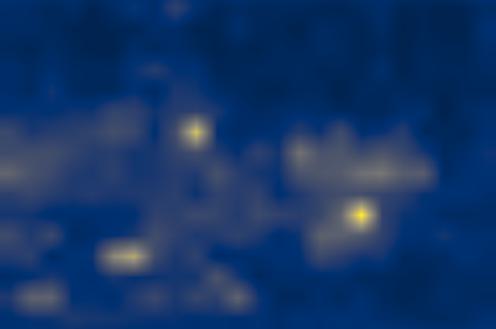}
\end{subfigure}
\begin{subfigure}{.139\textwidth}
  \centering
  \caption*{Attribution}
  \includegraphics[width=0.97\linewidth]{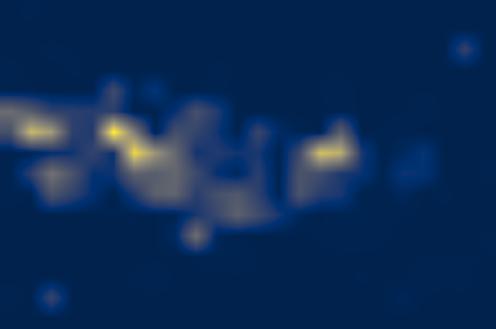}
\end{subfigure}
\begin{subfigure}{.139\textwidth}
  \centering
  \caption*{TAM}
  \includegraphics[width=0.97\linewidth]{images/1/vit_n01695060_5541.JPEG_t_a_m.jpg}
\end{subfigure}
}\vspace{2px}\\
\makebox[0.99\linewidth][c]{
\hspace{.139\textwidth}
\begin{subfigure}{.139\textwidth}
  \centering
  \includegraphics[width=0.97\linewidth]{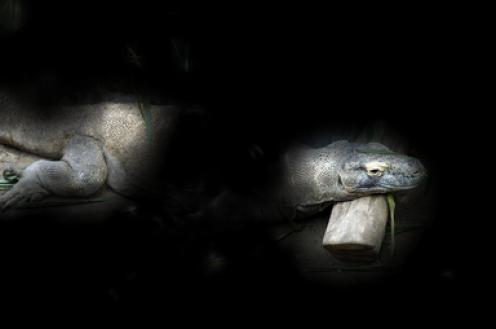}
  \caption*{MaRC Vis.}
\end{subfigure}
\begin{subfigure}{.139\textwidth}
  \centering
  \includegraphics[width=0.97\linewidth]{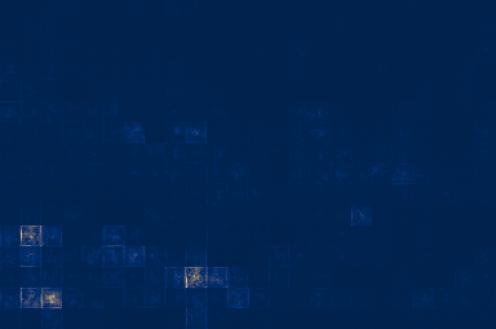}
  \caption*{Saliency}
\end{subfigure}
\begin{subfigure}{.139\textwidth}
  \centering
  \includegraphics[width=0.97\linewidth]{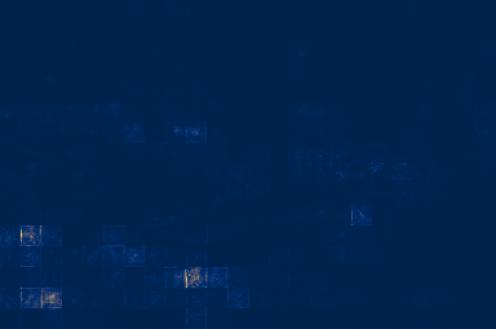}
  \caption*{InputXGrad}
\end{subfigure}
\begin{subfigure}{.139\textwidth}
  \centering
  \includegraphics[width=0.97\linewidth]{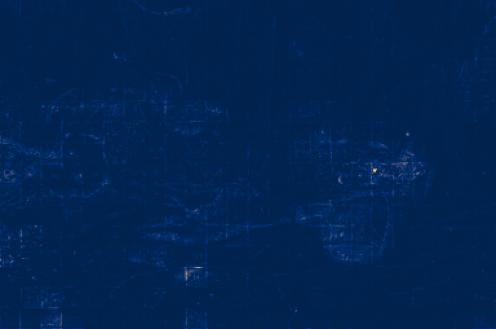}
  \caption*{Integr. Grads}
\end{subfigure}
\begin{subfigure}{.139\textwidth}
  \centering
  \includegraphics[width=0.97\linewidth]{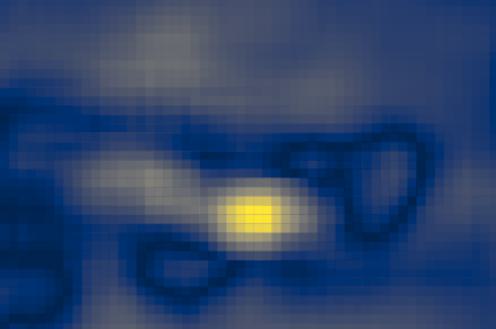}
  \caption*{Occlusion}
\end{subfigure}
\begin{subfigure}{.139\textwidth}
  \centering
  \includegraphics[width=0.97\linewidth]{images/1/vit_n01695060_5541.JPEG_grad_cam.jpg}
  \caption*{Grad-CAM}
\end{subfigure}
}
\caption{Komodo dragon}
\end{figure}

\begin{figure}[H]
\makebox[0.99\linewidth][c]{
\begin{subfigure}{.139\textwidth}
  \centering
  \caption*{Input}
  \includegraphics[width=0.97\linewidth]{images/gt/1.png}
\end{subfigure}
\begin{subfigure}{.139\textwidth}
  \centering
  \caption*{MaRC}
  \includegraphics[width=0.97\linewidth]{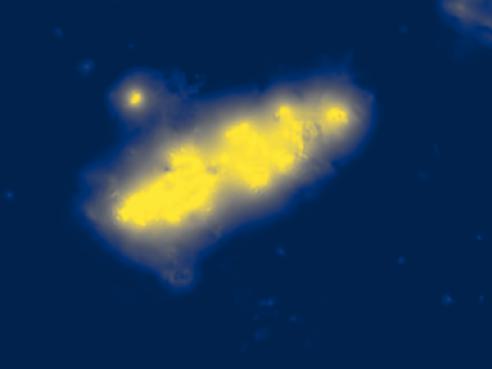}
\end{subfigure}
\begin{subfigure}{.139\textwidth}
  \centering
  \caption*{M-Perturb}
  \includegraphics[width=0.97\linewidth]{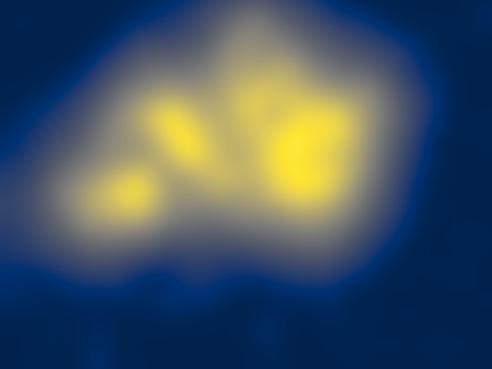}
\end{subfigure}
\begin{subfigure}{.139\textwidth}
  \centering
  \caption*{Raw attention}
  \includegraphics[width=0.97\linewidth]{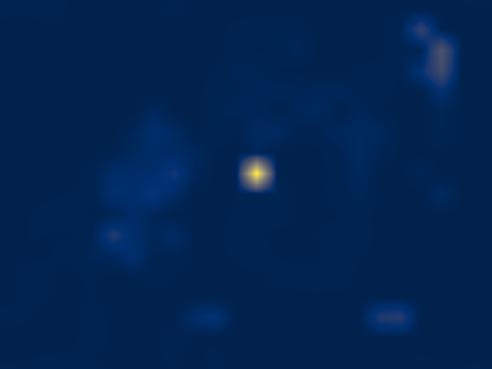}
\end{subfigure}
\begin{subfigure}{.139\textwidth}
  \centering
  \caption*{Rollout}
  \includegraphics[width=0.97\linewidth]{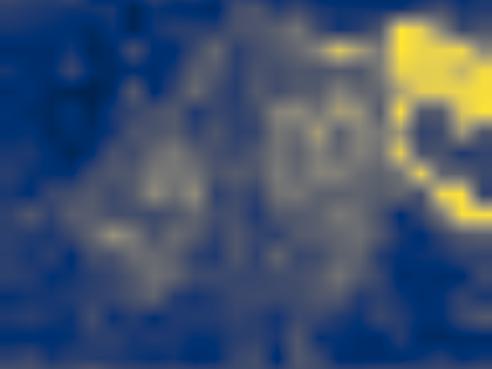}
\end{subfigure}
\begin{subfigure}{.139\textwidth}
  \centering
  \caption*{Attribution}
  \includegraphics[width=0.97\linewidth]{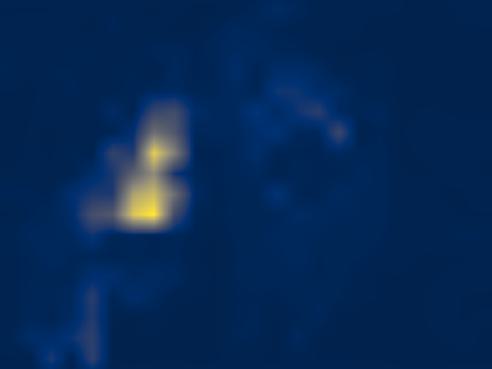}
\end{subfigure}
\begin{subfigure}{.139\textwidth}
  \centering
  \caption*{TAM}
  \includegraphics[width=0.97\linewidth]{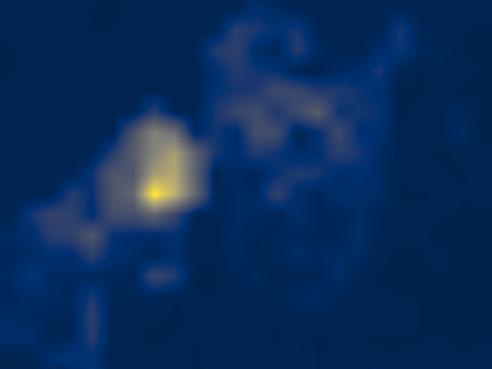}
\end{subfigure}
}\vspace{2px}\\
\makebox[0.99\linewidth][c]{
\hspace{.139\textwidth}
\begin{subfigure}{.139\textwidth}
  \centering
  \includegraphics[width=0.97\linewidth]{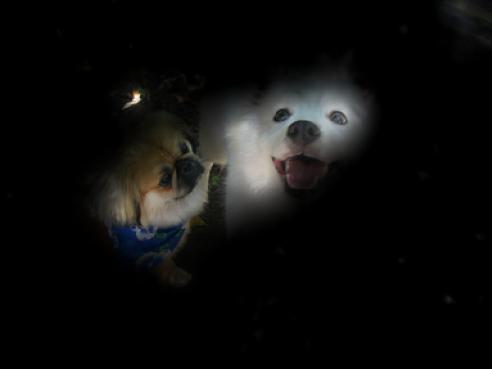}
  \caption*{MaRC Vis.}
\end{subfigure}
\begin{subfigure}{.139\textwidth}
  \centering
  \includegraphics[width=0.97\linewidth]{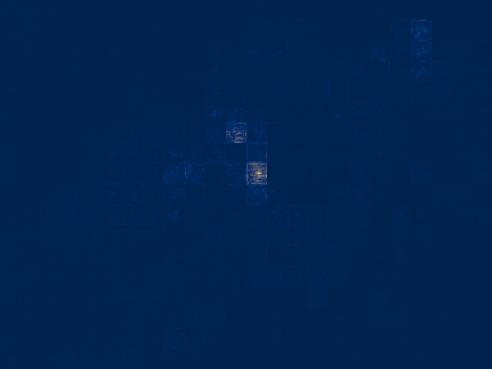}
  \caption*{Saliency}
\end{subfigure}
\begin{subfigure}{.139\textwidth}
  \centering
  \includegraphics[width=0.97\linewidth]{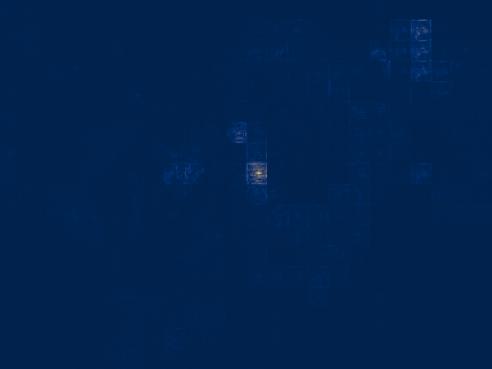}
  \caption*{InputXGrad}
\end{subfigure}
\begin{subfigure}{.139\textwidth}
  \centering
  \includegraphics[width=0.97\linewidth]{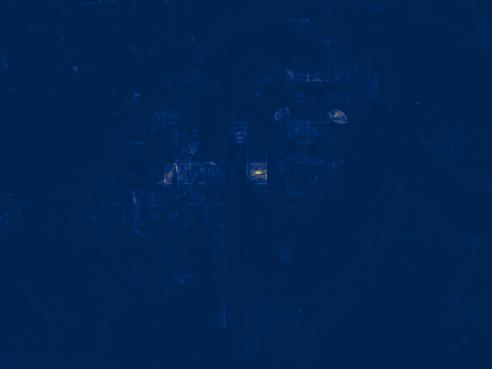}
  \caption*{Integr. Grads}
\end{subfigure}
\begin{subfigure}{.139\textwidth}
  \centering
  \includegraphics[width=0.97\linewidth]{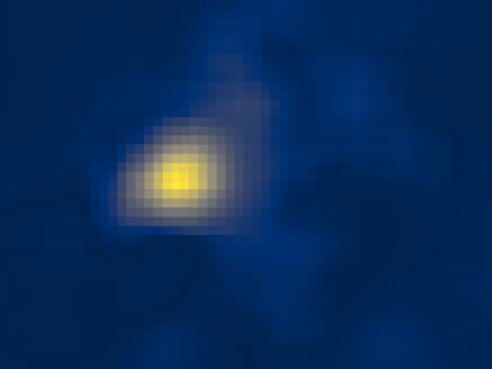}
  \caption*{Occlusion}
\end{subfigure}
\begin{subfigure}{.139\textwidth}
  \centering
  \includegraphics[width=0.97\linewidth]{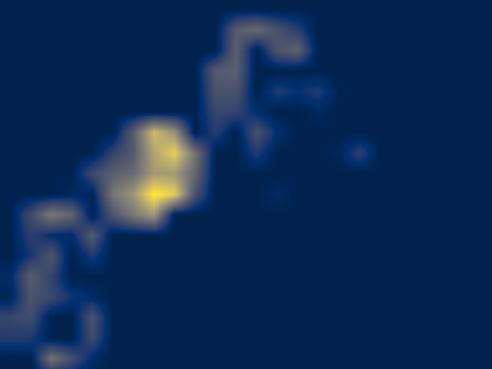}
  \caption*{Grad-CAM}
\end{subfigure}
}
\caption{Pekinese}

\end{figure}

\begin{figure}[H]
\makebox[0.99\linewidth][c]{
\begin{subfigure}{.139\textwidth}
  \centering
  \caption*{Input}
  \includegraphics[width=0.97\linewidth]{images/gt/2.png}
\end{subfigure}
\begin{subfigure}{.139\textwidth}
  \centering
  \caption*{MaRC}
  \includegraphics[width=0.97\linewidth]{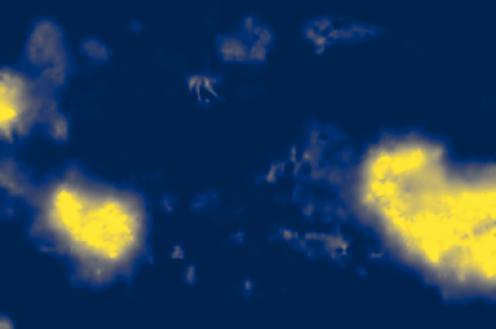}
\end{subfigure}
\begin{subfigure}{.139\textwidth}
  \centering
  \caption*{M-Perturb}
  \includegraphics[width=0.97\linewidth]{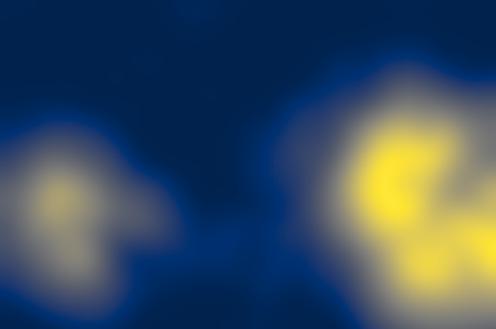}
\end{subfigure}
\begin{subfigure}{.139\textwidth}
  \centering
  \caption*{Raw attention}
  \includegraphics[width=0.97\linewidth]{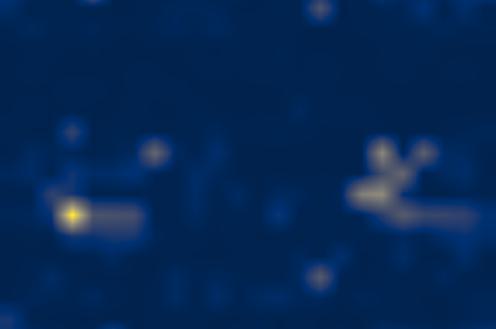}
\end{subfigure}
\begin{subfigure}{.139\textwidth}
  \centering
  \caption*{Rollout}
  \includegraphics[width=0.97\linewidth]{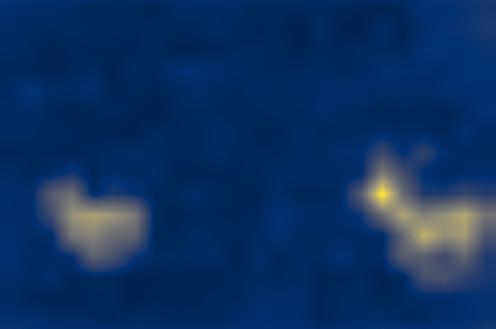}
\end{subfigure}
\begin{subfigure}{.139\textwidth}
  \centering
  \caption*{Attribution}
  \includegraphics[width=0.97\linewidth]{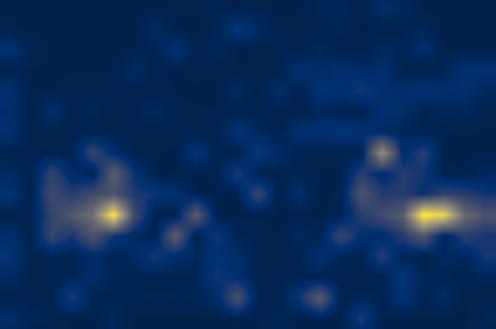}
\end{subfigure}
\begin{subfigure}{.139\textwidth}
  \centering
  \caption*{TAM}
  \includegraphics[width=0.97\linewidth]{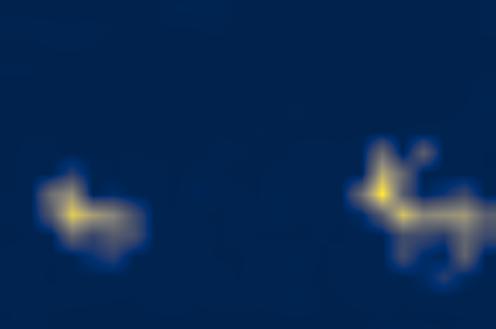}
\end{subfigure}
}\vspace{2px}\\
\makebox[0.99\linewidth][c]{
\hspace{.139\textwidth}
\begin{subfigure}{.139\textwidth}
  \centering
  \includegraphics[width=0.97\linewidth]{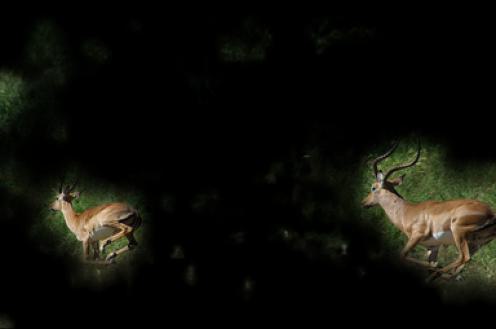}
  \caption*{MaRC Vis.}
\end{subfigure}
\begin{subfigure}{.139\textwidth}
  \centering
  \includegraphics[width=0.97\linewidth]{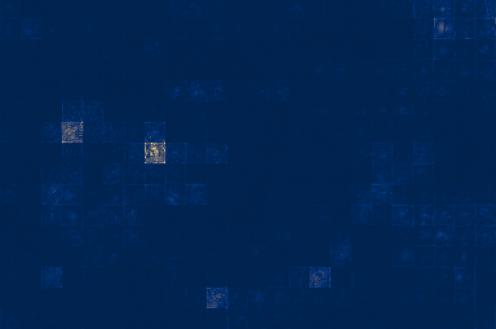}
  \caption*{Saliency}
\end{subfigure}
\begin{subfigure}{.139\textwidth}
  \centering
  \includegraphics[width=0.97\linewidth]{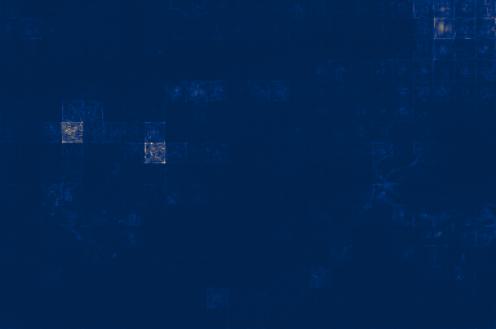}
  \caption*{InputXGrad}
\end{subfigure}
\begin{subfigure}{.139\textwidth}
  \centering
  \includegraphics[width=0.97\linewidth]{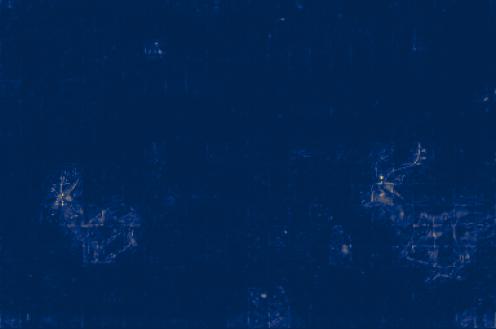}
  \caption*{Integr. Grads}
\end{subfigure}
\begin{subfigure}{.139\textwidth}
  \centering
  \includegraphics[width=0.97\linewidth]{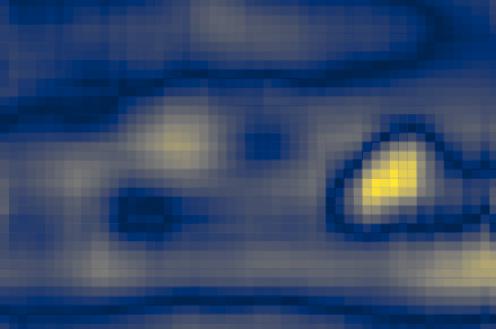}
  \caption*{Occlusion}
\end{subfigure}
\begin{subfigure}{.139\textwidth}
  \centering
  \includegraphics[width=0.97\linewidth]{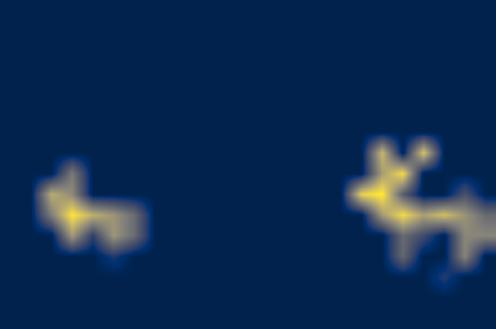}
  \caption*{Grad-CAM}
\end{subfigure}
}

\caption{Impala}

\end{figure}

\begin{figure}[H]
\makebox[0.99\linewidth][c]{
\begin{subfigure}{.139\textwidth}
  \centering
  \caption*{Input}
  \includegraphics[width=0.97\linewidth]{images/gt/3.png}
\end{subfigure}
\begin{subfigure}{.139\textwidth}
  \centering
  \caption*{MaRC}
  \includegraphics[width=0.97\linewidth]{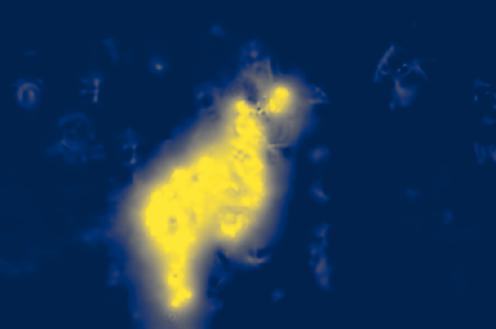}
\end{subfigure}
\begin{subfigure}{.139\textwidth}
  \centering
  \caption*{M-Perturb}
  \includegraphics[width=0.97\linewidth]{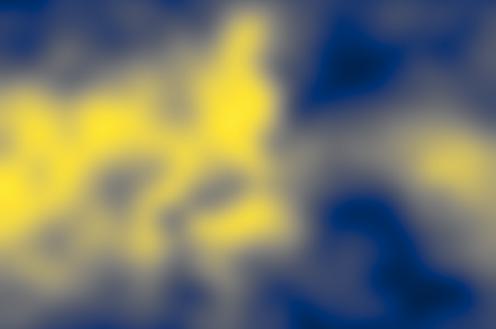}
\end{subfigure}
\begin{subfigure}{.139\textwidth}
  \centering
  \caption*{Raw attention}
  \includegraphics[width=0.97\linewidth]{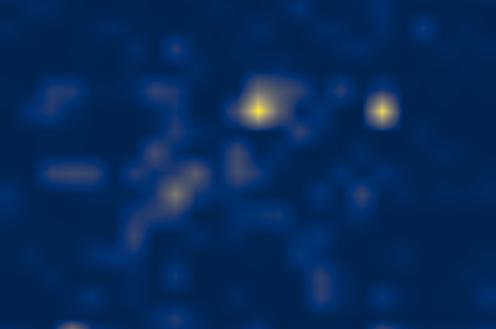}
\end{subfigure}
\begin{subfigure}{.139\textwidth}
  \centering
  \caption*{Rollout}
  \includegraphics[width=0.97\linewidth]{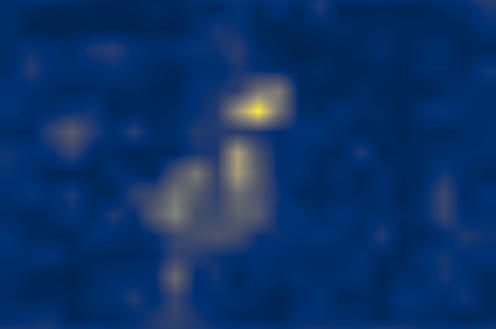}
\end{subfigure}
\begin{subfigure}{.139\textwidth}
  \centering
  \caption*{Attribution}
  \includegraphics[width=0.97\linewidth]{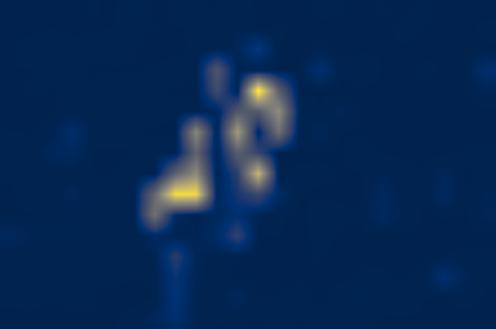}
\end{subfigure}
\begin{subfigure}{.139\textwidth}
  \centering
  \caption*{TAM}
  \includegraphics[width=0.97\linewidth]{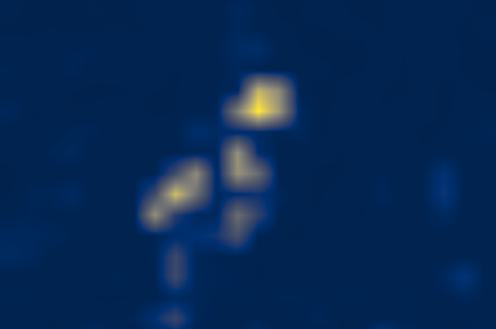}
\end{subfigure}
}\vspace{2px}\\
\makebox[0.99\linewidth][c]{
\hspace{.139\textwidth}
\begin{subfigure}{.139\textwidth}
  \centering
  \includegraphics[width=0.97\linewidth]{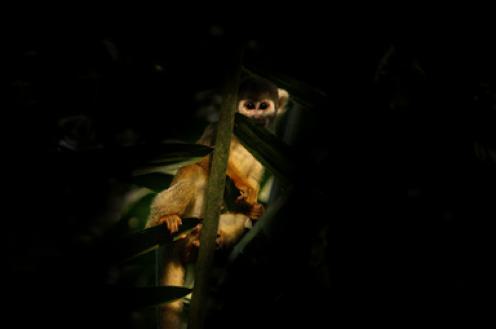}
  \caption*{MaRC Vis.}
\end{subfigure}
\begin{subfigure}{.139\textwidth}
  \centering
  \includegraphics[width=0.97\linewidth]{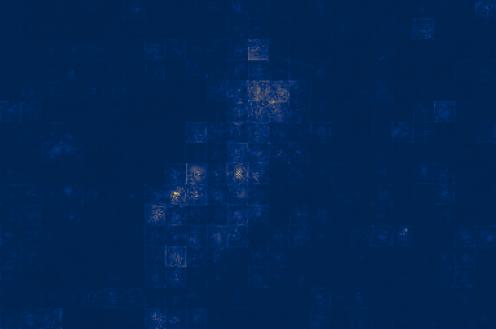}
  \caption*{Saliency}
\end{subfigure}
\begin{subfigure}{.139\textwidth}
  \centering
  \includegraphics[width=0.97\linewidth]{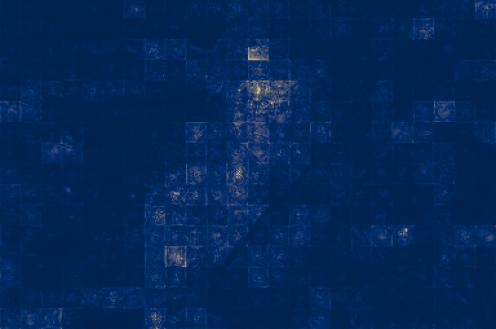}
  \caption*{InputXGrad}
\end{subfigure}
\begin{subfigure}{.139\textwidth}
  \centering
  \includegraphics[width=0.97\linewidth]{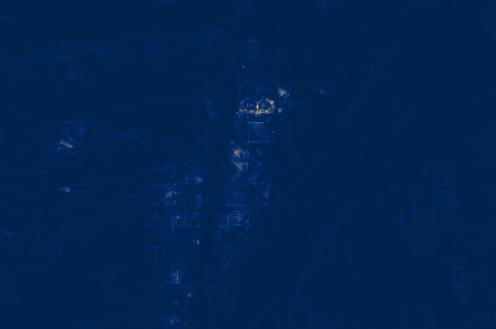}
  \caption*{Integr. Grads}
\end{subfigure}
\begin{subfigure}{.139\textwidth}
  \centering
  \includegraphics[width=0.97\linewidth]{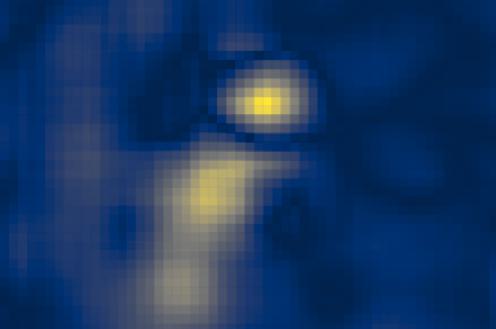}
  \caption*{Occlusion}
\end{subfigure}
\begin{subfigure}{.139\textwidth}
  \centering
  \includegraphics[width=0.97\linewidth]{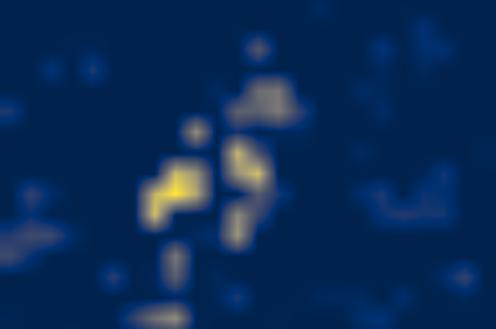}
  \caption*{Grad-CAM}
\end{subfigure}
}

\caption{Squirrel monkey}

\end{figure}

\begin{figure}[H]
\makebox[0.99\linewidth][c]{
\begin{subfigure}{.139\textwidth}
  \centering
  \caption*{Input}
  \includegraphics[width=0.97\linewidth]{images/gt/4.png}
\end{subfigure}
\begin{subfigure}{.139\textwidth}
  \centering
  \caption*{MaRC}
  \includegraphics[width=0.97\linewidth]{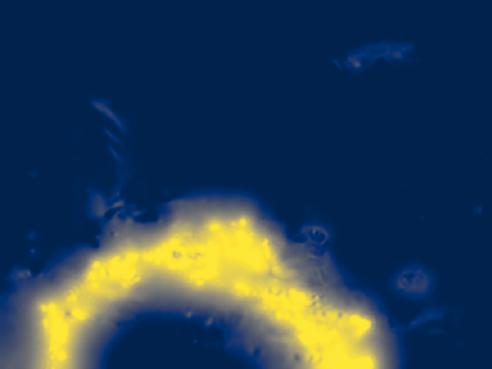}
\end{subfigure}
\begin{subfigure}{.139\textwidth}
  \centering
  \caption*{M-Perturb}
  \includegraphics[width=0.97\linewidth]{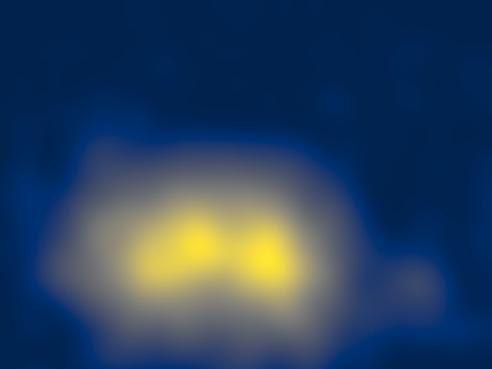}
\end{subfigure}
\begin{subfigure}{.139\textwidth}
  \centering
  \caption*{Raw attention}
  \includegraphics[width=0.97\linewidth]{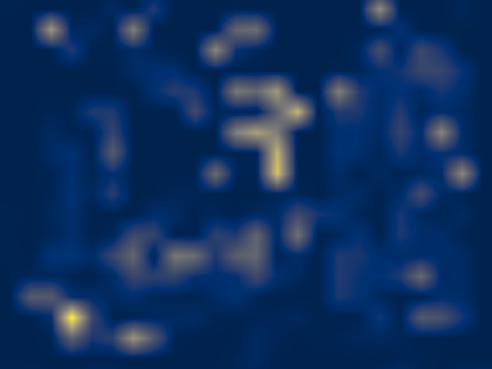}
\end{subfigure}
\begin{subfigure}{.139\textwidth}
  \centering
  \caption*{Rollout}
  \includegraphics[width=0.97\linewidth]{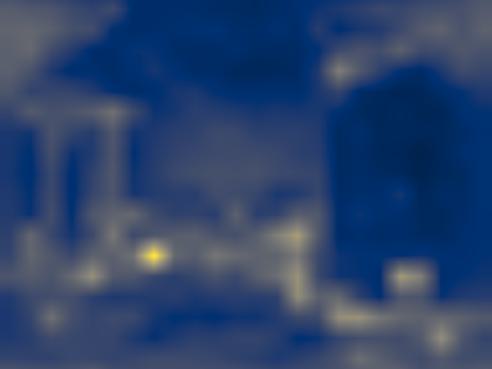}
\end{subfigure}
\begin{subfigure}{.139\textwidth}
  \centering
  \caption*{Attribution}
  \includegraphics[width=0.97\linewidth]{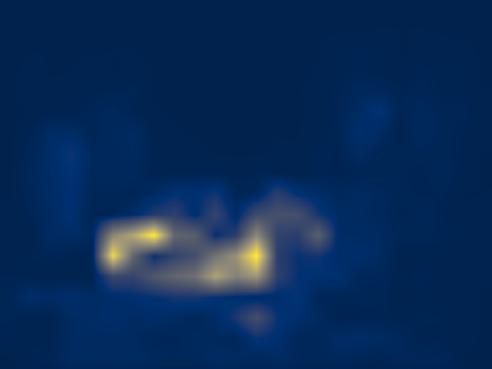}
\end{subfigure}
\begin{subfigure}{.139\textwidth}
  \centering
  \caption*{TAM}
  \includegraphics[width=0.97\linewidth]{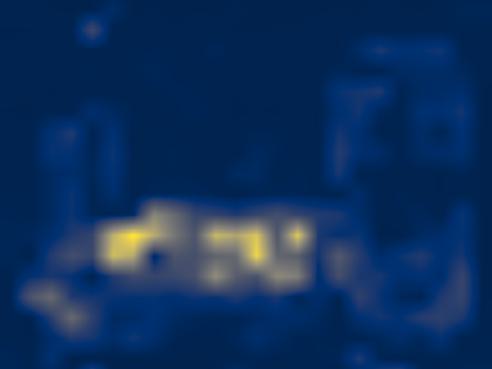}
\end{subfigure}
}\vspace{2px}\\
\makebox[0.99\linewidth][c]{
\hspace{.139\textwidth}
\begin{subfigure}{.139\textwidth}
  \centering
  \includegraphics[width=0.97\linewidth]{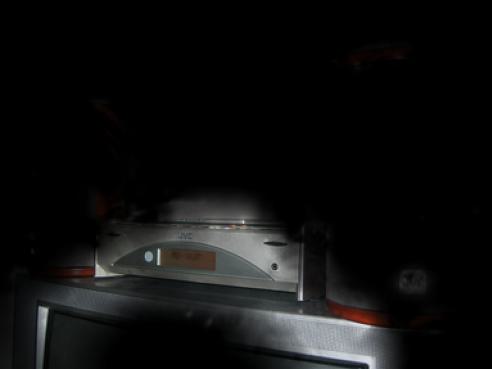}
  \caption*{MaRC Vis.}
\end{subfigure}
\begin{subfigure}{.139\textwidth}
  \centering
  \includegraphics[width=0.97\linewidth]{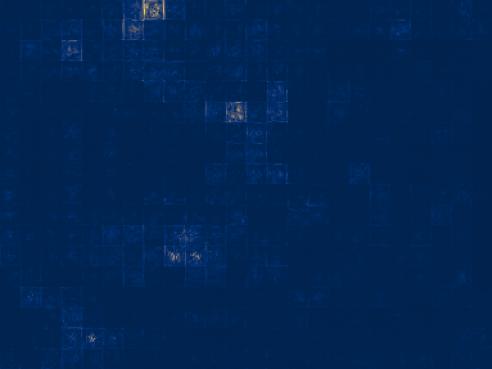}
  \caption*{Saliency}
\end{subfigure}
\begin{subfigure}{.139\textwidth}
  \centering
  \includegraphics[width=0.97\linewidth]{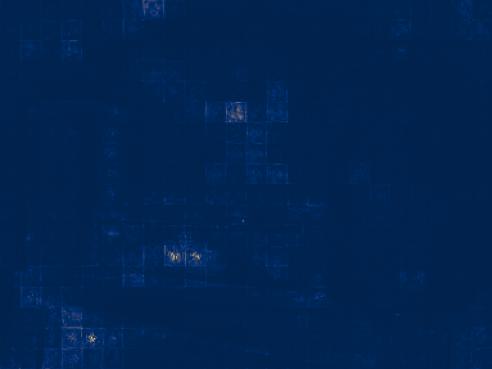}
  \caption*{InputXGrad}
\end{subfigure}
\begin{subfigure}{.139\textwidth}
  \centering
  \includegraphics[width=0.97\linewidth]{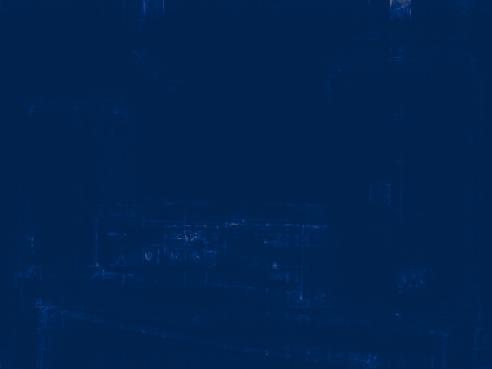}
  \caption*{Integr. Grads}
\end{subfigure}
\begin{subfigure}{.139\textwidth}
  \centering
  \includegraphics[width=0.97\linewidth]{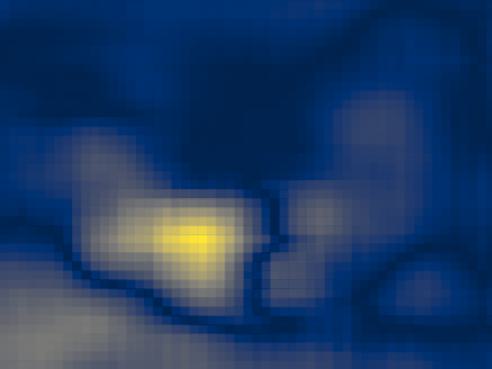}
  \caption*{Occlusion}
\end{subfigure}
\begin{subfigure}{.139\textwidth}
  \centering
  \includegraphics[width=0.97\linewidth]{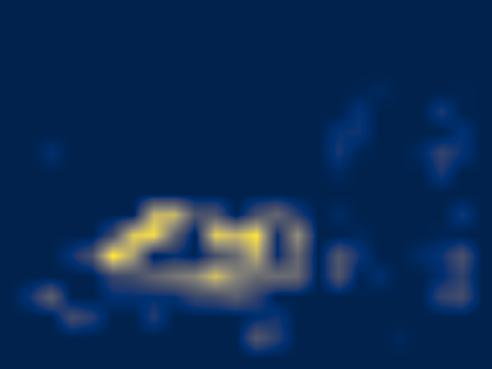}
  \caption*{Grad-CAM}
\end{subfigure}
}

\caption{CD player}

\end{figure}

\begin{figure}[H]
\makebox[0.99\linewidth][c]{
\begin{subfigure}{.139\textwidth}
  \centering
  \caption*{Input}
  \includegraphics[width=0.97\linewidth]{images/gt/5.png}
\end{subfigure}
\begin{subfigure}{.139\textwidth}
  \centering
  \caption*{MaRC}
  \includegraphics[width=0.97\linewidth]{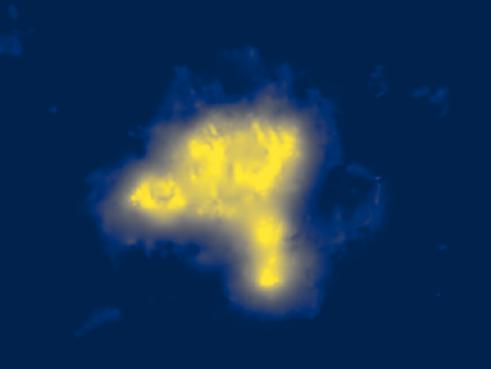}
\end{subfigure}
\begin{subfigure}{.139\textwidth}
  \centering
  \caption*{M-Perturb}
  \includegraphics[width=0.97\linewidth]{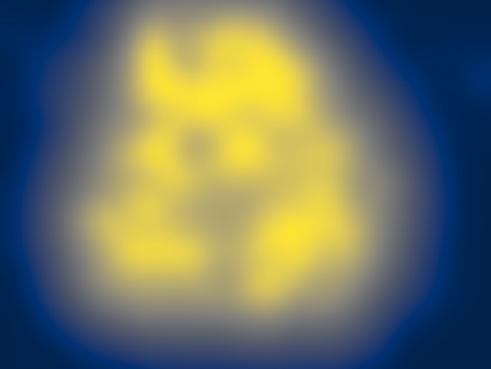}
\end{subfigure}
\begin{subfigure}{.139\textwidth}
  \centering
  \caption*{Raw attention}
  \includegraphics[width=0.97\linewidth]{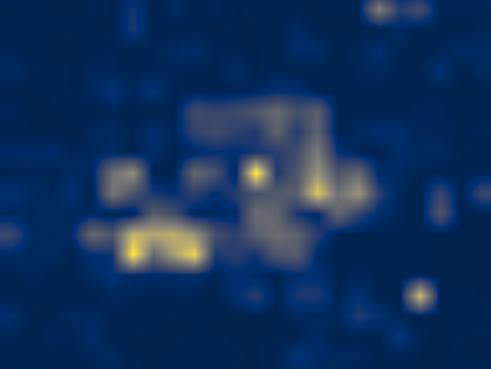}
\end{subfigure}
\begin{subfigure}{.139\textwidth}
  \centering
  \caption*{Rollout}
  \includegraphics[width=0.97\linewidth]{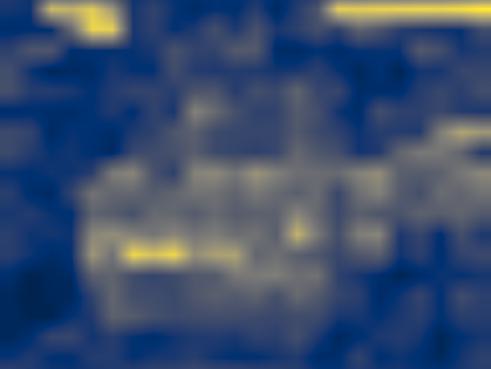}
\end{subfigure}
\begin{subfigure}{.139\textwidth}
  \centering
  \caption*{Attribution}
  \includegraphics[width=0.97\linewidth]{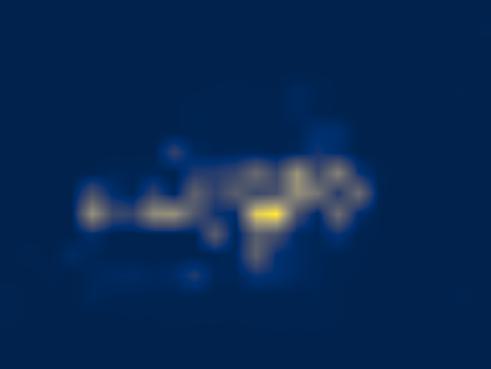}
\end{subfigure}
\begin{subfigure}{.139\textwidth}
  \centering
  \caption*{TAM}
  \includegraphics[width=0.97\linewidth]{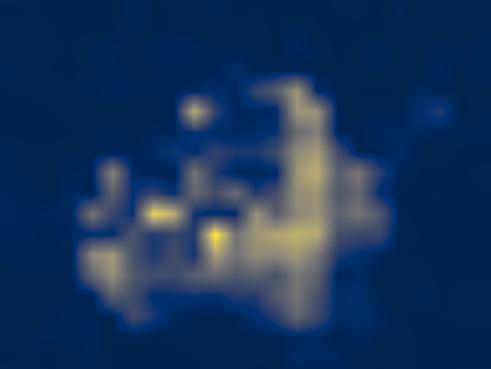}
\end{subfigure}
}\vspace{2px}\\
\makebox[0.99\linewidth][c]{
\hspace{.139\textwidth}
\begin{subfigure}{.139\textwidth}
  \centering
  \includegraphics[width=0.97\linewidth]{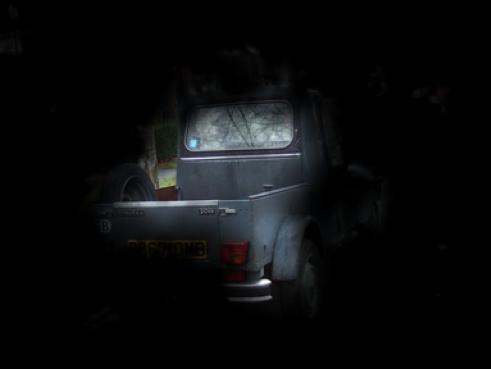}
  \caption*{MaRC Vis.}
\end{subfigure}
\begin{subfigure}{.139\textwidth}
  \centering
  \includegraphics[width=0.97\linewidth]{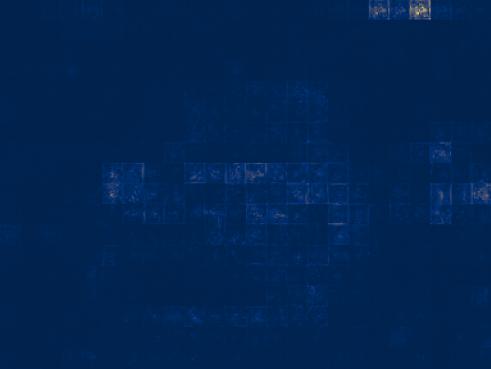}
  \caption*{Saliency}
\end{subfigure}
\begin{subfigure}{.139\textwidth}
  \centering
  \includegraphics[width=0.97\linewidth]{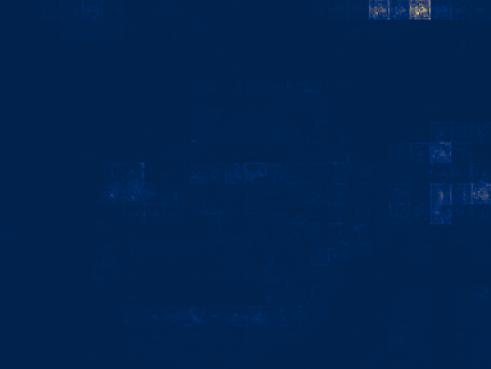}
  \caption*{InputXGrad}
\end{subfigure}
\begin{subfigure}{.139\textwidth}
  \centering
  \includegraphics[width=0.97\linewidth]{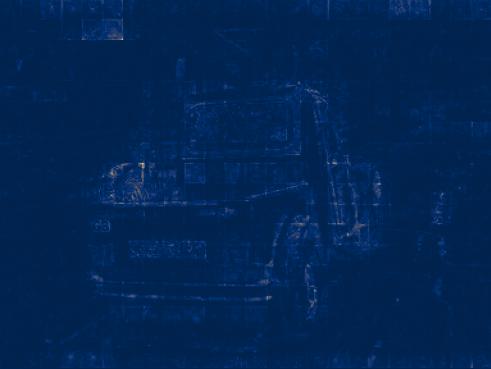}
  \caption*{Integr. Grads}
\end{subfigure}
\begin{subfigure}{.139\textwidth}
  \centering
  \includegraphics[width=0.97\linewidth]{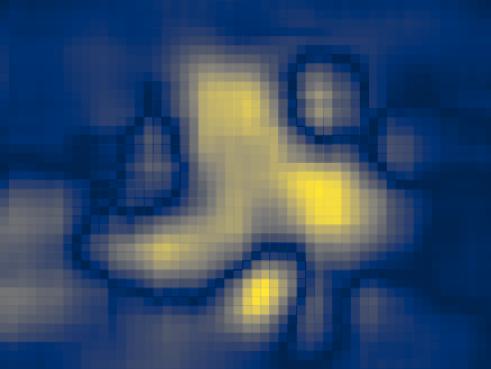}
  \caption*{Occlusion}
\end{subfigure}
\begin{subfigure}{.139\textwidth}
  \centering
  \includegraphics[width=0.97\linewidth]{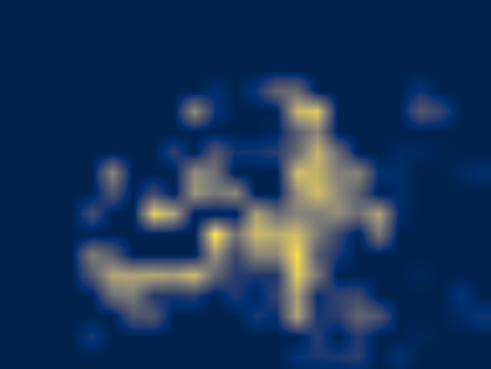}
  \caption*{Grad-CAM}
\end{subfigure}
}

\caption{Pickup}

\end{figure}

\begin{figure}[H]
\makebox[0.99\linewidth][c]{
\begin{subfigure}{.139\textwidth}
  \centering
  \caption*{Input}
  \includegraphics[width=0.97\linewidth]{images/gt/6.png}
\end{subfigure}
\begin{subfigure}{.139\textwidth}
  \centering
  \caption*{MaRC}
  \includegraphics[width=0.97\linewidth]{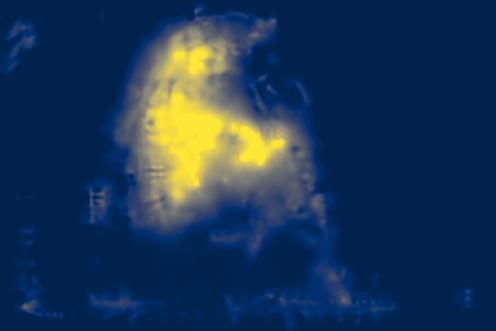}
\end{subfigure}
\begin{subfigure}{.139\textwidth}
  \centering
  \caption*{M-Perturb}
  \includegraphics[width=0.97\linewidth]{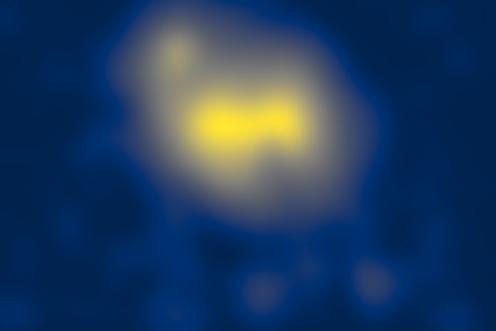}
\end{subfigure}
\begin{subfigure}{.139\textwidth}
  \centering
  \caption*{Raw attention}
  \includegraphics[width=0.97\linewidth]{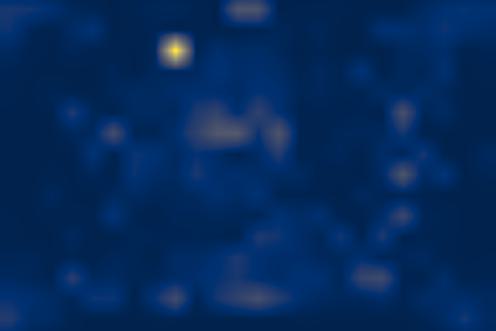}
\end{subfigure}
\begin{subfigure}{.139\textwidth}
  \centering
  \caption*{Rollout}
  \includegraphics[width=0.97\linewidth]{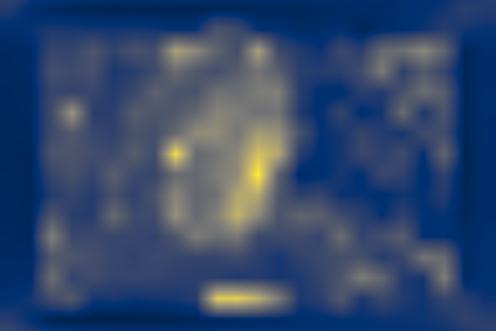}
\end{subfigure}
\begin{subfigure}{.139\textwidth}
  \centering
  \caption*{Attribution}
  \includegraphics[width=0.97\linewidth]{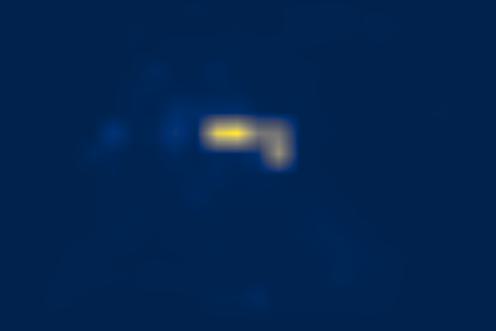}
\end{subfigure}
\begin{subfigure}{.139\textwidth}
  \centering
  \caption*{TAM}
  \includegraphics[width=0.97\linewidth]{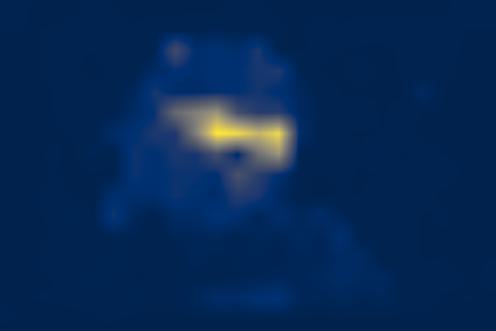}
\end{subfigure}
}\vspace{2px}\\
\makebox[0.99\linewidth][c]{
\hspace{.139\textwidth}
\begin{subfigure}{.139\textwidth}
  \centering
  \includegraphics[width=0.97\linewidth]{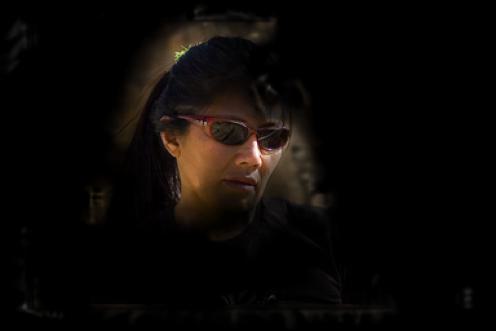}
  \caption*{MaRC Vis.}
\end{subfigure}
\begin{subfigure}{.139\textwidth}
  \centering
  \includegraphics[width=0.97\linewidth]{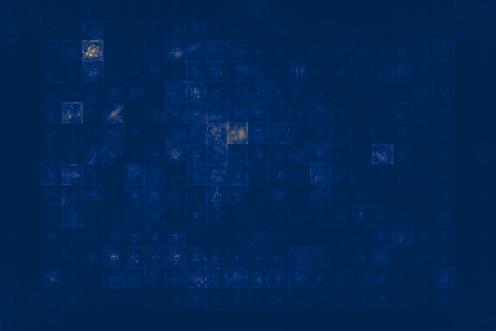}
  \caption*{Saliency}
\end{subfigure}
\begin{subfigure}{.139\textwidth}
  \centering
  \includegraphics[width=0.97\linewidth]{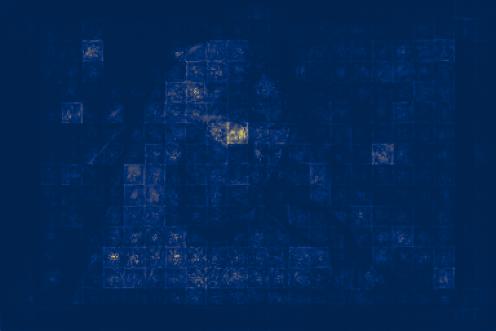}
  \caption*{InputXGrad}
\end{subfigure}
\begin{subfigure}{.139\textwidth}
  \centering
  \includegraphics[width=0.97\linewidth]{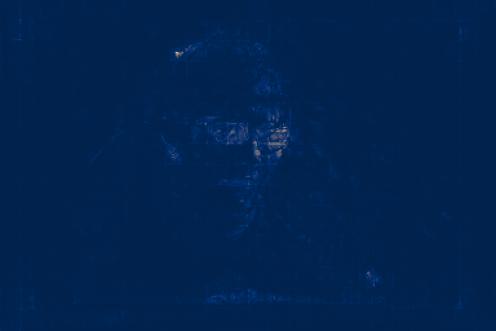}
  \caption*{Integr. Grads}
\end{subfigure}
\begin{subfigure}{.139\textwidth}
  \centering
  \includegraphics[width=0.97\linewidth]{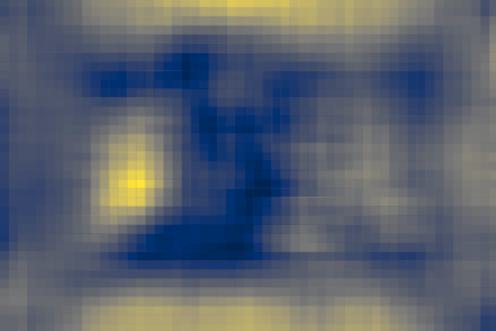}
  \caption*{Occlusion}
\end{subfigure}
\begin{subfigure}{.139\textwidth}
  \centering
  \includegraphics[width=0.97\linewidth]{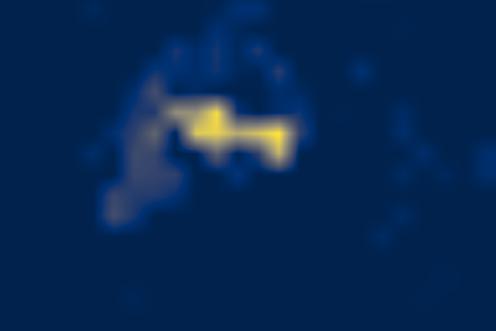}
  \caption*{Grad-CAM}
\end{subfigure}
}

\caption{Sunglasses}

\end{figure}

\begin{figure}[H]
\makebox[0.99\linewidth][c]{
\begin{subfigure}{.139\textwidth}
  \centering
  \caption*{Input}
  \includegraphics[width=0.97\linewidth]{images/gt/7.png}
\end{subfigure}
\begin{subfigure}{.139\textwidth}
  \centering
  \caption*{MaRC}
  \includegraphics[width=0.97\linewidth]{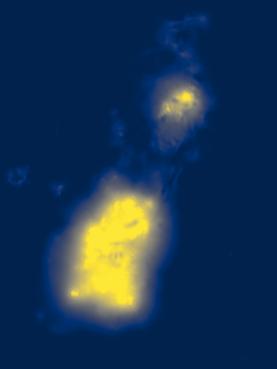}
\end{subfigure}
\begin{subfigure}{.139\textwidth}
  \centering
  \caption*{M-Perturb}
  \includegraphics[width=0.97\linewidth]{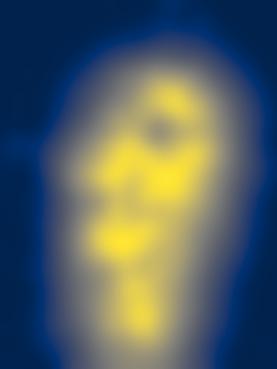}
\end{subfigure}
\begin{subfigure}{.139\textwidth}
  \centering
  \caption*{Raw attention}
  \includegraphics[width=0.97\linewidth]{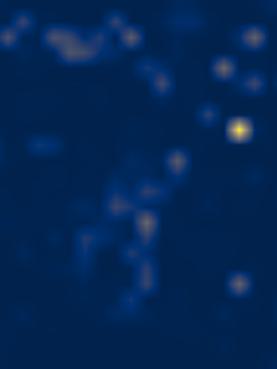}
\end{subfigure}
\begin{subfigure}{.139\textwidth}
  \centering
  \caption*{Rollout}
  \includegraphics[width=0.97\linewidth]{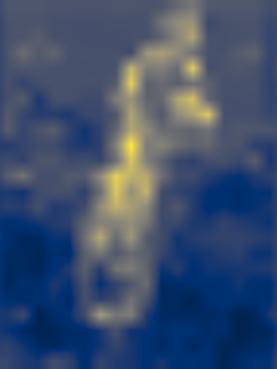}
\end{subfigure}
\begin{subfigure}{.139\textwidth}
  \centering
  \caption*{Attribution}
  \includegraphics[width=0.97\linewidth]{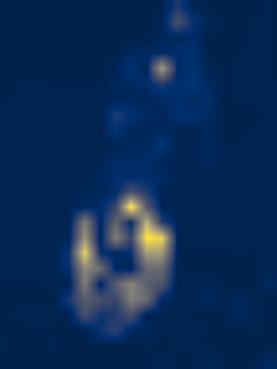}
\end{subfigure}
\begin{subfigure}{.139\textwidth}
  \centering
  \caption*{TAM}
  \includegraphics[width=0.97\linewidth]{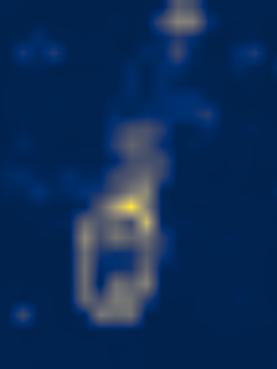}
\end{subfigure}
}\vspace{2px}\\
\makebox[0.99\linewidth][c]{
\hspace{.139\textwidth}
\begin{subfigure}{.139\textwidth}
  \centering
  \includegraphics[width=0.97\linewidth]{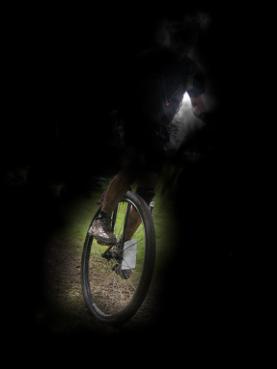}
  \caption*{MaRC Vis.}
\end{subfigure}
\begin{subfigure}{.139\textwidth}
  \centering
  \includegraphics[width=0.97\linewidth]{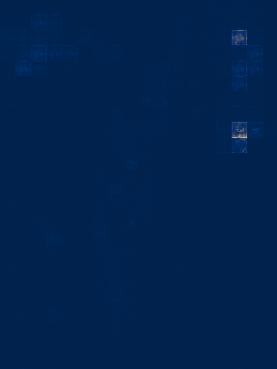}
  \caption*{Saliency}
\end{subfigure}
\begin{subfigure}{.139\textwidth}
  \centering
  \includegraphics[width=0.97\linewidth]{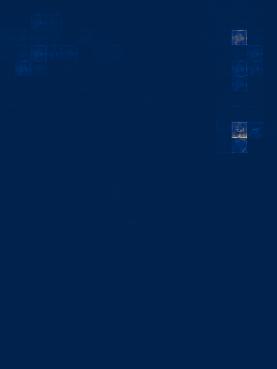}
  \caption*{InputXGrad}
\end{subfigure}
\begin{subfigure}{.139\textwidth}
  \centering
  \includegraphics[width=0.97\linewidth]{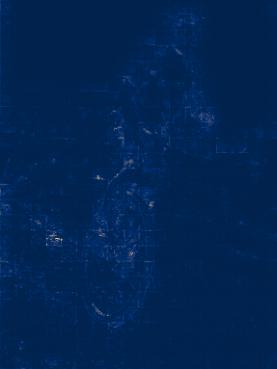}
  \caption*{Integr. Grads}
\end{subfigure}
\begin{subfigure}{.139\textwidth}
  \centering
  \includegraphics[width=0.97\linewidth]{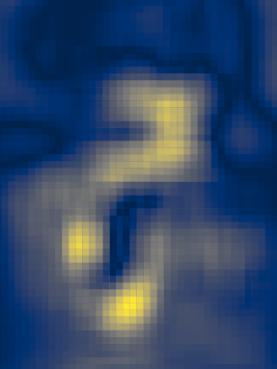}
  \caption*{Occlusion}
\end{subfigure}
\begin{subfigure}{.139\textwidth}
  \centering
  \includegraphics[width=0.97\linewidth]{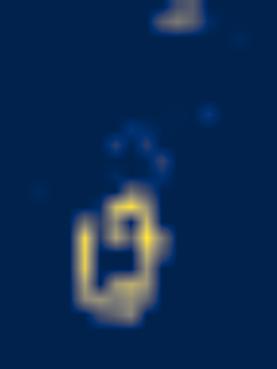}
  \caption*{Grad-CAM}
\end{subfigure}
}

\caption{Unicycle}
\end{figure}

\begin{figure}[H]
\makebox[0.99\linewidth][c]{
\begin{subfigure}{.139\textwidth}
  \centering
  \caption*{Input}
  \includegraphics[width=0.97\linewidth]{images/gt/8.png}
\end{subfigure}
\begin{subfigure}{.139\textwidth}
  \centering
  \caption*{MaRC}
  \includegraphics[width=0.97\linewidth]{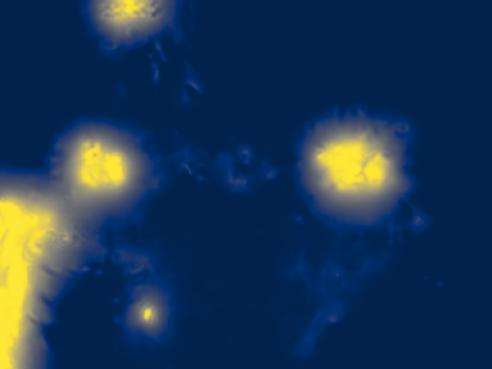}
\end{subfigure}
\begin{subfigure}{.139\textwidth}
  \centering
  \caption*{M-Perturb}
  \includegraphics[width=0.97\linewidth]{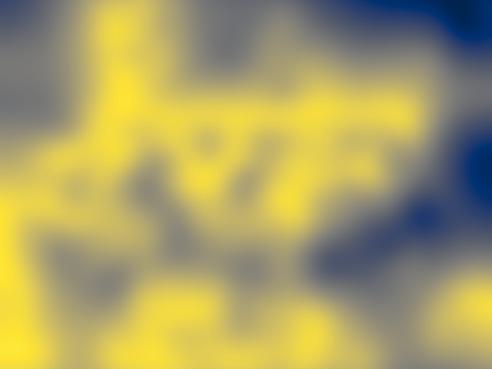}
\end{subfigure}
\begin{subfigure}{.139\textwidth}
  \centering
  \caption*{Raw attention}
  \includegraphics[width=0.97\linewidth]{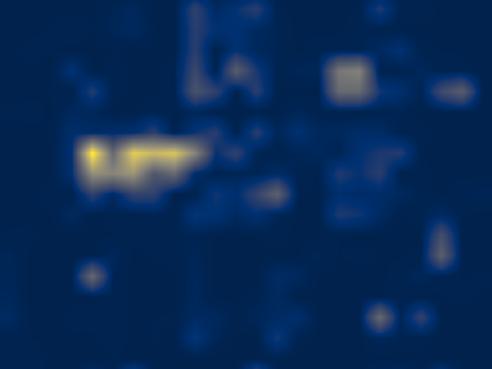}
\end{subfigure}
\begin{subfigure}{.139\textwidth}
  \centering
  \caption*{Rollout}
  \includegraphics[width=0.97\linewidth]{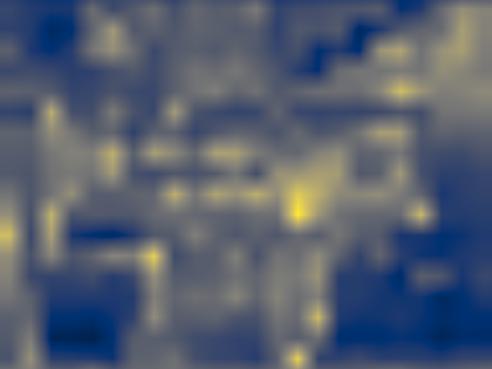}
\end{subfigure}
\begin{subfigure}{.139\textwidth}
  \centering
  \caption*{Attribution}
  \includegraphics[width=0.97\linewidth]{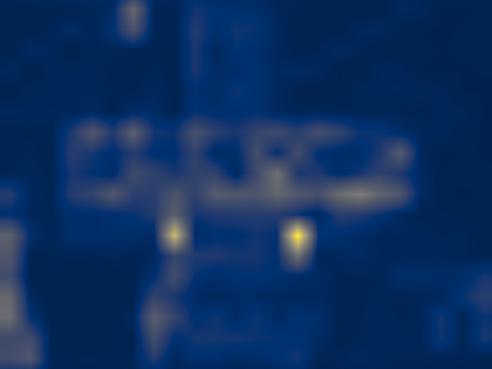}
\end{subfigure}
\begin{subfigure}{.139\textwidth}
  \centering
  \caption*{TAM}
  \includegraphics[width=0.97\linewidth]{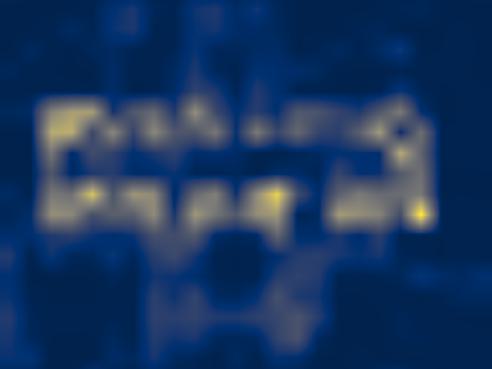}
\end{subfigure}
}\vspace{2px}\\
\makebox[0.99\linewidth][c]{
\hspace{.139\textwidth}
\begin{subfigure}{.139\textwidth}
  \centering
  \includegraphics[width=0.97\linewidth]{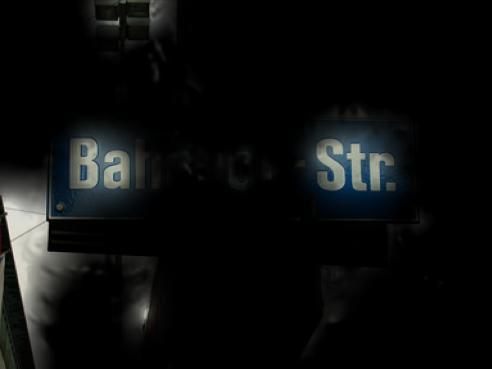}
  \caption*{MaRC Vis.}
\end{subfigure}
\begin{subfigure}{.139\textwidth}
  \centering
  \includegraphics[width=0.97\linewidth]{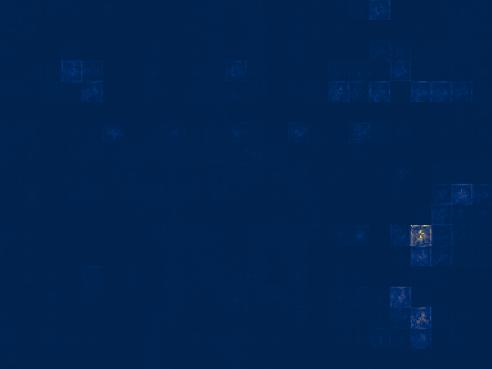}
  \caption*{Saliency}
\end{subfigure}
\begin{subfigure}{.139\textwidth}
  \centering
  \includegraphics[width=0.97\linewidth]{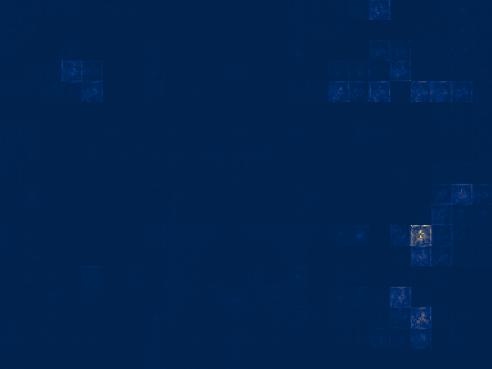}
  \caption*{InputXGrad}
\end{subfigure}
\begin{subfigure}{.139\textwidth}
  \centering
  \includegraphics[width=0.97\linewidth]{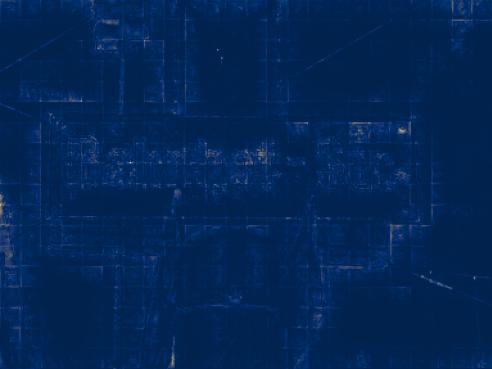}
  \caption*{Integr. Grads}
\end{subfigure}
\begin{subfigure}{.139\textwidth}
  \centering
  \includegraphics[width=0.97\linewidth]{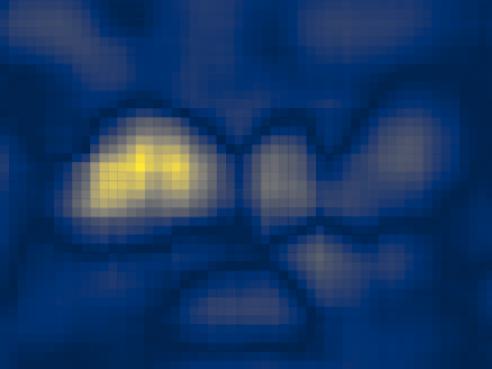}
  \caption*{Occlusion}
\end{subfigure}
\begin{subfigure}{.139\textwidth}
  \centering
  \includegraphics[width=0.97\linewidth]{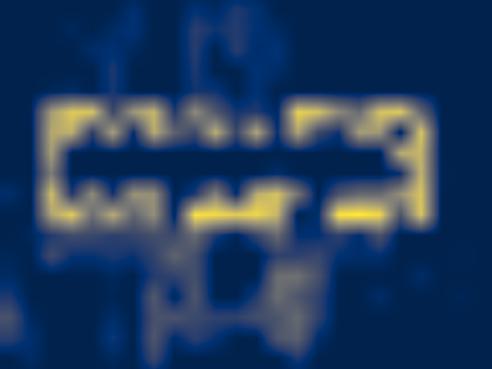}
  \caption*{Grad-CAM}
\end{subfigure}
}

\caption{Unicycle}

\end{figure}

\begin{figure}[H]
\makebox[0.99\linewidth][c]{
\begin{subfigure}{.139\textwidth}
  \centering
  \caption*{Input}
  \includegraphics[width=0.97\linewidth]{images/gt/9.png}
\end{subfigure}
\begin{subfigure}{.139\textwidth}
  \centering
  \caption*{MaRC}
  \includegraphics[width=0.97\linewidth]{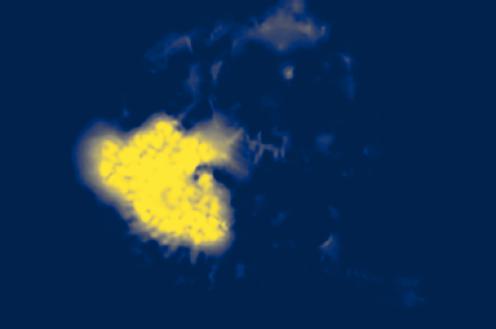}
\end{subfigure}
\begin{subfigure}{.139\textwidth}
  \centering
  \caption*{M-Perturb}
  \includegraphics[width=0.97\linewidth]{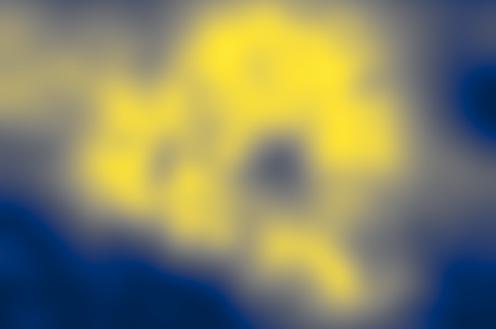}
\end{subfigure}
\begin{subfigure}{.139\textwidth}
  \centering
  \caption*{Raw attention}
  \includegraphics[width=0.97\linewidth]{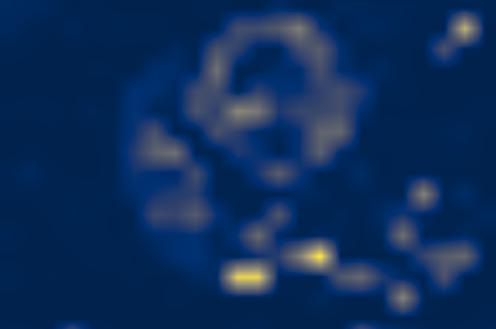}
\end{subfigure}
\begin{subfigure}{.139\textwidth}
  \centering
  \caption*{Rollout}
  \includegraphics[width=0.97\linewidth]{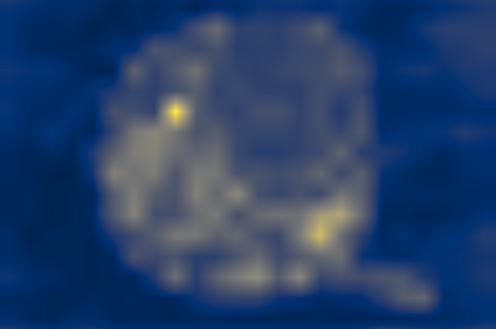}
\end{subfigure}
\begin{subfigure}{.139\textwidth}
  \centering
  \caption*{Attribution}
  \includegraphics[width=0.97\linewidth]{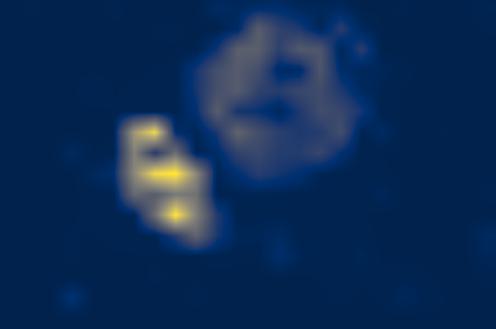}
\end{subfigure}
\begin{subfigure}{.139\textwidth}
  \centering
  \caption*{TAM}
  \includegraphics[width=0.97\linewidth]{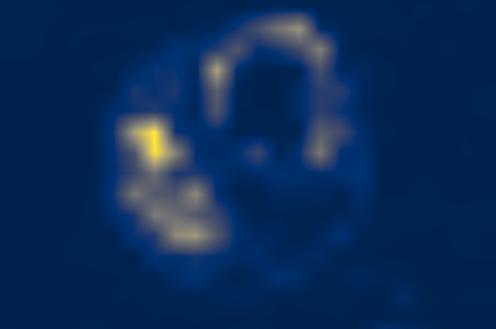}
\end{subfigure}
}\vspace{2px}\\
\makebox[0.99\linewidth][c]{
\hspace{.139\textwidth}
\begin{subfigure}{.139\textwidth}
  \centering
  \includegraphics[width=0.97\linewidth]{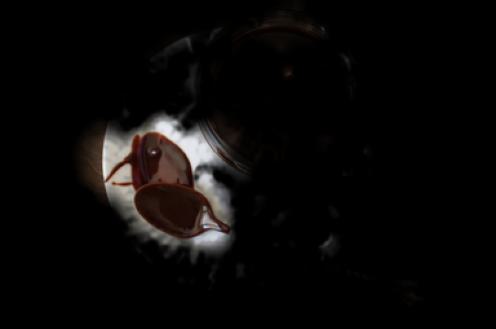}
  \caption*{MaRC Vis.}
\end{subfigure}
\begin{subfigure}{.139\textwidth}
  \centering
  \includegraphics[width=0.97\linewidth]{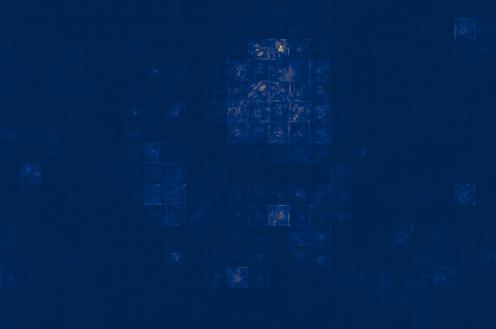}
  \caption*{Saliency}
\end{subfigure}
\begin{subfigure}{.139\textwidth}
  \centering
  \includegraphics[width=0.97\linewidth]{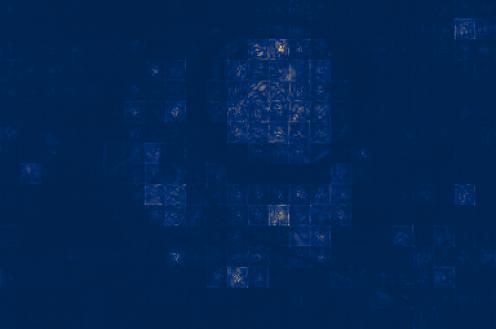}
  \caption*{InputXGrad}
\end{subfigure}
\begin{subfigure}{.139\textwidth}
  \centering
  \includegraphics[width=0.97\linewidth]{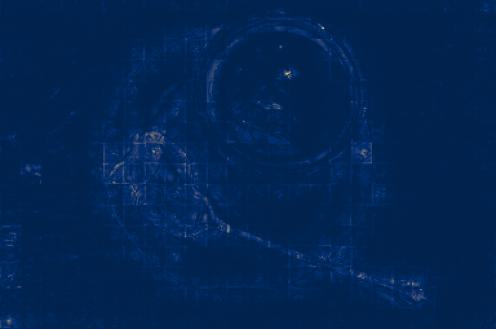}
  \caption*{Integr. Grads}
\end{subfigure}
\begin{subfigure}{.139\textwidth}
  \centering
  \includegraphics[width=0.97\linewidth]{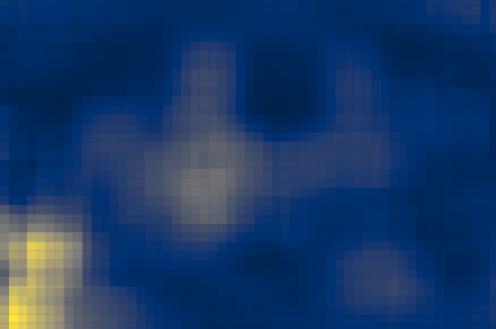}
  \caption*{Occlusion}
\end{subfigure}
\begin{subfigure}{.139\textwidth}
  \centering
  \includegraphics[width=0.97\linewidth]{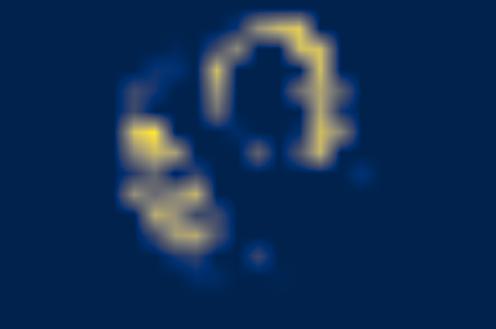}
  \caption*{Grad-CAM}
\end{subfigure}
}

\caption{Chocolate sauce}

\end{figure}

\begin{figure}[H]
\makebox[0.99\linewidth][c]{
\begin{subfigure}{.139\textwidth}
  \centering
  \caption*{Input}
  \includegraphics[width=0.97\linewidth]{images/gt/10.png}
\end{subfigure}
\begin{subfigure}{.139\textwidth}
  \centering
  \caption*{MaRC}
  \includegraphics[width=0.97\linewidth]{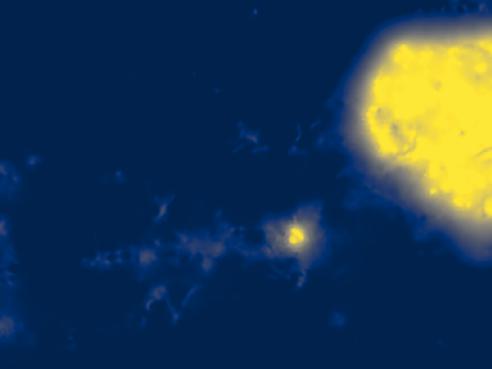}
\end{subfigure}
\begin{subfigure}{.139\textwidth}
  \centering
  \caption*{M-Perturb}
  \includegraphics[width=0.97\linewidth]{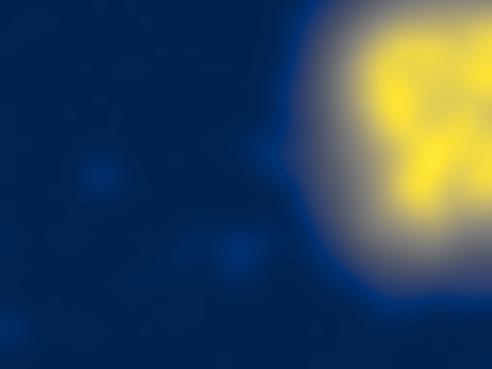}
\end{subfigure}
\begin{subfigure}{.139\textwidth}
  \centering
  \caption*{Raw attention}
  \includegraphics[width=0.97\linewidth]{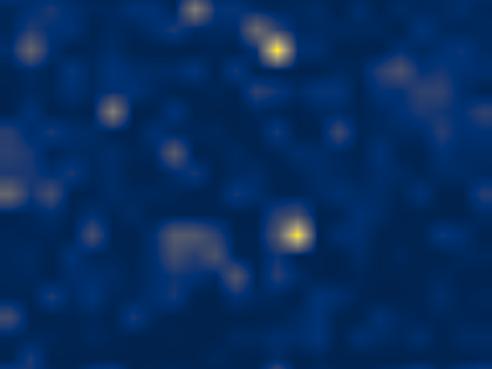}
\end{subfigure}
\begin{subfigure}{.139\textwidth}
  \centering
  \caption*{Rollout}
  \includegraphics[width=0.97\linewidth]{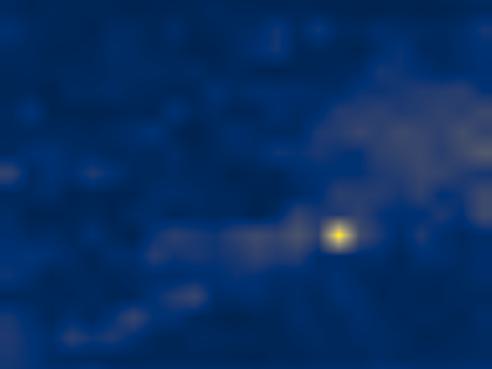}
\end{subfigure}
\begin{subfigure}{.139\textwidth}
  \centering
  \caption*{Attribution}
  \includegraphics[width=0.97\linewidth]{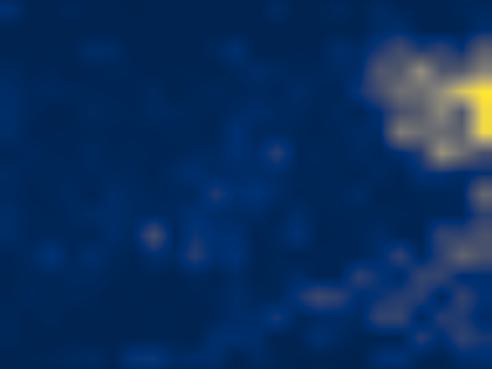}
\end{subfigure}
\begin{subfigure}{.139\textwidth}
  \centering
  \caption*{TAM}
  \includegraphics[width=0.97\linewidth]{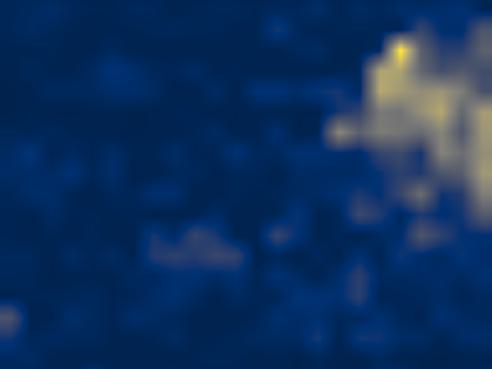}
\end{subfigure}
}\vspace{2px}\\
\makebox[0.99\linewidth][c]{
\hspace{.139\textwidth}
\begin{subfigure}{.139\textwidth}
  \centering
  \includegraphics[width=0.97\linewidth]{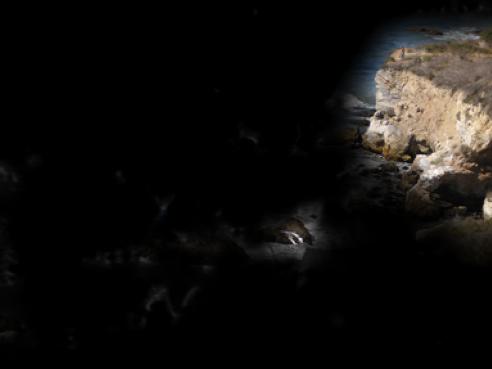}
  \caption*{MaRC Vis.}
\end{subfigure}
\begin{subfigure}{.139\textwidth}
  \centering
  \includegraphics[width=0.97\linewidth]{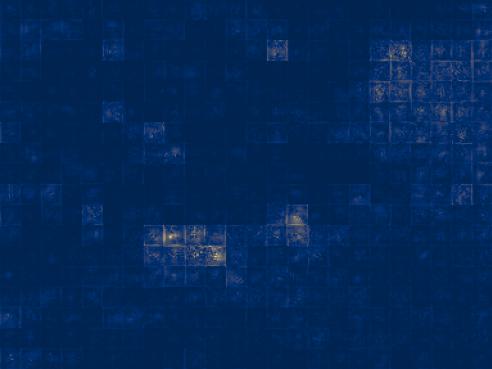}
  \caption*{Saliency}
\end{subfigure}
\begin{subfigure}{.139\textwidth}
  \centering
  \includegraphics[width=0.97\linewidth]{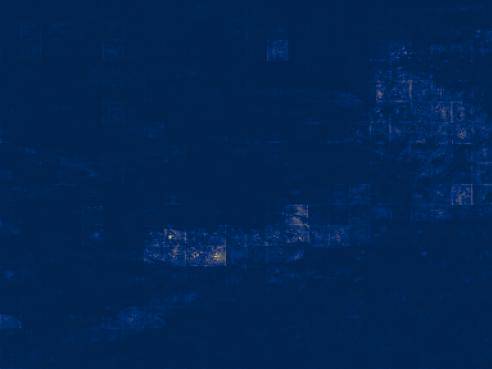}
  \caption*{InputXGrad}
\end{subfigure}
\begin{subfigure}{.139\textwidth}
  \centering
  \includegraphics[width=0.97\linewidth]{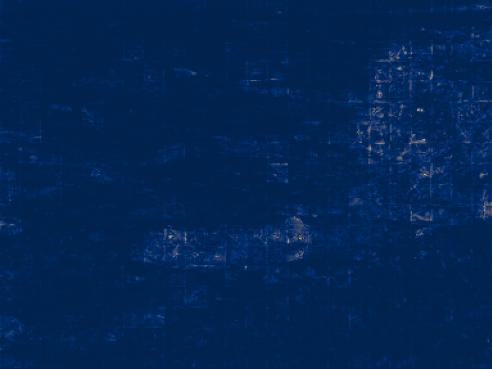}
  \caption*{Integr. Grads}
\end{subfigure}
\begin{subfigure}{.139\textwidth}
  \centering
  \includegraphics[width=0.97\linewidth]{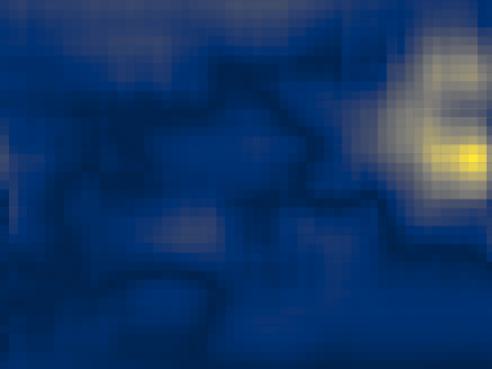}
  \caption*{Occlusion}
\end{subfigure}
\begin{subfigure}{.139\textwidth}
  \centering
  \includegraphics[width=0.97\linewidth]{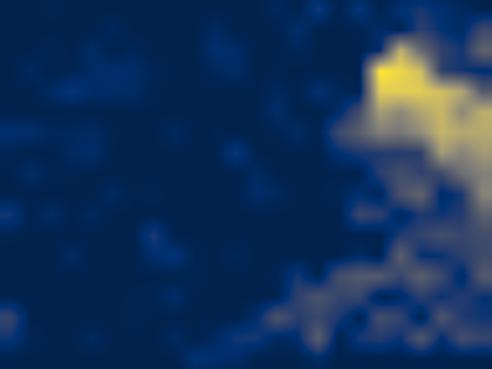}
  \caption*{Grad-CAM}
\end{subfigure}
}

\caption{Cliff}

\end{figure}
\end{document}